\documentclass{article}

\usepackage[preprint]{neurips_2026}

\usepackage[utf8]{inputenc} 
\usepackage[T1]{fontenc}    
\usepackage{hyperref}       
\usepackage{url}            
\usepackage{booktabs}       
\usepackage{amsfonts}       
\usepackage{nicefrac}       
\usepackage{microtype}      
\usepackage{xcolor}         
\usepackage{amsmath, amssymb, amsfonts}
\usepackage{subcaption}
\usepackage{graphicx}
\usepackage{makecell}
\usepackage{multirow}
\usepackage{cleveref}
\usepackage{tcolorbox}
\usepackage{enumitem}

\definecolor{AccuracyColor}{HTML}{1F77B4}
\definecolor{FOneColor}{HTML}{FF7F0E}
\definecolor{AUROCColor}{HTML}{2CA02C}
\definecolor{AUPRCColor}{HTML}{D62728}

\renewcommand{\thesubfigure}{\roman{subfigure}}

\title{Mechanistic Analysis of Alignment Algorithms in Language Models}

\author{%
  Aarush Sinha \\
  University of Copenhagen \\
  \And
  Ishan Garg \\
  Independent \\
  \And
  Veeraraju Elluru \\
  IIT Jodhpur \\
  \And
  Arth Singh \\
  NIT Agartala \\
  \And
  Kushal Garg \\
  Narris \\
}

\begin{document}

\maketitle

\begin{abstract}
    Post-training alignment algorithms are predominantly evaluated as black boxes, obscuring how they reshape language models' internal computations. We present a systematic mechanistic analysis of six preference-optimization methods: PPO, DPO, SimPO, ORPO, GRPO, and KTO across three open-weight model families. By integrating layer-wise linear probing, Sparse Autoencoders, and crosscoders, we localize preference representations and quantify alignment-induced geometric transformations in latent space. We find that preference signals consistently concentrate in early--mid or mid--late layers, but different objectives induce qualitatively distinct representational shifts. KTO and GRPO enhance linear separability through constructive feature sharing and sparse, high-salience recruitment. In contrast, DPO and ORPO degrade separability via non-constructive geometric rotation and feature attenuation, while PPO and SimPO largely preserve baseline geometry. These transformations exhibit architecture-dependent variability, demonstrating that behavioral alignment does not imply uniform internal restructuring. Our findings establish alignment as a heterogeneous intervention, motivate standardized feature-level auditing for safety and interpretability, and highlight the need for mechanism-aware optimization objectives.
\end{abstract}

\section{Introduction}
The rapid scaling of Large Language Models (LLMs) \cite{GPT5.4,glm5team2026glm5vibecodingagentic,geminiteam2025geminifamilyhighlycapable} has yielded systems with remarkable capabilities, yet these models frequently exhibit behaviors that are misaligned with human values and safety requirements. To bridge this gap, the field has converged on post-training algorithms designed to steer model behavior toward helpfulness, harmlessness, and honesty. Reinforcement Learning from Human Feedback (RLHF), typically implemented via Proximal Policy Optimization (PPO), established the initial paradigm for this process \citep{ouyang2022training, schulman2017proximal}. The landscape has since diversified to include direct or reward-model-free preference objectives such as Direct Preference Optimization (DPO) \citep{rafailov2023direct} and other methods, including Simple Preference Optimization (SimPO) \citep{meng2024simpo}, Odds Ratio Preference Optimization (ORPO) \citep{hong2024orpo}, Kahneman-Tversky Optimization (KTO) \citep{ethayarajh2024kto}, and Group Relative Policy Optimization (GRPO) \citep{shao2024deepseekmath}.

Despite the growing algorithmic zoo, the evaluation of these methods has remained largely behavioral. We assess alignment success through aggregate metrics on benchmarks or human preference ratings, effectively treating the model as a black box. While these evaluations confirm that alignment occurs, they offer little insight into how the model's internal computations are reconfigured. This diagnostic gap is critical: without understanding the mechanistic underpinnings of these algorithms, we cannot rigorously predict unintended side effects, such as the degradation of specific capabilities or the emergence of deceptive behaviors. As the community strives for more transparent and reliable AI, the need to move beyond output-level evaluation to a mechanistic understanding of alignment is paramount. 

In this paper, we present a comprehensive diagnostic analysis of the internal effects of alignment algorithms. We compare a representative suite of methods - PPO, DPO, SimPO, ORPO, KTO, and GRPO - to investigate whether these diverse optimization objectives converge on similar internal representations or produce distinct feature-level changes. To do so, we employ several tools commonly used to study model internals: we train Sparse Autoencoders (SAEs) \citep{cunningham2023sae} to decompose superposed activations into interpretable features, use crosscoders \cite{SparseCr41:online} to compare how base-model and aligned-model features share, rotate, or become model-specific, and employ linear probing \cite{tigges2023linearrepresentationssentimentlarge} to detect and localize preference-relevant representations within the residual stream. Prior work has offered important insights into the internal effects of specific alignment methods, but a comparative picture across the most widely used objectives and model architectures remains limited.

Our contributions are as follows:

\begin{enumerate}
    \item \textbf{Comprehensive alignment-diagnostics.} We evaluate \textbf{six} post-training methods across \textbf{three} open-weight model families on the \texttt{UltraFeedback} dataset. We demonstrate that each alignment fine-tuning method shows distinct characteristics through white-box comparisons on the internal representations rather than only model outputs.

    \item \textbf{Feature-level evidence for distinct alignment signatures.} Linear probes, SAEs, and crosscoders reveal that KTO and GRPO improve, while PPO and SimPO maintain (linear) preference decodability. Conversely, DPO and ORPO reduce the (linear) separability on Llama-3.2 and Qwen3 through different feature-geometric changes: DPO is associated with non-constructuve rotation of shared features, whereas ORPO attenuates the preference-relevant activations.

    \item \textbf{Architecture-dependent internal effects.} The same alignment objective can produce inconsistent feature distributions across model families. This shows that claims about the internal effect of an alignment method must be scoped to the base architecture and layer being analyzed.
\end{enumerate}

\begin{figure*}
    \centering
    \includegraphics[width=\linewidth]{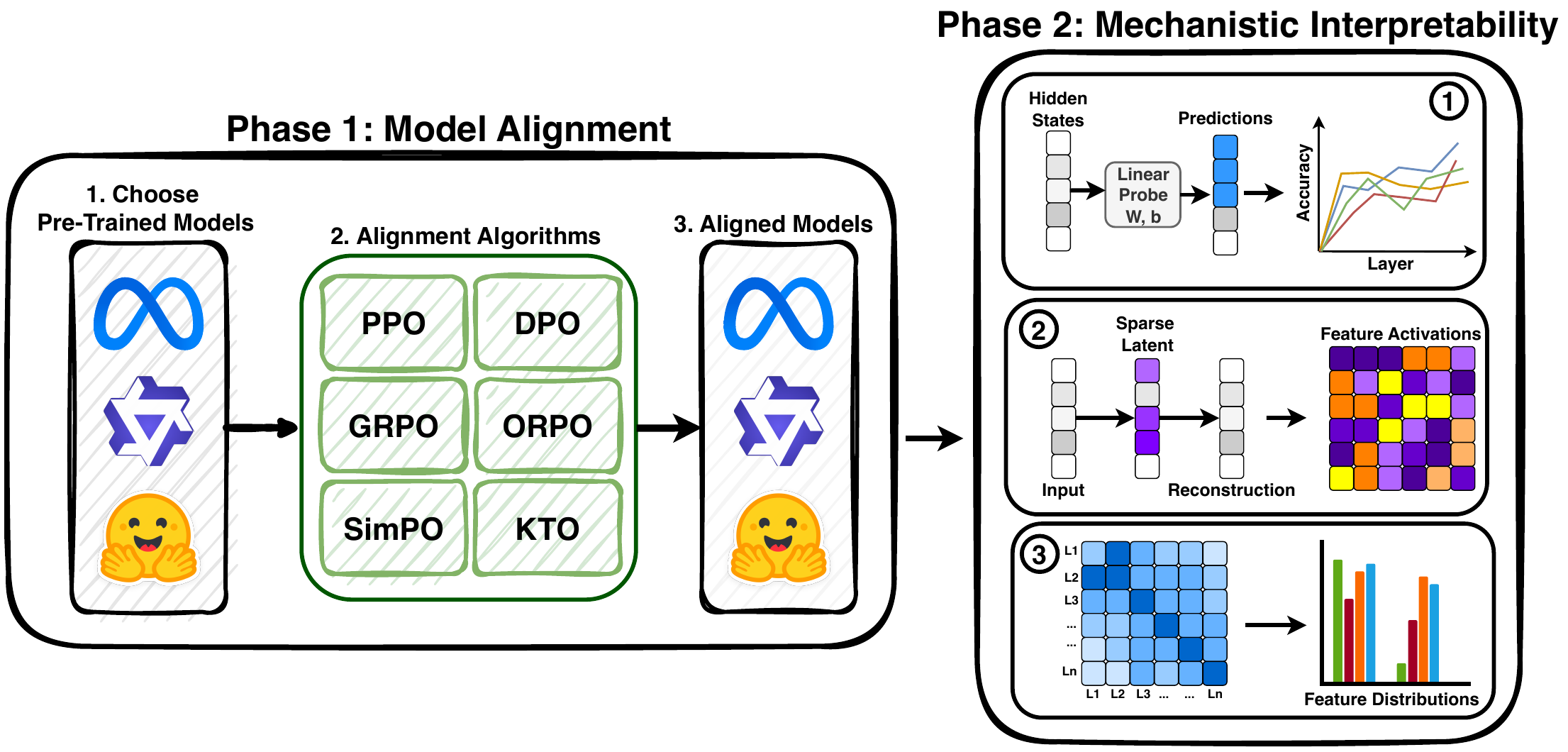}
    \caption{Overall pipeline of our framework consisting of two phases. In Phase 1, we select pre-trained language models for alignment, including Llama-3.2, Qwen3, and SmolLM3, and apply multiple alignment algorithms to obtain aligned variants of each model. In Phase 2, we perform mechanistic interpretability analysis on both the base and aligned models to study representation, feature, and layer-level changes: (1) linear probing is used to identify and compare the most informative layers before and after alignment; (2) Sparse Autoencoders (SAEs) are trained on the selected layers to analyze differences in feature activations; and (3) Crosscoders are trained to compare shared and shifted feature distributions between base and aligned models.}
    \label{fig:main}
\end{figure*}

\section{Related Works}
\begin{table*}[t]
\centering
\caption{MT-Bench \cite{bai2024mt} scores by model family and alignment. \textbf{Ma}/\textbf{Re}: math/reasoning; \textbf{Co}/\textbf{St}: coding/stem; \textbf{Ex}/\textbf{Hu}: extraction/humanities; \textbf{Wr}/\textbf{Rp}: writing/roleplay. Scores are judge means ($n{=}80$ turns per model). \textit{Base} is the public instruct checkpoint for that family. We use GPT-5.4-mini as our LLM evaluator via the OpenAI API.}
\label{tab:mtbench-by-family}
\small
\setlength{\tabcolsep}{4pt}
\resizebox{\textwidth}{!}{%
\begin{tabular}{llrrrrrrrrr}
\toprule
\multicolumn{2}{c}{} & \textbf{Avg.} & \multicolumn{2}{c}{\textbf{Quantitative}} & \multicolumn{2}{c}{\textbf{Technical}} & \multicolumn{2}{c}{\textbf{Knowledge}} & \multicolumn{2}{c}{\textbf{Open-ended}} \\
\cmidrule(lr){3-3}\cmidrule(lr){4-5}\cmidrule(lr){6-7}\cmidrule(lr){8-9}\cmidrule(lr){10-11}
\textbf{Family} & \textbf{Align.} & \textbf{Avg.} & \textbf{Ma} & \textbf{Re} & \textbf{Co} & \textbf{St} & \textbf{Ex} & \textbf{Hu} & \textbf{Wr} & \textbf{Rp} \\
\midrule
\multirow{7}{*}{\textbf{SmolLM-3B}} & Base & 4.84 & 7.30 & 5.35 & 4.75 & 4.95 & 4.50 & 4.10 & 3.80 & 3.95 \\
 & DPO & 4.70 & 7.10 & 5.80 & 4.10 & 5.30 & 4.70 & 3.95 & 2.90 & 3.75 \\
 & GRPO & 4.69 & 7.60 & 5.75 & 3.85 & 5.30 & 4.35 & 3.40 & 3.65 & 3.65 \\
 & KTO & 4.70 & 7.20 & 4.85 & 3.70 & 4.65 & 4.90 & 3.70 & 4.55 & 4.05 \\
 & ORPO & 4.70 & 6.85 & 5.40 & 3.70 & 4.85 & 4.75 & 3.95 & 4.00 & 4.10 \\
 & PPO & 4.94 & 7.55 & 6.35 & 3.75 & 5.05 & 4.95 & 4.30 & 3.70 & 3.85 \\
 & SimPO & 4.59 & 7.20 & 4.75 & 4.80 & 4.65 & 4.80 & 3.75 & 3.30 & 3.45 \\
\addlinespace
\multirow{7}{*}{\textbf{Llama-3.2-3B}} & Base & 6.33 & 8.25 & 6.30 & 4.80 & 5.35 & 7.15 & 5.70 & 6.70 & 6.40 \\
 & DPO & 6.45 & 8.60 & 6.30 & 5.15 & 5.70 & 7.50 & 6.00 & 6.30 & 6.05 \\
 & GRPO & 6.36 & 8.25 & 6.05 & 4.95 & 5.50 & 7.55 & 6.30 & 6.30 & 5.95 \\
 & KTO & 6.31 & 7.80 & 6.00 & 4.90 & 5.55 & 7.50 & 5.70 & 6.80 & 6.25 \\
 & ORPO & 2.12 & 2.75 & 3.10 & 1.45 & 2.80 & 2.20 & 1.30 & 1.70 & 1.65 \\
 & PPO & 6.23 & 7.75 & 5.85 & 5.05 & 5.85 & 6.95 & 6.25 & 6.30 & 5.85 \\
 & SimPO & 6.62 & 8.65 & 6.55 & 5.30 & 5.80 & 7.30 & 6.20 & 6.80 & 6.40 \\
\addlinespace
\multirow{7}{*}{\textbf{Qwen3-4B-Instruct}} & Base & 7.35 & 8.45 & 7.15 & 6.50 & 7.05 & 9.00 & 6.10 & 7.35 & 7.20 \\
 & DPO & 7.39 & 8.85 & 6.95 & 6.45 & 7.00 & 8.85 & 6.70 & 7.50 & 6.80 \\
 & GRPO & 7.35 & 8.55 & 6.40 & 6.20 & 7.20 & 9.20 & 6.45 & 7.50 & 7.30 \\
 & KTO & 7.28 & 8.90 & 6.75 & 6.45 & 6.65 & 9.10 & 6.50 & 7.05 & 6.80 \\
 & ORPO & 5.41 & 6.55 & 5.60 & 4.55 & 5.10 & 6.95 & 4.25 & 6.20 & 4.10 \\
 & PPO & 7.37 & 8.75 & 7.40 & 6.10 & 7.20 & 9.30 & 6.35 & 6.95 & 6.90 \\
 & SimPO & 7.33 & 8.00 & 6.45 & 6.35 & 7.35 & 9.40 & 6.70 & 7.10 & 7.30 \\
\bottomrule
\end{tabular}
}
\end{table*}

\textbf{Preference Optimization Algorithms}
Post-training alignment via RLHF \citep{christiano2023deepreinforcementlearninghuman, bakker2022finetuning, stiennon2022learningsummarizehumanfeedback} optimizes language models using human preference signals and has become a standard stage of model development \citep{ouyang2022training, bai2022traininghelpfulharmlessassistant, touvron2023llama2openfoundation}. Proximal Policy Optimization (PPO) \citep{schulman2017proximal} has been widely used in this setting, though its computational cost and instability have motivated other preference-optimization methods. Direct Preference Optimization (DPO) \citep{rafailov2023direct} is not an RLHF algorithm in the PPO sense; it formulates preference learning as supervised optimization over paired data. Subsequent methods further expand the objective design space: KTO \citep{ethayarajh2024kto} adopts a prospect-theoretic formulation, SimPO \citep{meng2024simpo} removes the reference model, ORPO \citep{hong2024orpo} integrates an odds-ratio penalty into the SFT objective, and GRPO \citep{shao2024deepseekmath} stabilizes policy gradients using group-relative rewards.

\textbf{Mechanistic Interpretability}
Mechanistic interpretability studies both feature-level representations and causal circuits underlying model computation \citep{olah2020zoom, elhage2021mathematical}. Transformer representations arise from interactions between attention heads and MLP layers, with the residual stream acting as a shared communication channel \citep{olsson2022induction}. A central challenge is superposition, where models encode more features than available dimensions \citep{elhage2021mathematical}. Sparse Autoencoders (SAEs) \citep{cunningham2023sae, templeton2024scaling} address this by decomposing activations into sparse, interpretable features. Extensions such as Gemma Scope \citep{lieberum2024gemmascopeopensparse} provide layer-wise SAE coverage for systematic analysis. Linear probing \citep{alain2018understandingintermediatelayersusing} provides a complementary, non-causal measure of whether a concept is linearly decodable from a representation.

Crosscoders \citep{SparseCr41:online} extend sparse feature analysis to model comparison by jointly encoding activations from two related models or layers and then examining the paired decoder geometry. This makes it possible to distinguish features that are shared, base-specific, aligned-specific, amplified, attenuated, or redirected after fine-tuning. Recent work shows that sparsity pressure can create spurious model-exclusive features and proposes improved training and evaluation practices for more reliable model-diffing \citep{minder2026overcoming}.

Another line of interpretability work stems from the causal circuit analysis. Here, causal interventions such as activation patching or component ablations are used to test whether particular attention heads, MLPs, or residual-stream directions are necessary for a behavior. Such work has identified circuits for indirect object identification \citep{wang2022interpretabilitywildcircuitindirect}, induction behavior \citep{olsson2022induction}, and localized factual recall mechanisms in MLP layers \citep{meng2023locatingeditingfactualassociations, geva2021transformerfeedforwardlayerskeyvalue}. Recent work extends causal mechanistic analyses to looped reasoning architectures, demonstrating convergence to cyclic fixed points and stabilization of attention behavior \citep{blayney2026mechanisticanalysisloopedreasoning}. \textbf{Our study uses feature-level diagnostics} and does not work on component-level analyses.

\textbf{Alignment and Internal Representations}
Alignment training alters the geometry of internal representations in ways directly associated with their post-training objectives. Prior work has shown that fine-tuning and RLHF redistribute representations across layers and modulate directions in the residual stream \citep{konen-etal-2024-style}. Crosscoder analyses indicate that chat fine-tuning introduces localized features associated with template tokens \citep{minder2026overcoming}. These results are especially relevant in safety contexts, where post-training may change not only model behavior but also the internal signals used to interpret that behavior. Recent work argues that safety fine-tuning operates through specific internal transformations rather than purely output-level changes \citep{jain2024what}. Post-training algorithms such as RLVR can shift internal states in ways that weaken or evade probe-based detection, even when behavior appears improved \citep{taufeeque2026obfuscationatlas}. DPO reduces linear decodability of toxic features in early layers while shifting them to later layers \citep{lee2024mechanisticunderstandingalignmentalgorithms}. However, these works are not comprehensive across standard alignment fine-tuning methods and model families.
\vspace{-2mm}

\section{Methodology and Experiments}
To ensure the robustness and generality of our findings across diverse architectures, we evaluate three distinct language models: Llama-3.2-3B-Instruct \cite{grattafiori2024llama3herdmodels}, SmolLM3-3B \cite{bakouch2025smollm3}, and Qwen3-4B-Instruct \cite{yang2025qwen3technicalreport}. These models represent state-of-the-art open-weights architectures at the 4B parameter scale, providing a comprehensive testbed for analyzing alignment representations. We utilize the \texttt{Transformers}  \cite{wolf2020huggingfacestransformersstateoftheartnatural} and \texttt{TRL} \cite{vonwerra2020trl} library to align our models.

\subsection{Alignment}
We investigate the internal representations of alignment fine-tuning (AFT) through LoRA fine-tuning the base models using six prominent preference optimization algorithms: PPO, SimPO, GRPO, DPO, ORPO, and KTO. We utilize different variants of the UltraFeedback \cite{cui2024ultrafeedbackboostinglanguagemodels} dataset to ensure compatibility with each alignment method. For DPO and SimPO, we utilize the vanilla version\footnote{\href{https://huggingface.co/datasets/argilla/ultrafeedback-binarized-preferences-cleaned}{argilla/ultrafeedback-binarized-preferences-cleaned}}. For PPO and GRPO, which necessitate multiple generation samples per prompt, we employ the \texttt{multi-binarized} variant\footnote{\href{https://huggingface.co/datasets/argilla/ultrafeedback-multi-binarized-preferences-cleaned}{argilla/ultrafeedback-multi-binarized-preferences-cleaned}}. Similarly for KTO, we use the corresponding version\footnote{\href{https://huggingface.co/datasets/argilla/ultrafeedback-binarized-preferences-cleaned-kto}{argilla/ultrafeedback-binarized-preferences-cleaned-kto}} which provides independent prompt-completion pairs with binary desirability labels. All models are fine-tuned using Low-Rank Adaptation (LoRA) \cite{hu2021loralowrankadaptationlarge} on the query, key, value, and output projection matrices ($r=16$, $\alpha=32$) to maintain computational efficiency while allowing sufficient expressivity for alignment. The hyperparameters are outlined in Appendix$\S$\ref{app:hyperparams}.

\subsection{Linear Probes}
To decode the layer-wise emergence of preference representations, we train linear probes on the models' internal activations. For a given prompt, we extract the final token's residual stream representation for both the chosen ($x_c$) and rejected ($x_r$) completions, independently for each layer. We formulate this as a contrastive classification task by computing the difference vector $\Delta x = x_c - x_r$. To prevent the probe from relying on absolute activation magnitudes, we construct a symmetric dataset comprising positive examples ($+\Delta x$, label 1) and negative examples ($-\Delta x$, label 0). 

We train a logistic regression classifier with balanced class weights on these representations. The probes are optimized using Adam \cite{kingma2017adammethodstochasticoptimization} (learning rate $0.05$), and evaluated using Accuracy, F1-score, AUROC, and AUPRC. Furthermore, in \cref{fig:lp_pca_grid_smollm3}--\cref{fig:lp_pca_grid_qwen} in Appendix~\ref{subsec-appendix:linear-probe=diagnostics}, we show the varied degrees of linear separability of the chosen and rejected representations at the best layer via PCA.

\subsection{Sparse Autoencoders}
While the probes help us localize the maximal preference-specific representations, these are inherently \textit{superposed}. To decompose the dense, polysemantic representations of the aligned models into interpretable, monosemantic features, we train Batch Top-$K$ Sparse Autoencoders (SAEs) \cite{bussmann2024batchtopk, bricken2023monosemanticity}. The SAEs are trained on activations extracted from the \texttt{UltraChat}\footnote{\href{https://huggingface.co/datasets/openbmb/UltraChat}{openbmb/UltraChat}}\cite{ding2023enhancingchatlanguagemodels} dataset, utilizing a context window of $1024$ tokens. 

The SAEs consist of a dictionary size of $d_{\text{SAE}} = 4096$ and an $L_0$ of $K=64$. The models are trained for $200,000$ tokens with a batch size of $2048$ tokens. Optimization is performed using Adam ($\beta_1=0.9$, $\beta_2=0.999$) with a peak learning rate of $3 \times 10^{-4}$, incorporating a $100$-step linear warmup. We apply an auxiliary loss coefficient of $1.0$ to encourage dead feature revival, initialize the decoder norm to $0.1$, and set the Top-$K$ threshold learning rate to $0.01$. This configuration ensures a high-fidelity reconstruction of the original activations while strictly bounding the $L_0$ norm of the feature activations.

\paragraph{Finding monosemantic features.} For each model family, we identify an ``anchor'' feature by computing the mean activation across the entire set of prompts and selecting the feature (decoder column) with the maximum mean activation. We then measure the activation of this anchor feature in the aligned models at the corresponding token positions. This readout uses the base model SAE to ensure a consistent latent space for comparison.

Formally, for each alignment fine-tuned model $M_\alpha$, we extract the residual activations $h_L(x)$ at two layers: (1) $L_{\text{same}}$, the layer where the base SAE was trained, and (2) $L_{\text{best}}$, the layer with highest linear probe AUROC for the alignment task. We apply the base SAE encoder to these residuals and extract the activation of the fixed anchor feature: $a^{(\alpha, L)} = S_0(h_L^{(\alpha)}(x))_{f^*}$. 

To avoid selecting generic or template-driven features, this discovery is restricted to semantic content-bearing token positions. We exclude special tokens, early template-specific positions, and positions where the same token appears across many prompts. We also limit the number of anchor positions per prompt to avoid repeated patterns dominating the anchor set. The final set of anchors are chosen such that they are relevant to AFT, e.g., separating preferred from dispreferred responses or aligning with the layer-wise preference probe signal. We summarize the 

\subsection{Crosscoders}
The SAE experiments help us interpret the polysemantic latent space within aligned models. To more directly compare the base and aligned representations, we make use of crosscoders~\cite{SparseCr41:online}, with modifications from \cite{mishrasharma-2025} and \cite{nasiri-sarvi2026sparc} for universality. A crosscoder jointly learns sparse features over paired hidden states from the base and aligned models, allowing us to compare whether a learned feature is \textit{shared} across both models or \textit{preferentially reconstructed through model-exclusive decoders}. From both the base and aligned models, the residual-stream activations are extracted from \textbf{three} layers (``best'' layer $\pm1$ from linear probing), and averaged for a more robust intervention. We use the \texttt{UltraFeedBack} preference dataset with pooling over the final prompt token. Activations from both models are normalized and passed through two independently parameterized encoders $E_b, E_a : \mathbb{R}^{d} \to \mathbb{R}^{M}$ with a shared global Top-$K$ activation budget. The decoders $D_b, D_a$ reconstruct each stream.

The crosscoder dictionary has an expansion factor of $8$, Top-$K = 400$, and a forced-shared subspace comprising $6\%$ of features to discourage the learned dictionary from explaining shared structure using only model-exclusive features~\cite{mishrasharma-2025}. The training objective consists of three terms: (i) per-stream reconstruction loss, (ii) cross-reconstruction loss, and (iii) sparsity and shared-subspace regularization over the learned feature activations and decoder geometry. The details of the training hyperparameters are summarized in Appendix$\S$\ref{app:crosscoder-details}.

Following \cite{elluru2026mechanisticallyinterpretingcompressionvisionlanguage}'s feature geometry evaluation, after training, each feature is classified using two geometric statistics derived from the decoder columns: $\rho = \|W_{a,\text{dec}}\| / (\|W_{b,\text{dec}}\| + \|W_{a,\text{dec}}\|)$ measures the share of the feature's total decoder norm carried by the aligned stream, and $\theta$ is the angle between $W_{b,\text{dec}}$ and $W_{a,\text{dec}}$. A Gaussian Mixture Model (GMM) is fit to the distribution of $\rho$ values to estimate decision boundaries for base-only, shared, and aligned-only features; angular thresholds then separate shared features that preserve, amplify, attenuate, or redirect the base-model direction. The exact values of these thresholds are deferred to Appendix $\S$\ref{app:crosscoder_metrics}. 

The MT-Bench results (Table \ref{tab:mtbench-by-family}) show that alignment gains are not uniform across model families or objectives. Qwen remains strong across most settings, while Llama benefits from DPO/SimPO-style tuning but collapses under ORPO, dropping from a 6.33 base average to 2.12 with consistently poor category scores. SmolLM is comparatively flat, suggesting that aggregate black-box scores alone can hide whether an alignment method is improving behavior, leaving the base model mostly unchanged, or damaging internal capabilities. This makes a strong case for white-box evaluations: in the ORPO case especially, surface-level outputs reveal the failure, but white-box probes are needed to diagnose whether the degradation comes from representation drift, over-optimization of preference signals, loss of instruction-following circuits, or other internal changes that are invisible from final answer scores alone.

\section{Results}
In this section, we present the results from the probing experiments on the three models across the six alignment algorithms. We then use these results for performing feature interpretations at the ``best'' layer in the SAE latent space. Lastly, we show how the feature geometry changes due to alignment at this layer. 

\begin{figure*}[t!]
    \centering
    \scriptsize

    \begin{minipage}{\textwidth}
        \centering
        \setcounter{subfigure}{0}
        \renewcommand{\thesubfigure}{\alph{subfigure}}
        \begin{subfigure}[!ht]{0.125\textwidth}
            \includegraphics[width=\linewidth]{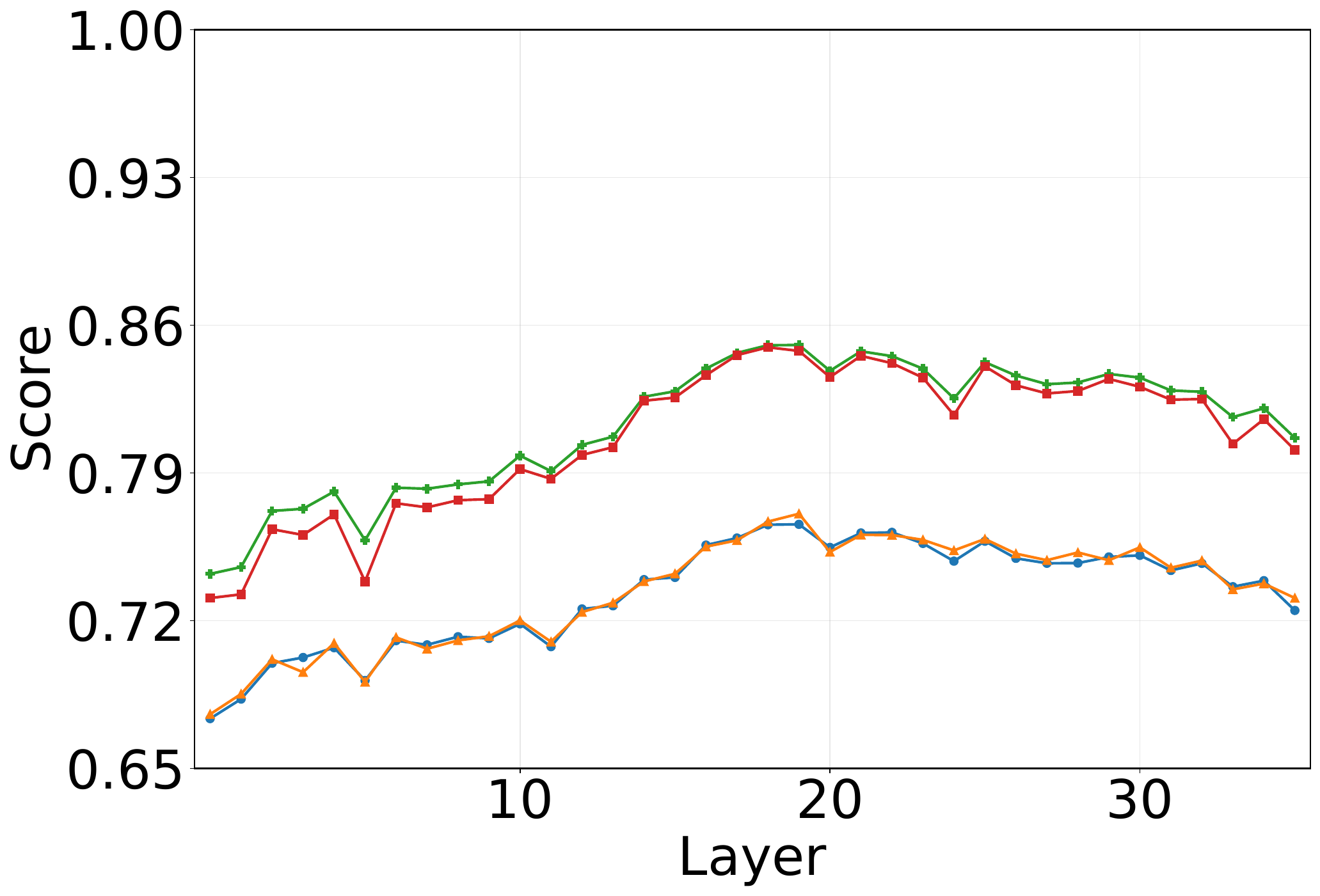}
            \caption{Base}
        \end{subfigure}\hfill
        \begin{subfigure}[!ht]{0.125\textwidth}
            \includegraphics[width=\linewidth]{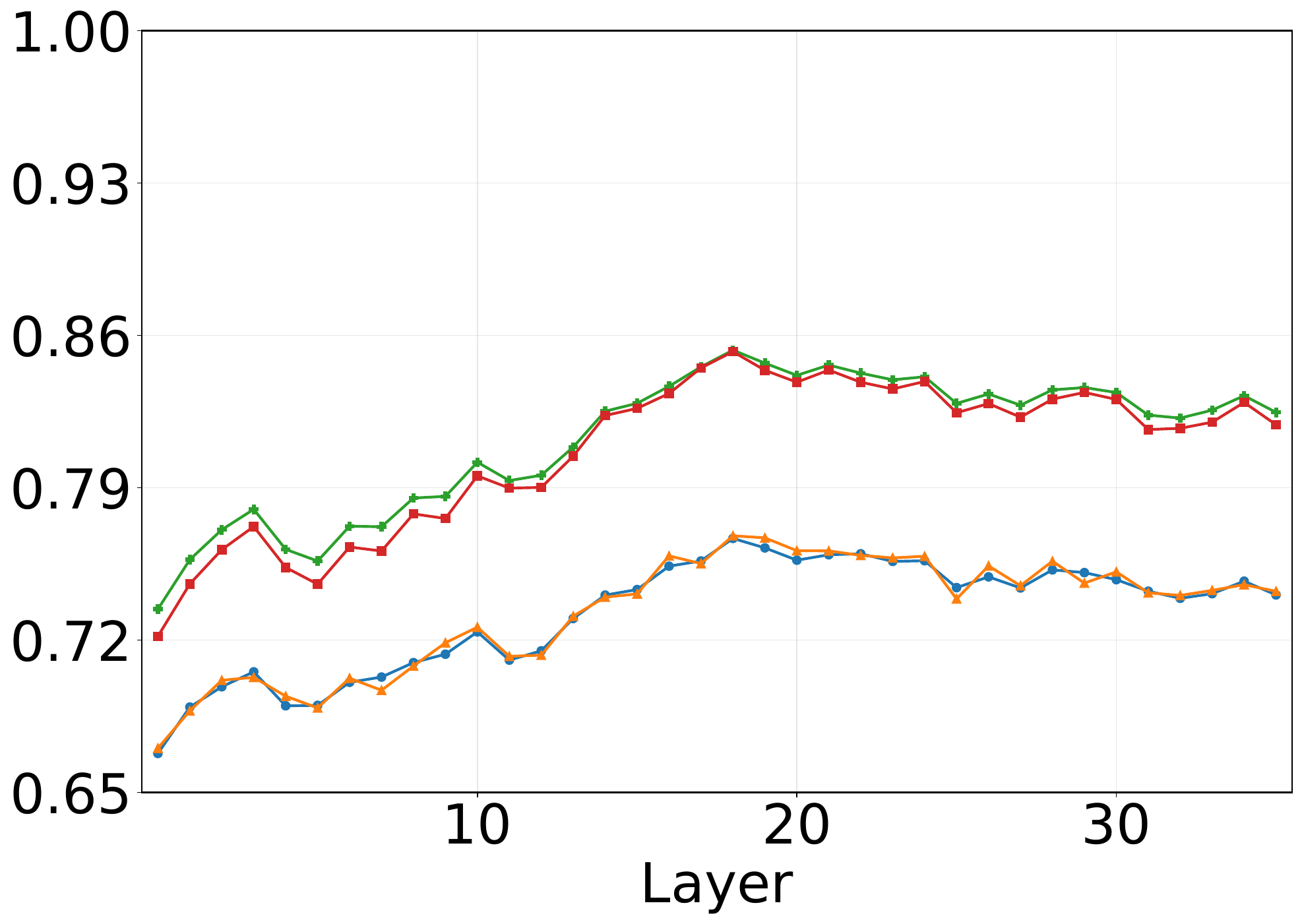}
            \caption{DPO}
        \end{subfigure}\hfill
        \begin{subfigure}[!ht]{0.125\textwidth}
            \includegraphics[width=\linewidth]{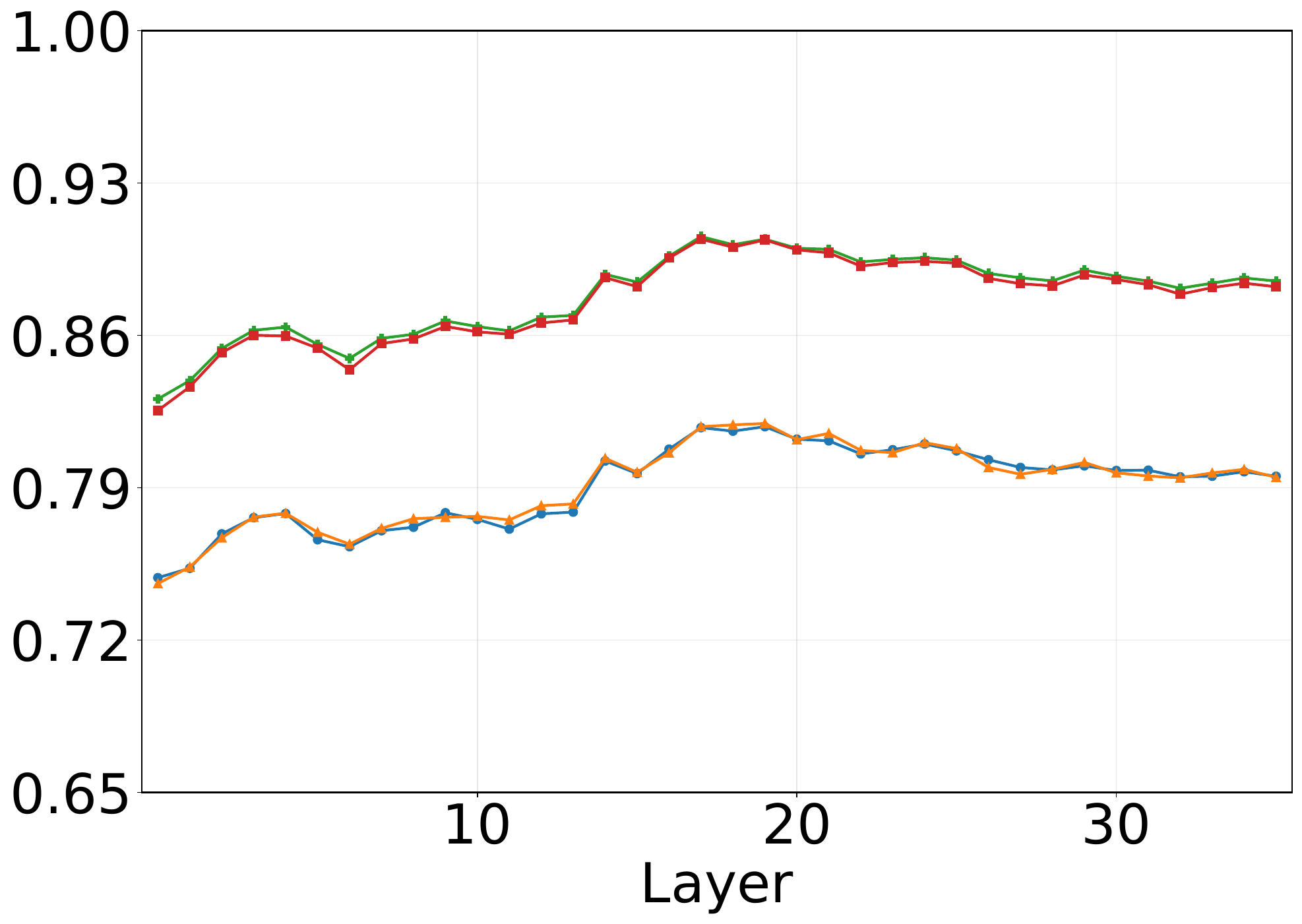}
            \caption{GRPO}
        \end{subfigure}\hfill
        \begin{subfigure}[!ht]{0.125\textwidth}
            \includegraphics[width=\linewidth]{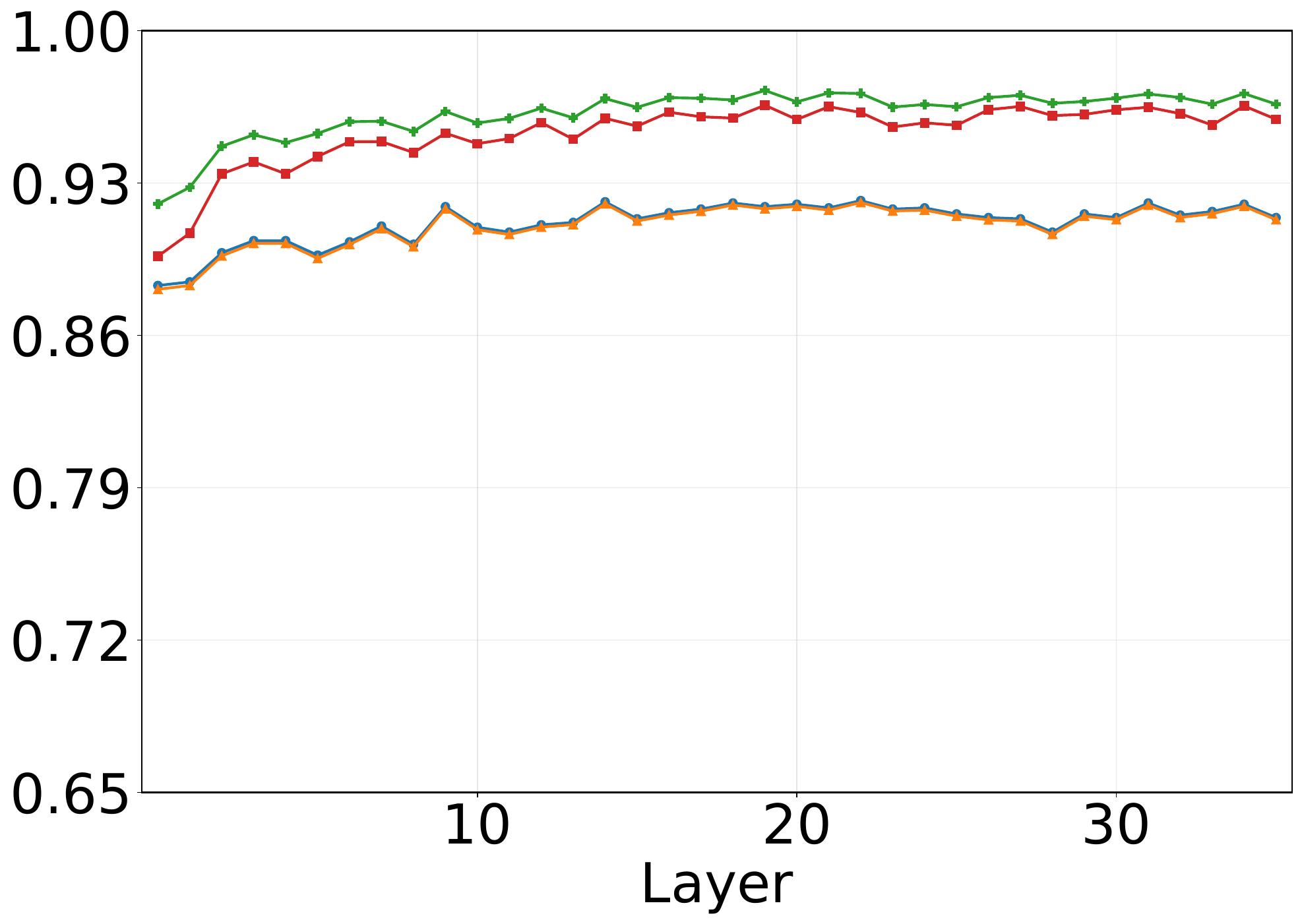}
            \caption{KTO}
        \end{subfigure}\hfill
        \begin{subfigure}[!ht]{0.125\textwidth}
            \includegraphics[width=\linewidth]{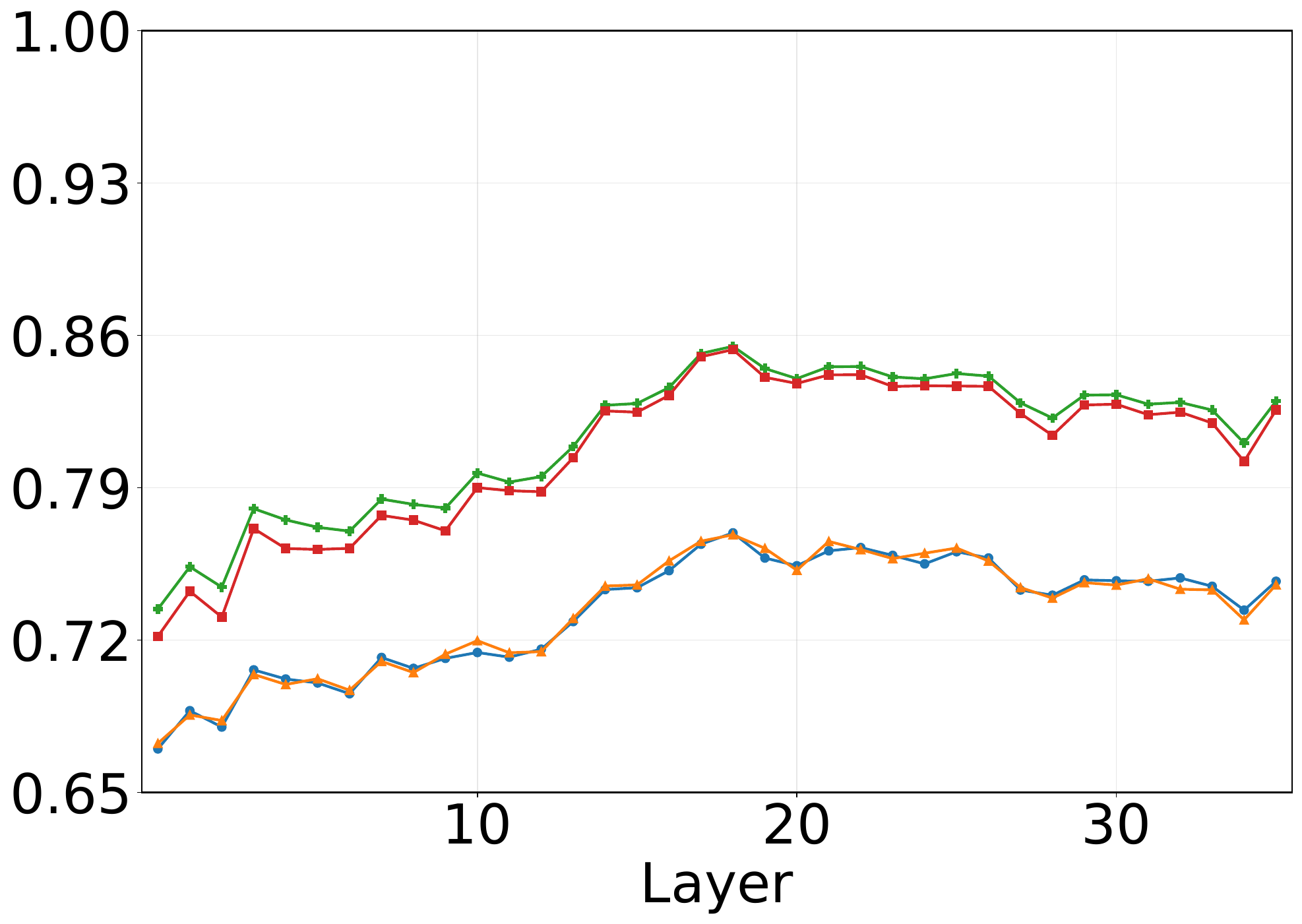}
            \caption{ORPO}
        \end{subfigure}\hfill
        \begin{subfigure}[!ht]{0.125\textwidth}
            \includegraphics[width=\linewidth]{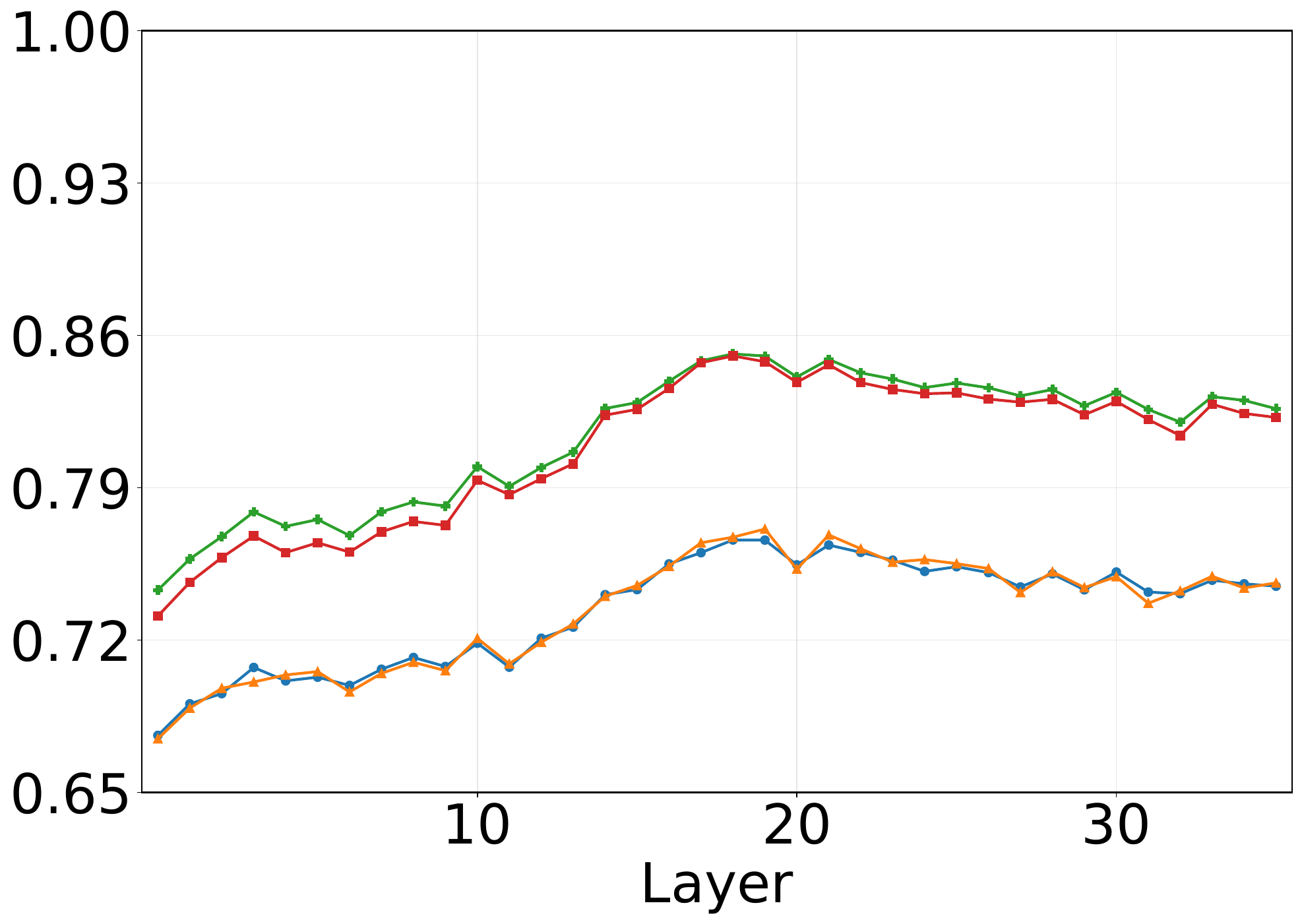}
            \caption{PPO}
        \end{subfigure}\hfill
        \begin{subfigure}[!ht]{0.125\textwidth}
            \includegraphics[width=\linewidth]{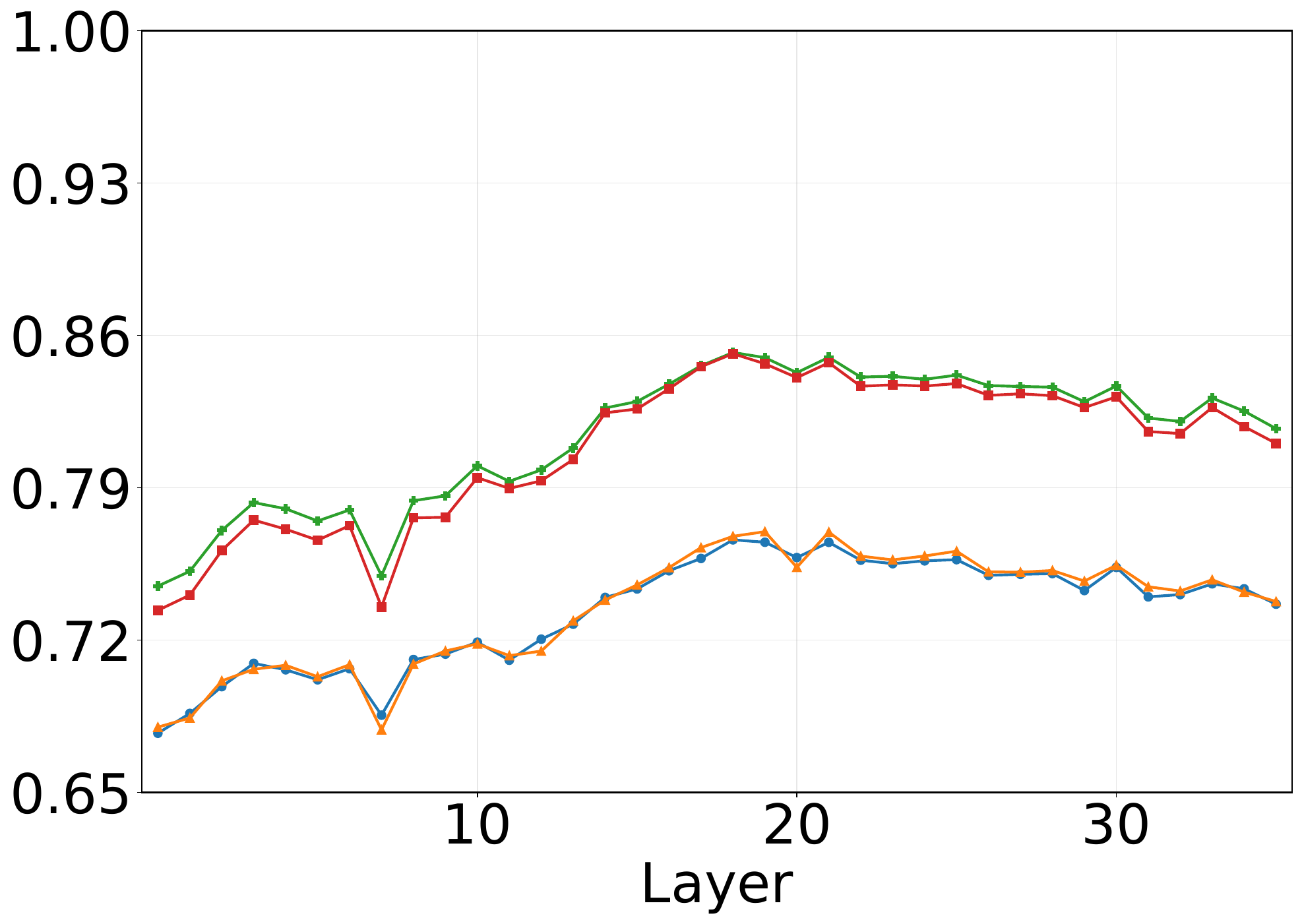}
            \caption{SimPO}
        \end{subfigure}

        \vspace{0.3em}
        \textbf{(i) SmolLM3-3B}
    \end{minipage}

    \vspace{0.6em}

    \begin{minipage}{\textwidth}
        \centering
        \setcounter{subfigure}{0}
        \renewcommand{\thesubfigure}{\alph{subfigure}}
        \begin{subfigure}[!ht]{0.125\textwidth}
            \includegraphics[width=\linewidth]{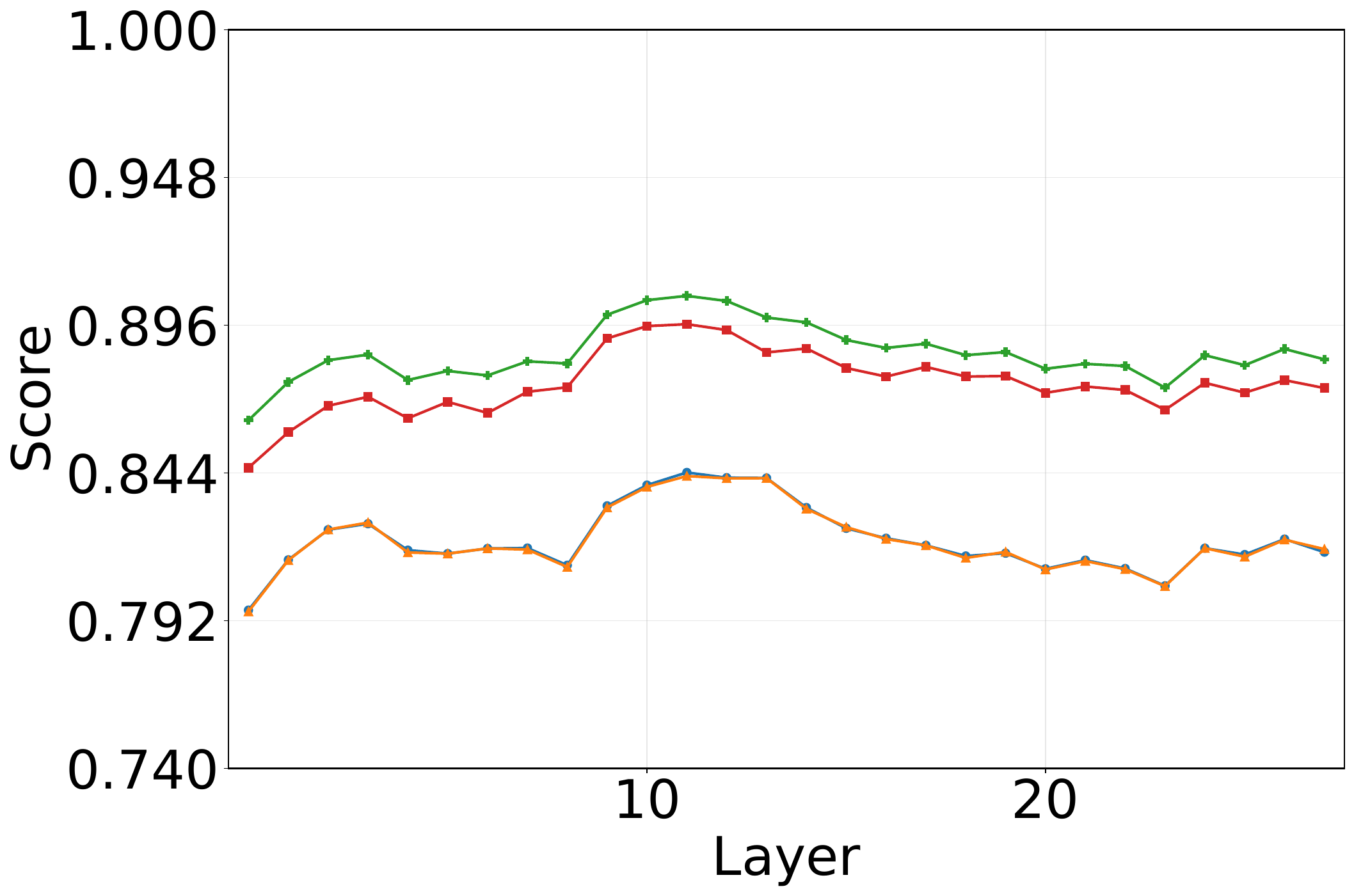}
            \caption{Base}
        \end{subfigure}\hfill
        \begin{subfigure}[!ht]{0.125\textwidth}
            \includegraphics[width=\linewidth]{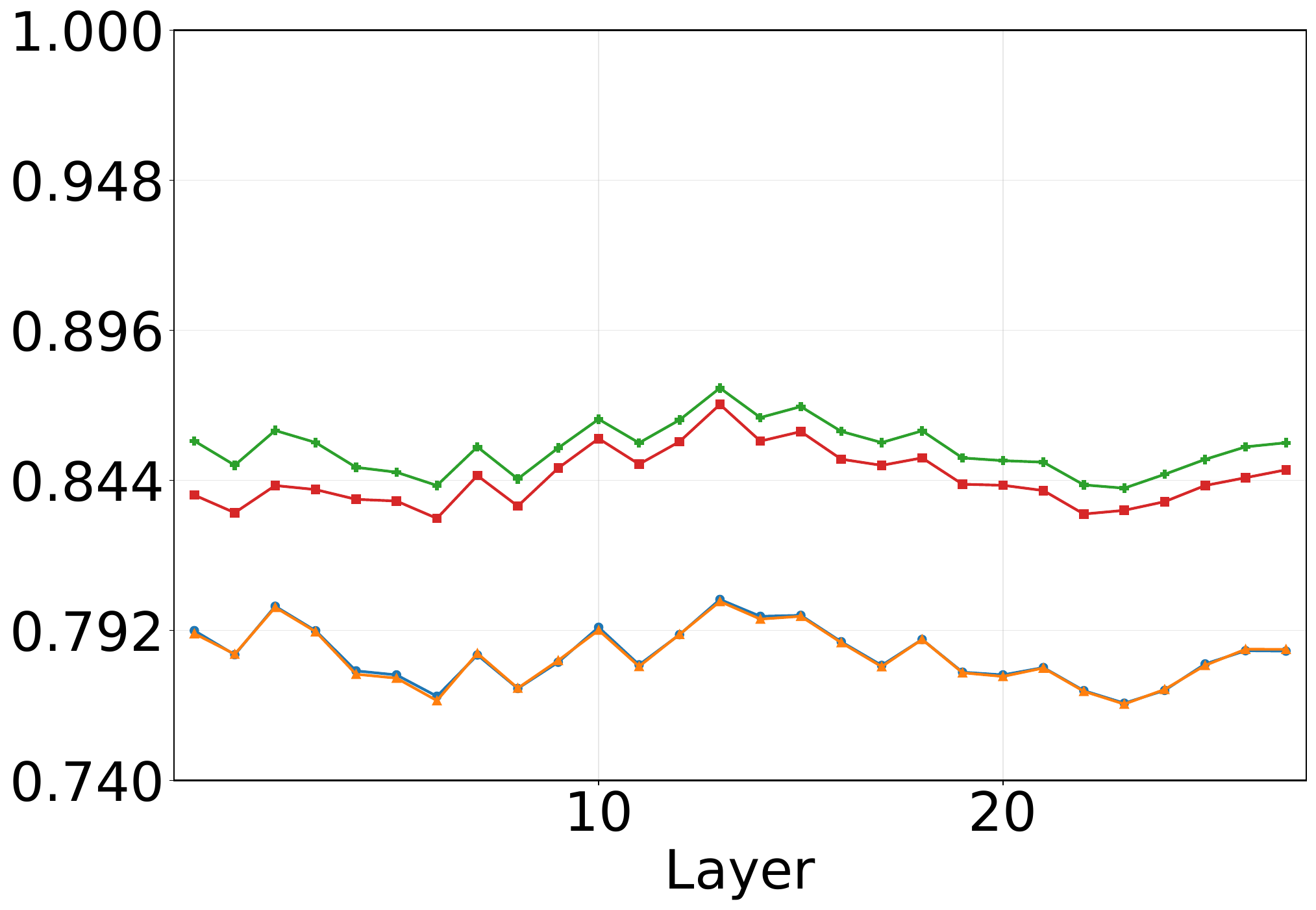}
            \caption{DPO}
        \end{subfigure}\hfill
        \begin{subfigure}[!ht]{0.125\textwidth}
            \includegraphics[width=\linewidth]{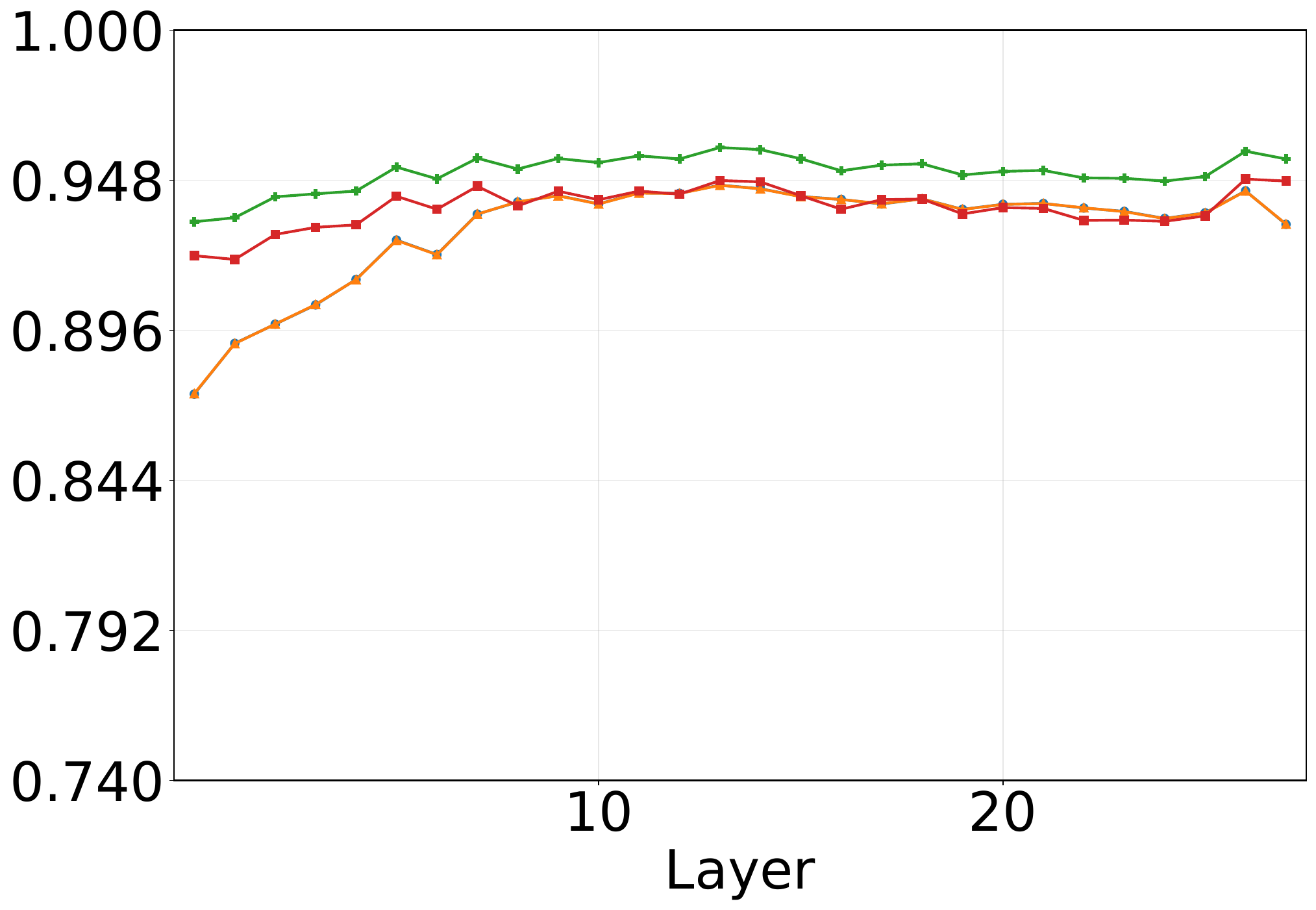}
            \caption{GRPO}
        \end{subfigure}\hfill
        \begin{subfigure}[!ht]{0.125\textwidth}
            \includegraphics[width=\linewidth]{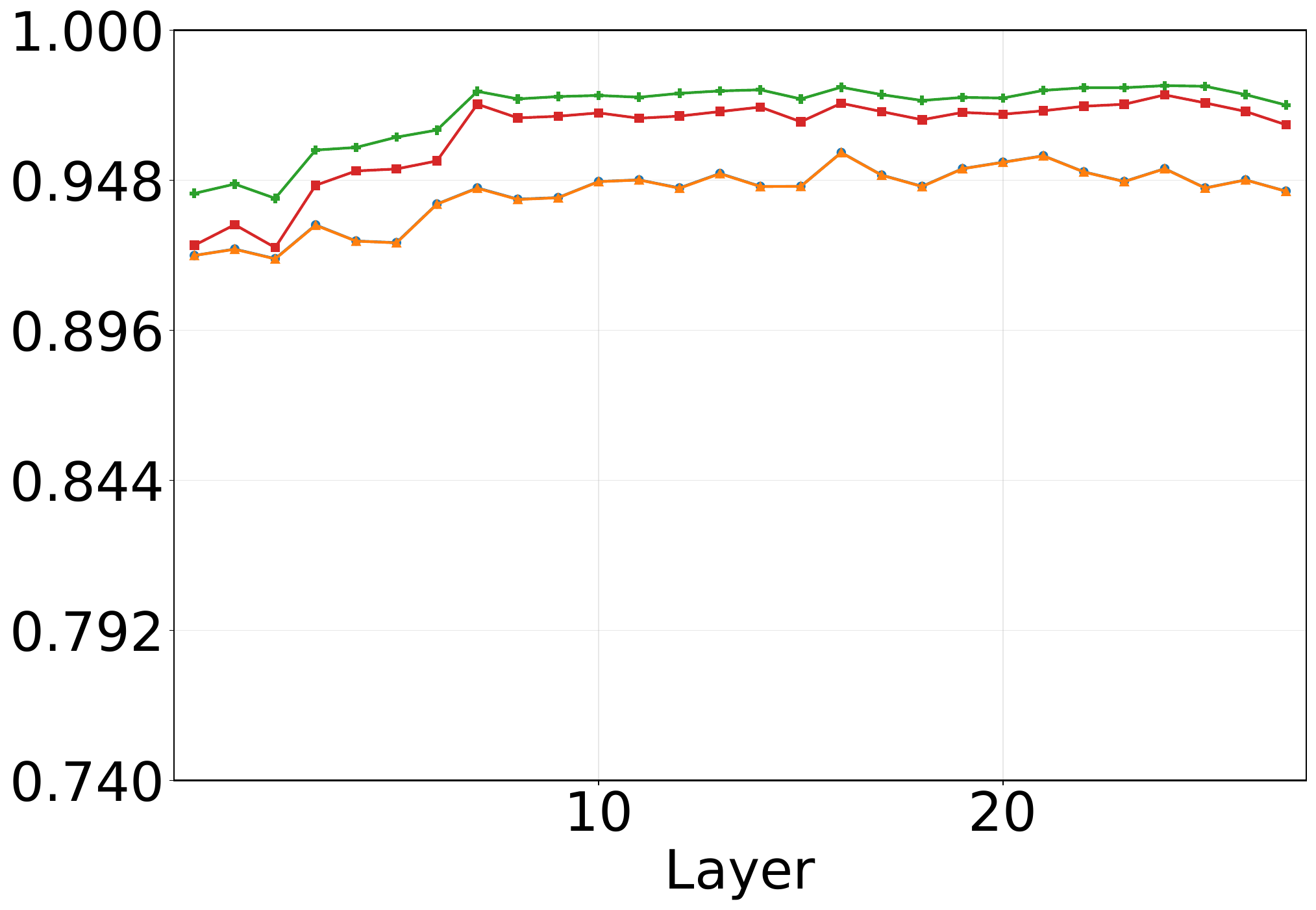}
            \caption{KTO}
        \end{subfigure}\hfill
        \begin{subfigure}[!ht]{0.125\textwidth}
            \includegraphics[width=\linewidth]{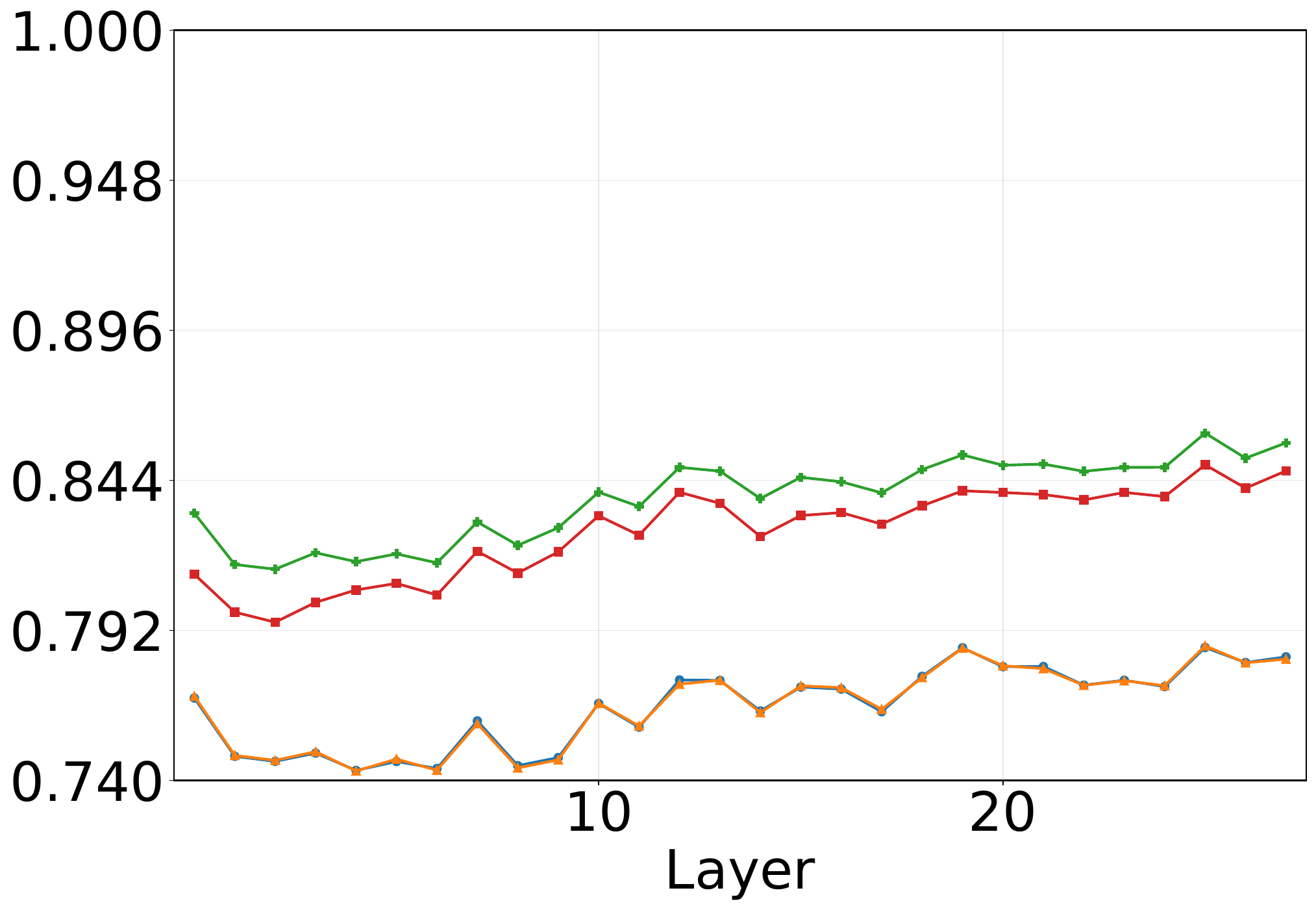}
            \caption{ORPO}
        \end{subfigure}\hfill
        \begin{subfigure}[!ht]{0.125\textwidth}
            \includegraphics[width=\linewidth]{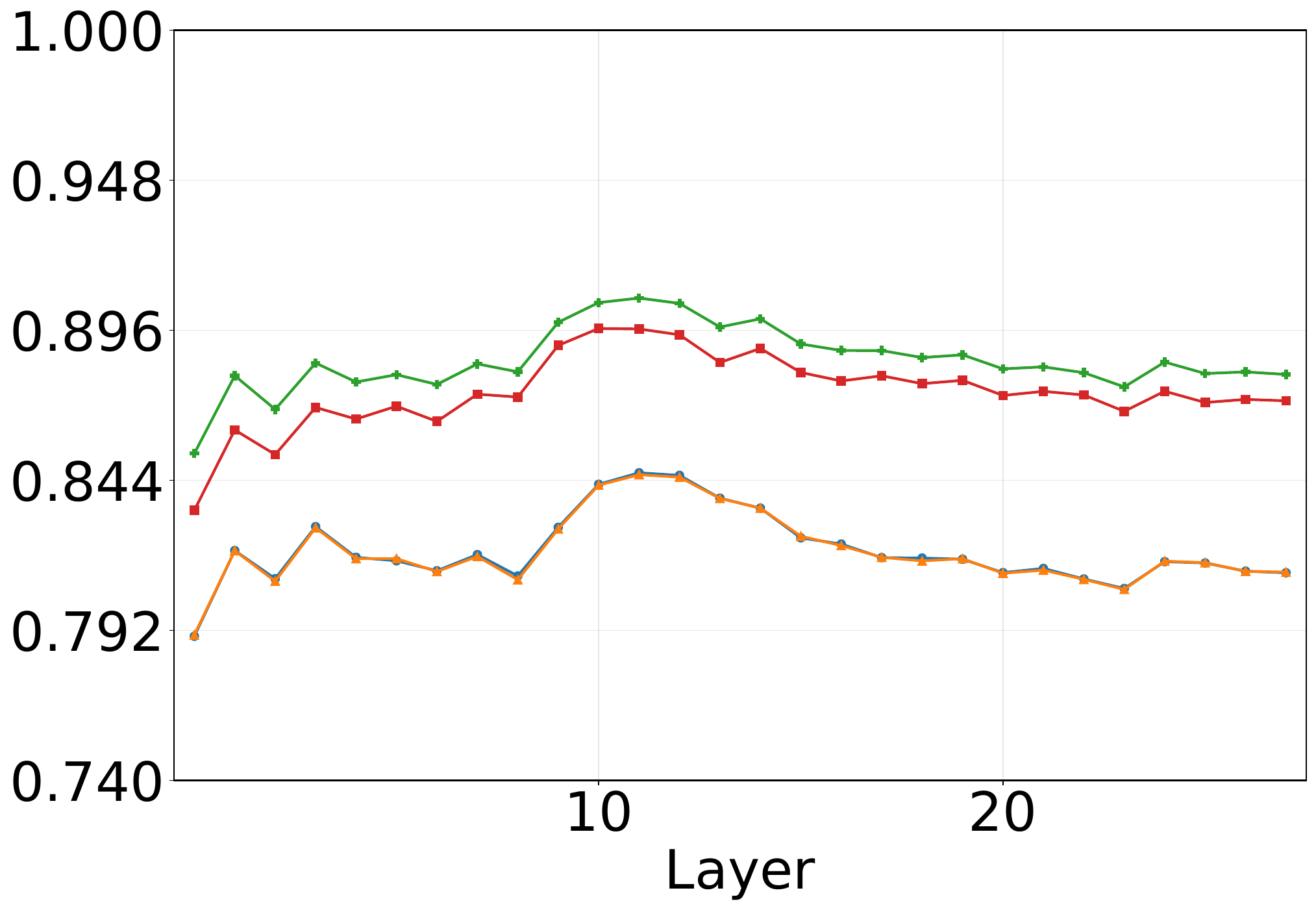}
            \caption{PPO}
        \end{subfigure}\hfill
        \begin{subfigure}[!ht]{0.125\textwidth}
            \includegraphics[width=\linewidth]{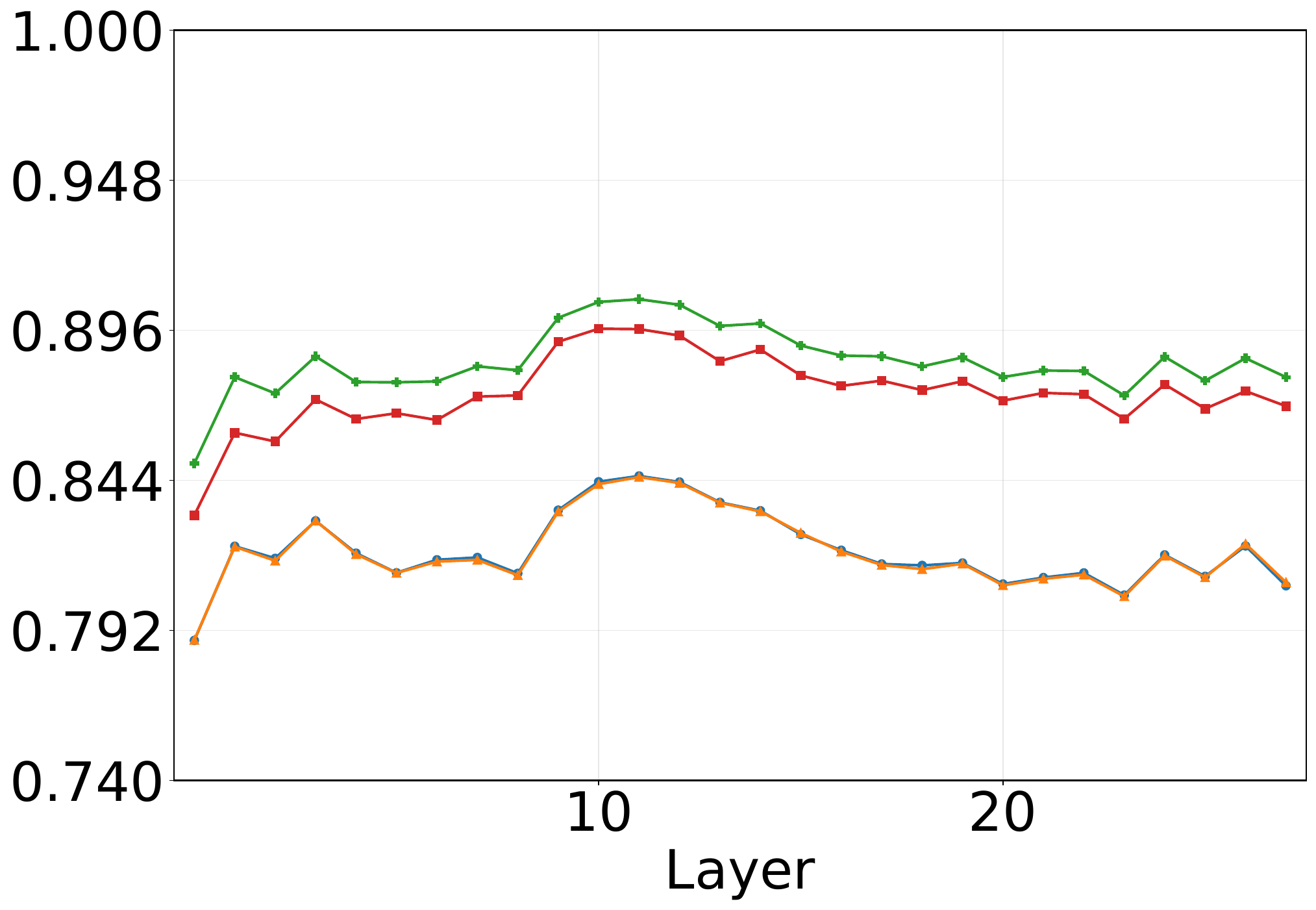}
            \caption{SimPO}
        \end{subfigure}

        \vspace{0.3em}
        \textbf{(ii) Llama-3.2-3B}
    \end{minipage}

    \vspace{0.6em}

    \begin{minipage}{\textwidth}
        \centering
        \setcounter{subfigure}{0}
        \renewcommand{\thesubfigure}{\alph{subfigure}}
        \begin{subfigure}[!ht]{0.125\textwidth}
            \includegraphics[width=\linewidth]{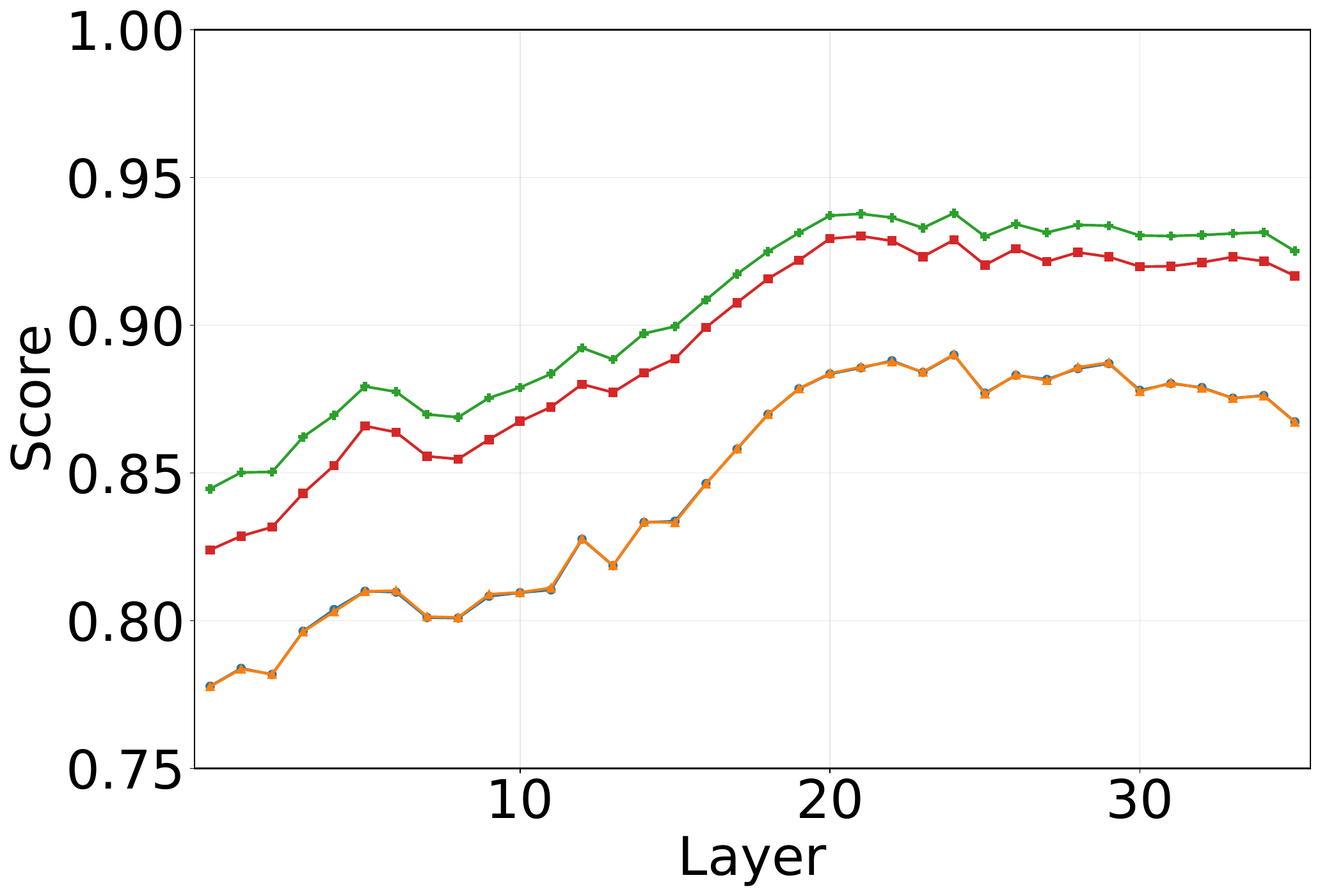}
            \caption{Base}
        \end{subfigure}\hfill
        \begin{subfigure}[!ht]{0.125\textwidth}
            \includegraphics[width=\linewidth]{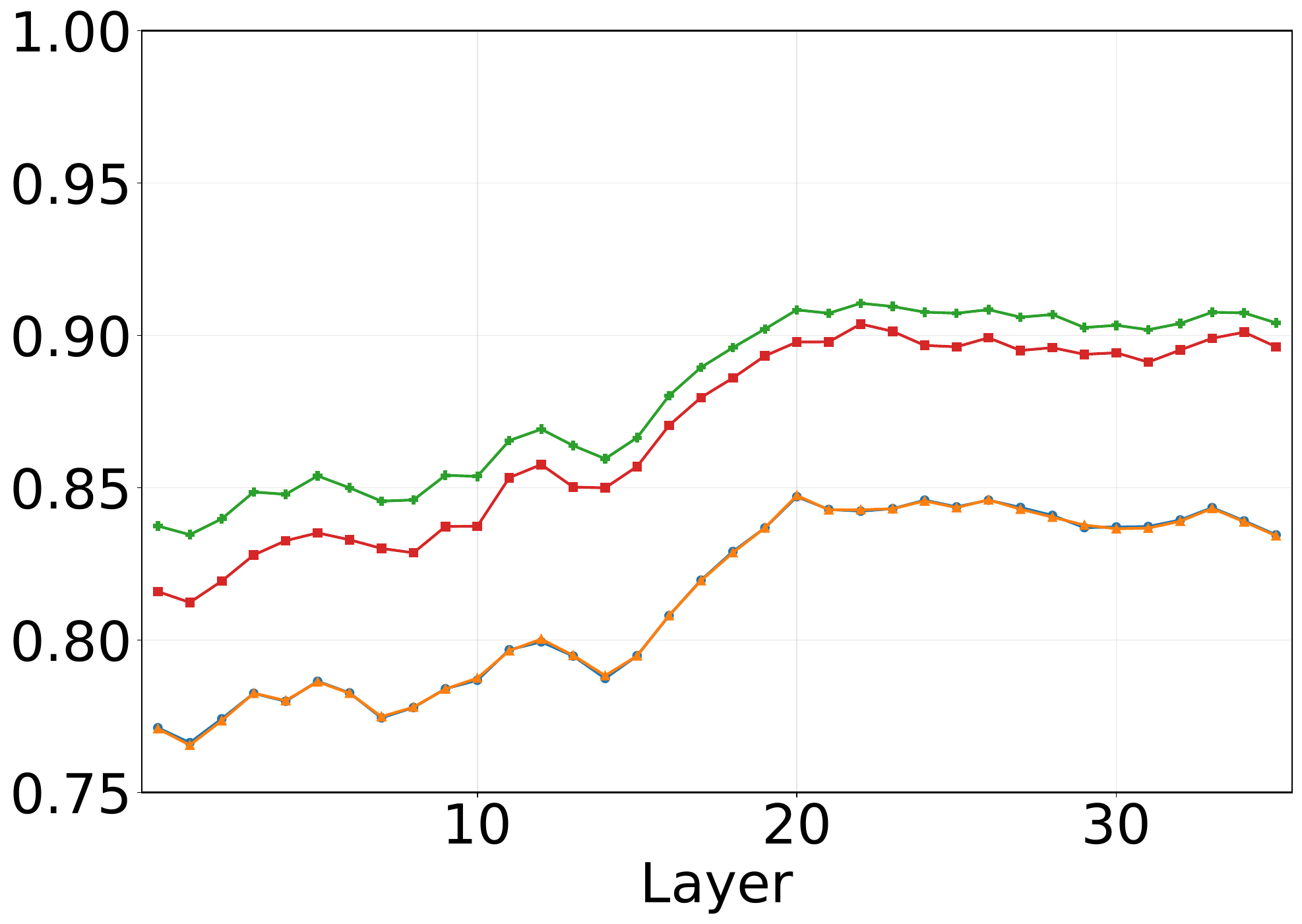}
            \caption{DPO}
        \end{subfigure}\hfill
        \begin{subfigure}[!ht]{0.125\textwidth}
            \includegraphics[width=\linewidth]{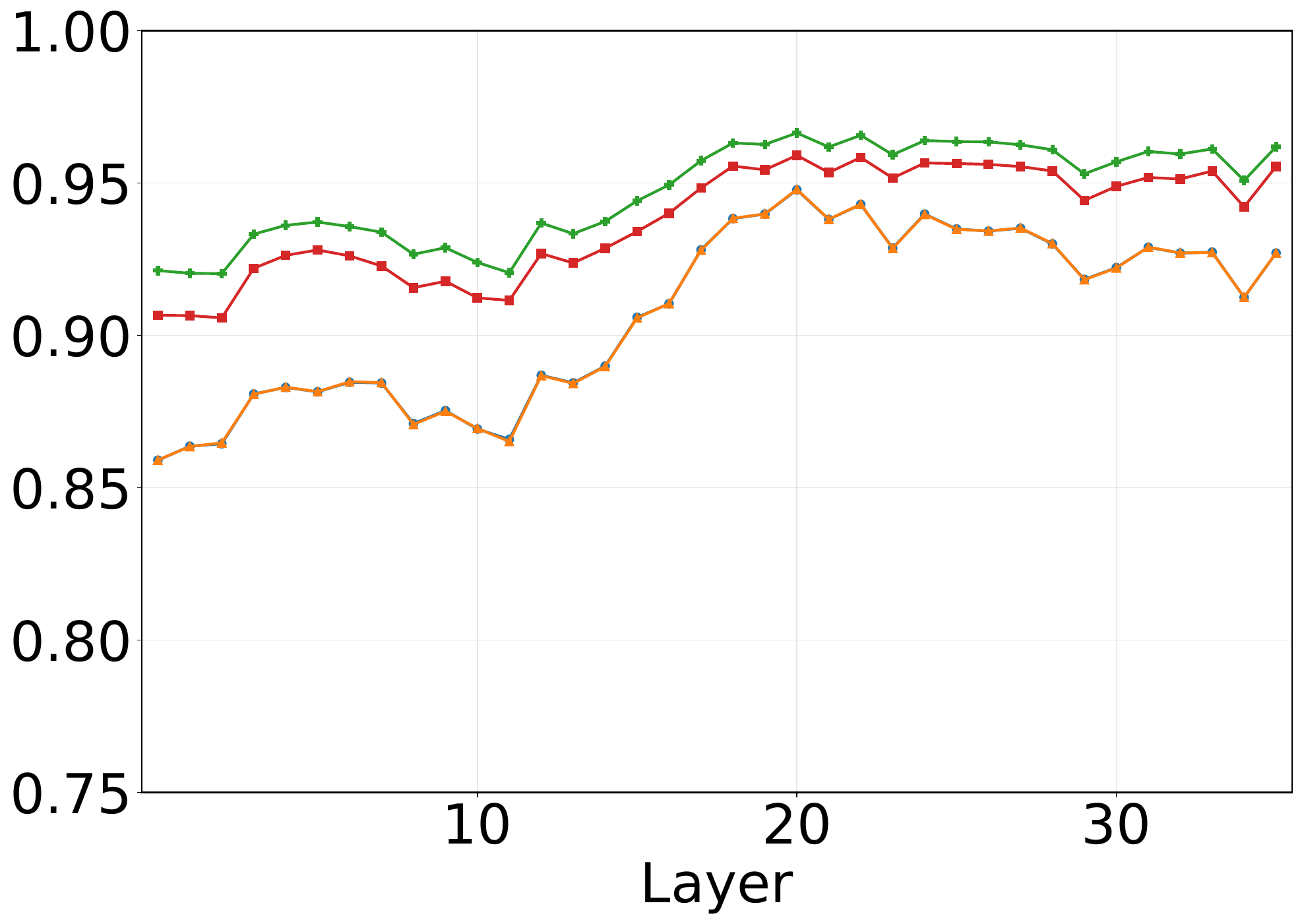}
            \caption{GRPO}
        \end{subfigure}\hfill
        \begin{subfigure}[!ht]{0.125\textwidth}
            \includegraphics[width=\linewidth]{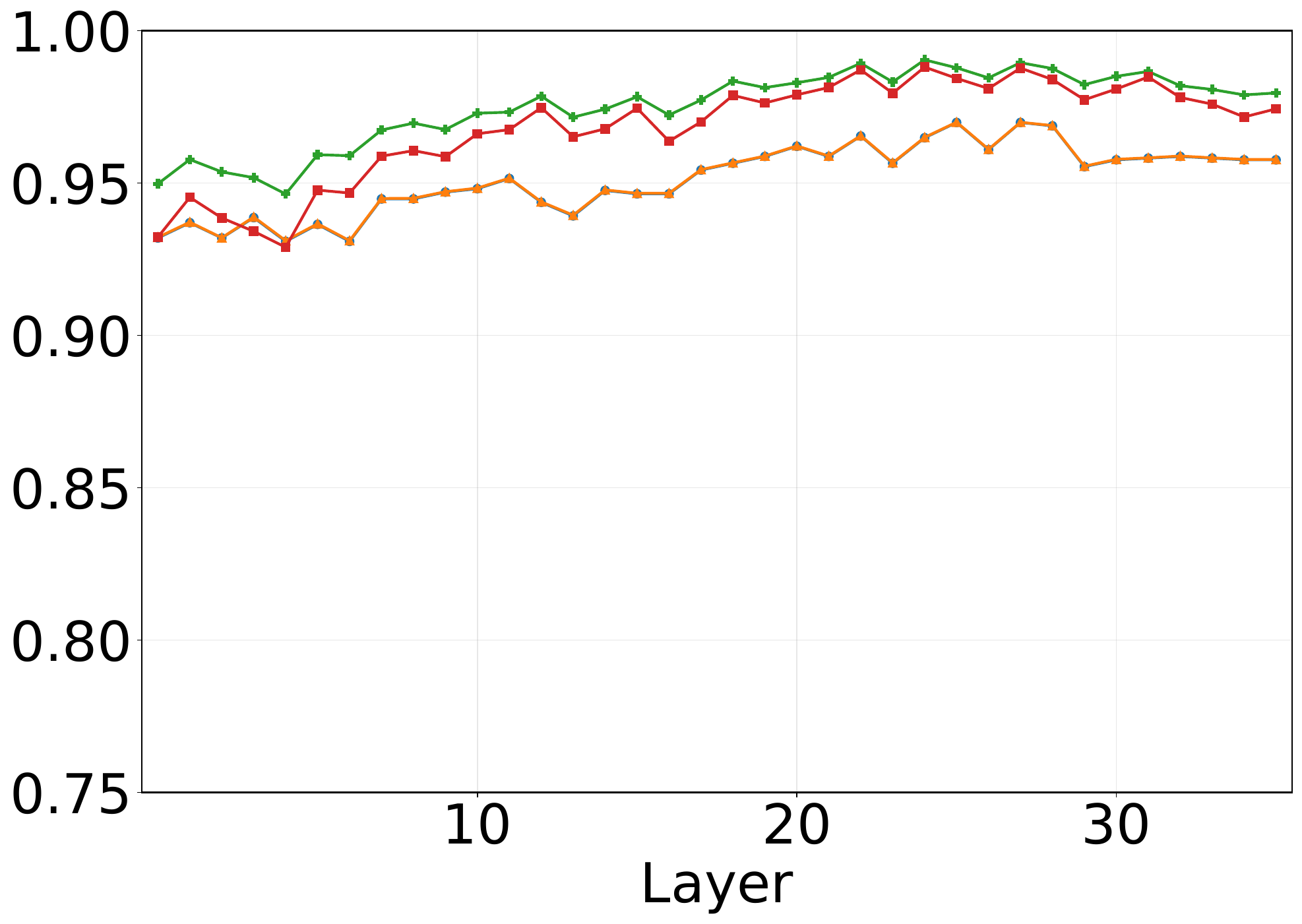}
            \caption{KTO}
        \end{subfigure}\hfill
        \begin{subfigure}[!ht]{0.125\textwidth}
            \includegraphics[width=\linewidth]{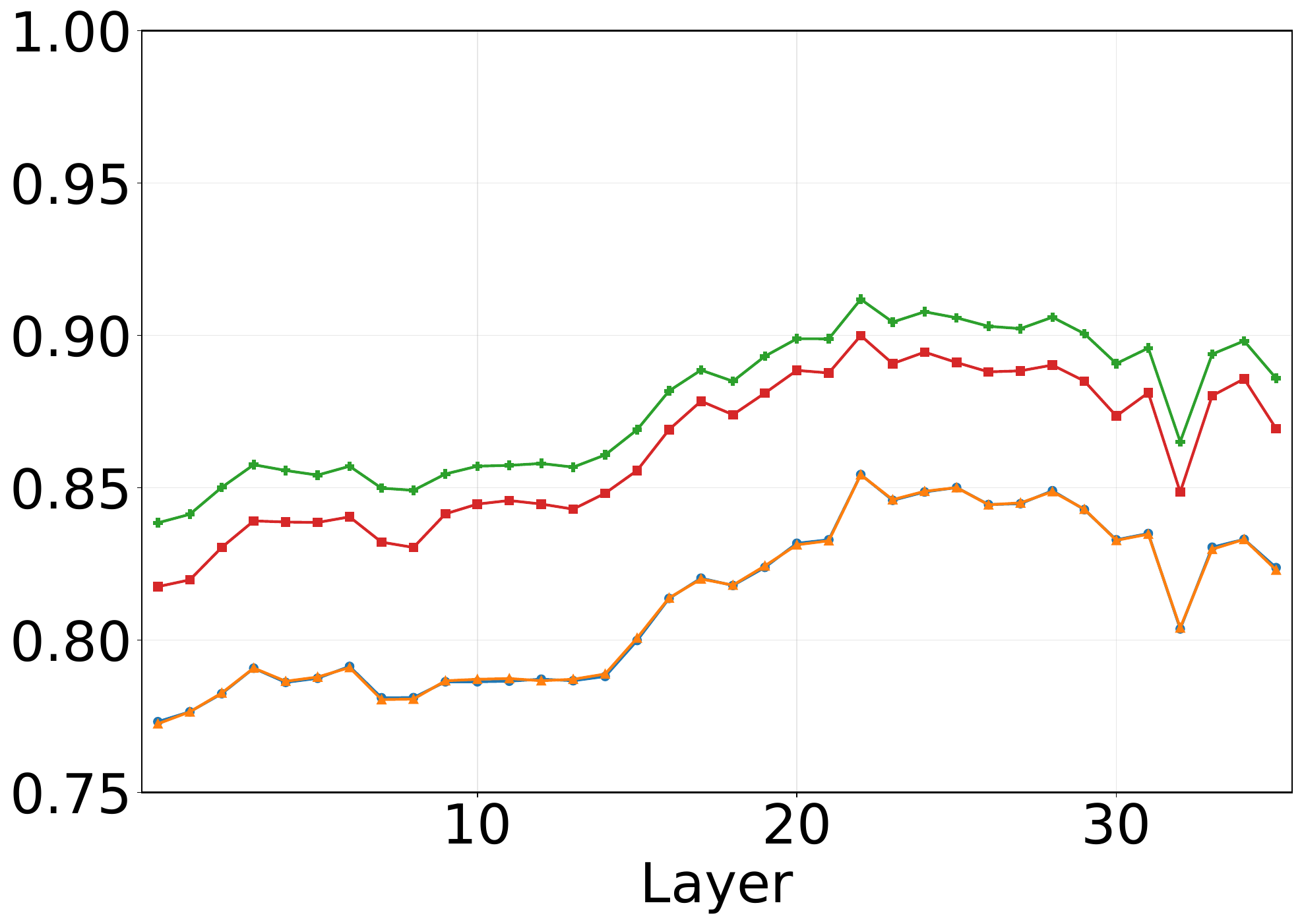}
            \caption{ORPO}
        \end{subfigure}\hfill
        \begin{subfigure}[!ht]{0.125\textwidth}
            \includegraphics[width=\linewidth]{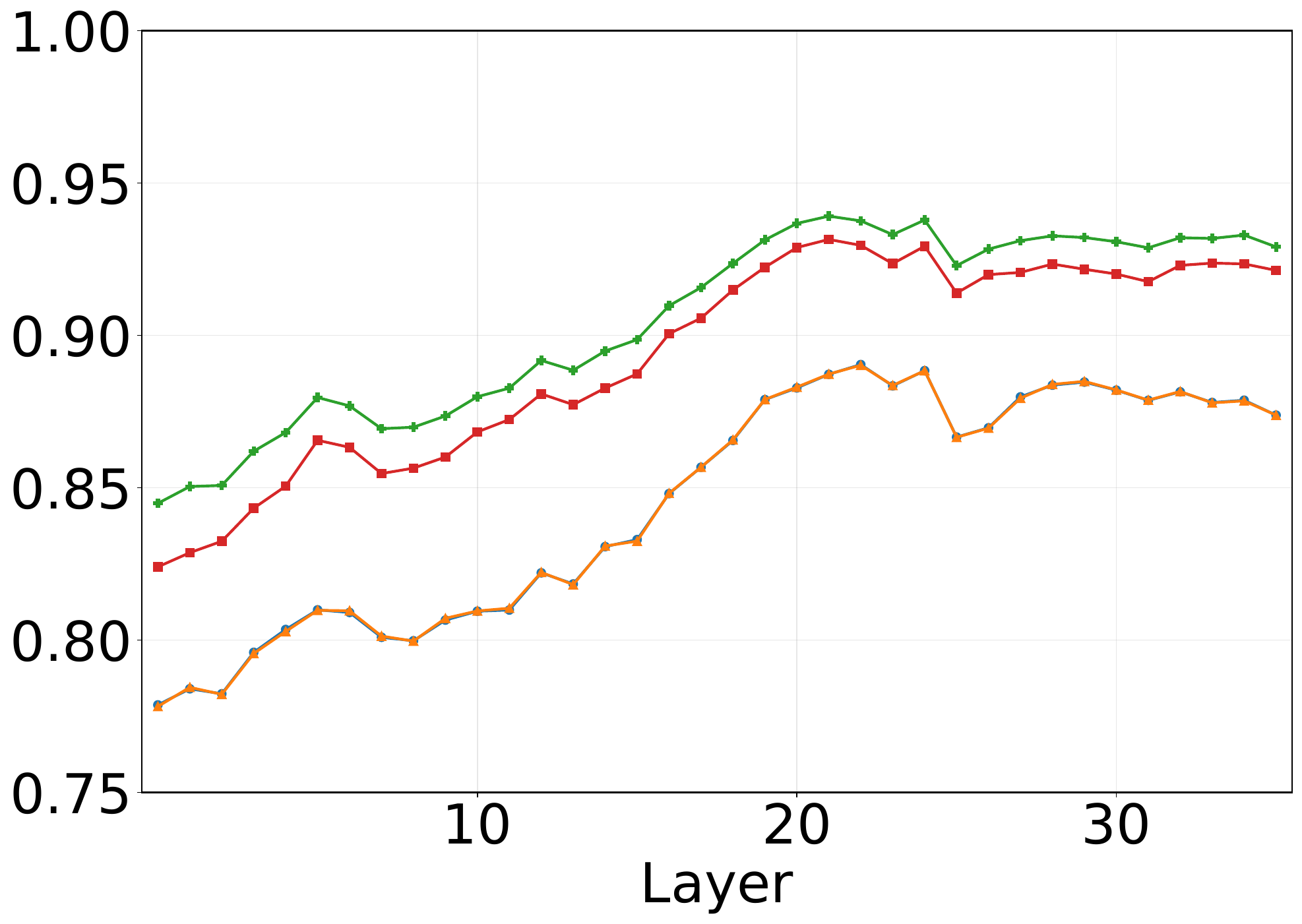}
            \caption{PPO}
        \end{subfigure}\hfill
        \begin{subfigure}[!ht]{0.125\textwidth}
            \includegraphics[width=\linewidth]{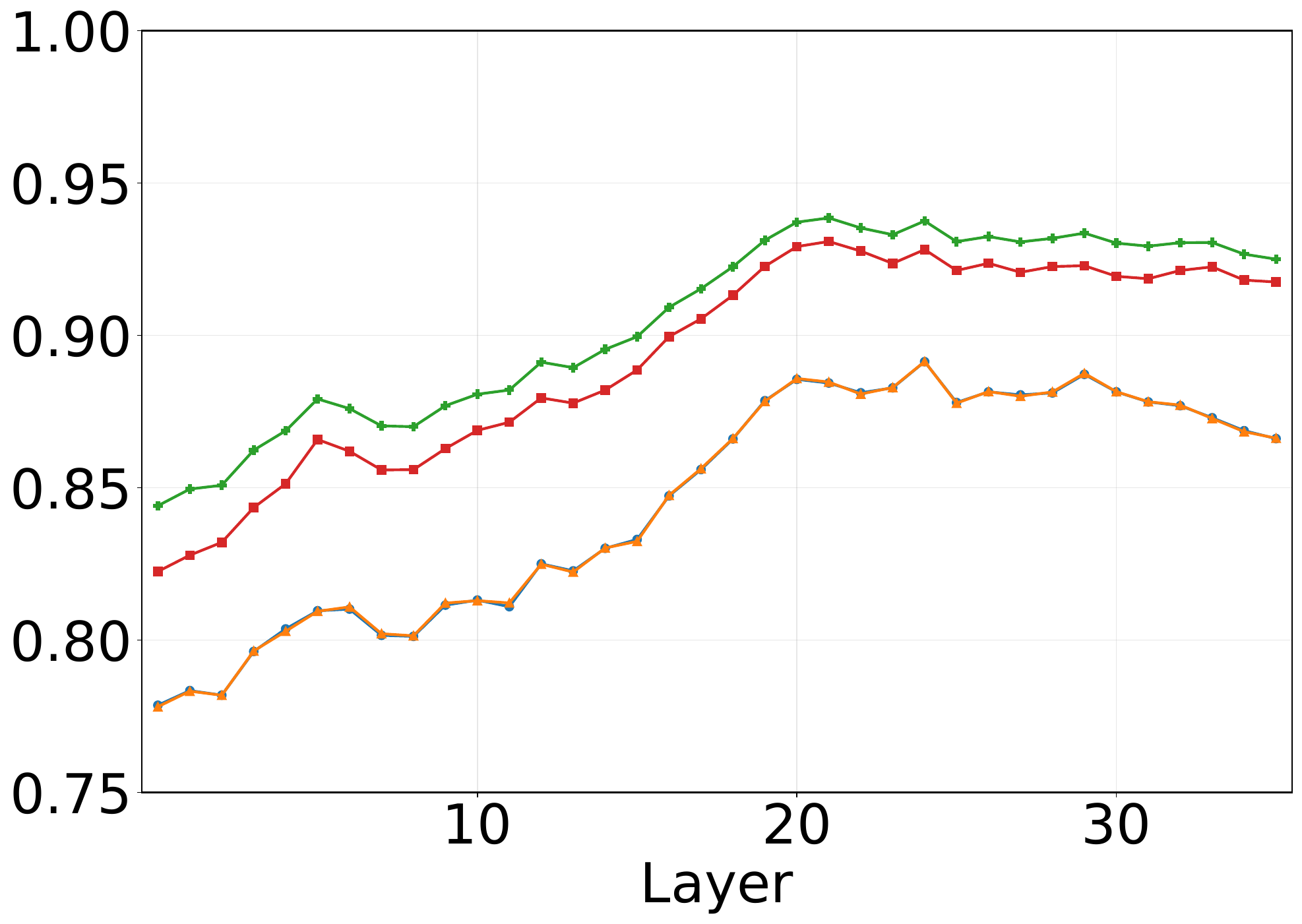}
            \caption{SimPO}
        \end{subfigure}

        \vspace{0.3em}
        \textbf{(iii) Qwen3-4B}
    \end{minipage}

    \caption{Layer-wise linear probe metrics across models: {\color{AccuracyColor}Accuracy}, {\color{FOneColor}F1 Score}, {\color{AUROCColor}AUROC}, and {\color{AUPRCColor}AUPRC}.}
    \label{fig:lp_layerwise_all_models}
\end{figure*}

\subsection{Linear Probing}
\label{sec:linear_probing}

We summarize the probing results in Figure \ref{fig:lp_layerwise_all_models} with exact values in Table \ref{tab:app_lp_full}. We observe that the objective functions of the different alignment algorithms affect the linear encoding of preferences. 

The classical \textit{camel-hump} pattern is evident across model architectures and alignment algorithms. Most of the discriminability characteristics concentrate in either the early--middle or middle--late layers. 

\textbf{PPO and SimPO} do not improve the linear separation of preferences when compared to the baseline.

\textbf{KTO}'s asymmetric utility weighting, with the loss-aversion parameter of $\lambda > 1$~\cite{ethayarajh2024kto}, penalizes undesirable outputs more strongly, thereby amplifying the gradients for rejected completions. This asymmetry leads to the \textbf{best} discriminative power without amplifying the relative coefficient norms ($\ell_2$-norm is less than half of the maximum at ``best'' layer, \cref{tab:app_lp_full} and \cref{fig:lp_layerwise_coef_norm_all_models} in Appendix$\S$\ref{app:linear_probes}). Next, \textbf{GRPO}'s advantage normalization within each generation group ($A_i = r_i - \bar{r}_G$) amplifies gradients in proportion to intra-group variance, with coefficient norms up to $47.98$ at the ``best'' layer (\cref{tab:app_lp_full} and \cref{fig:lp_layerwise_coef_norm_all_models} in Appendix$\S$\ref{app:linear_probes}) on Llama-3.2-3B. Yet, the overall performance is below KTO.

\textbf{DPO and ORPO} degrade overall linear separability for both Llama-3.2 and Qwen3 relative to the base model and most other alignment methods. Since DPO is an offline preference objective rather than RLHF, this can be attributed to non-constructive interfere with (linear) preference structure that is already present after instruction tuning. We present more analyses on the feature geometry in $\S$\ref{sec:crosscoder}.

\subsection{SAE Anchor Transfer Across Alignment Methods}
\label{sec:sae-results}

\begin{figure}[!htbp]
\centering
\includegraphics[width=0.8\linewidth]{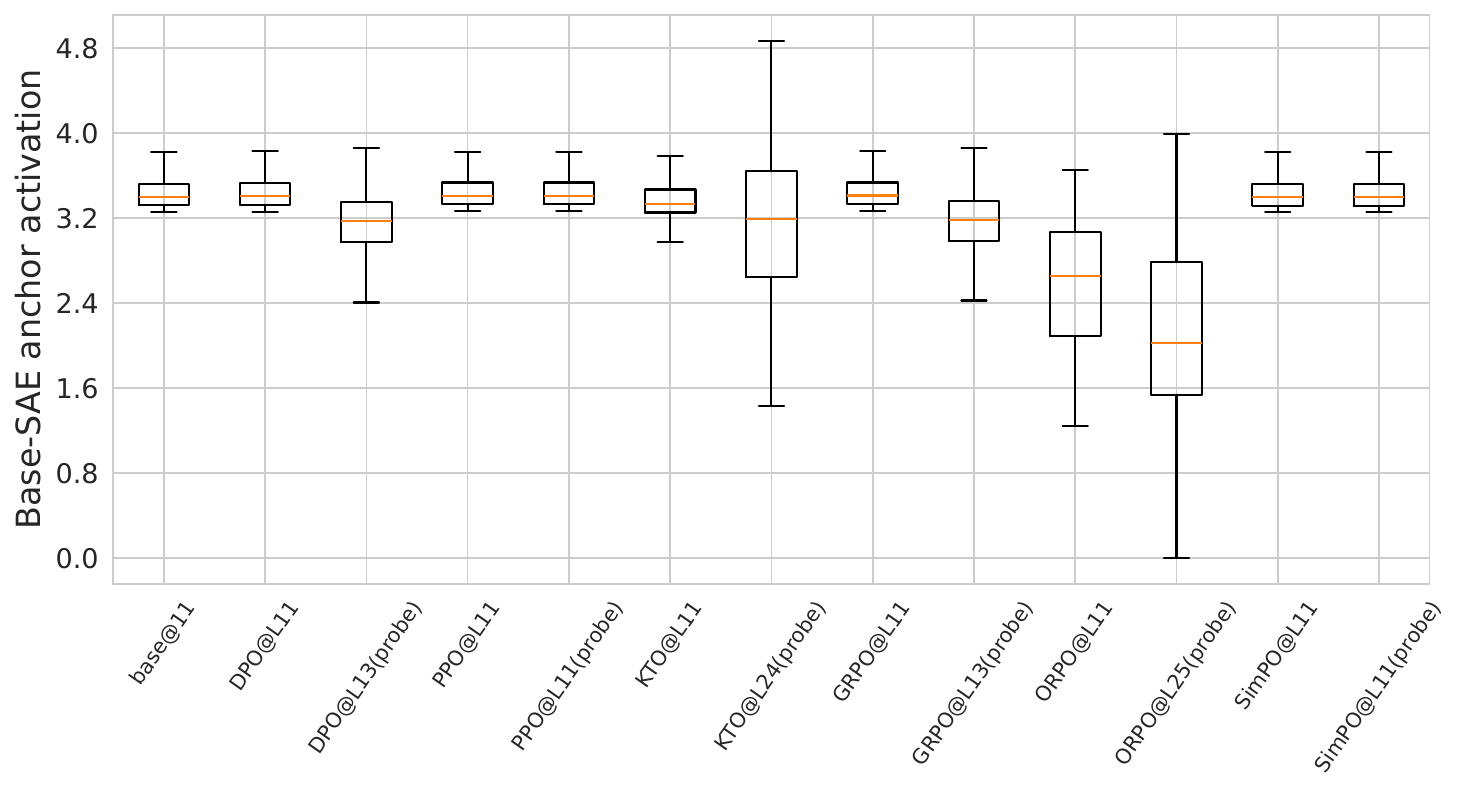}
\caption{Distribution of anchor feature activations for Llama-3.2-3B (Feature 15132) across alignment methods.}
\label{fig:anchor-llama}
\end{figure}

\begin{figure}[!htbp]
\centering
\includegraphics[width=0.8\textwidth]{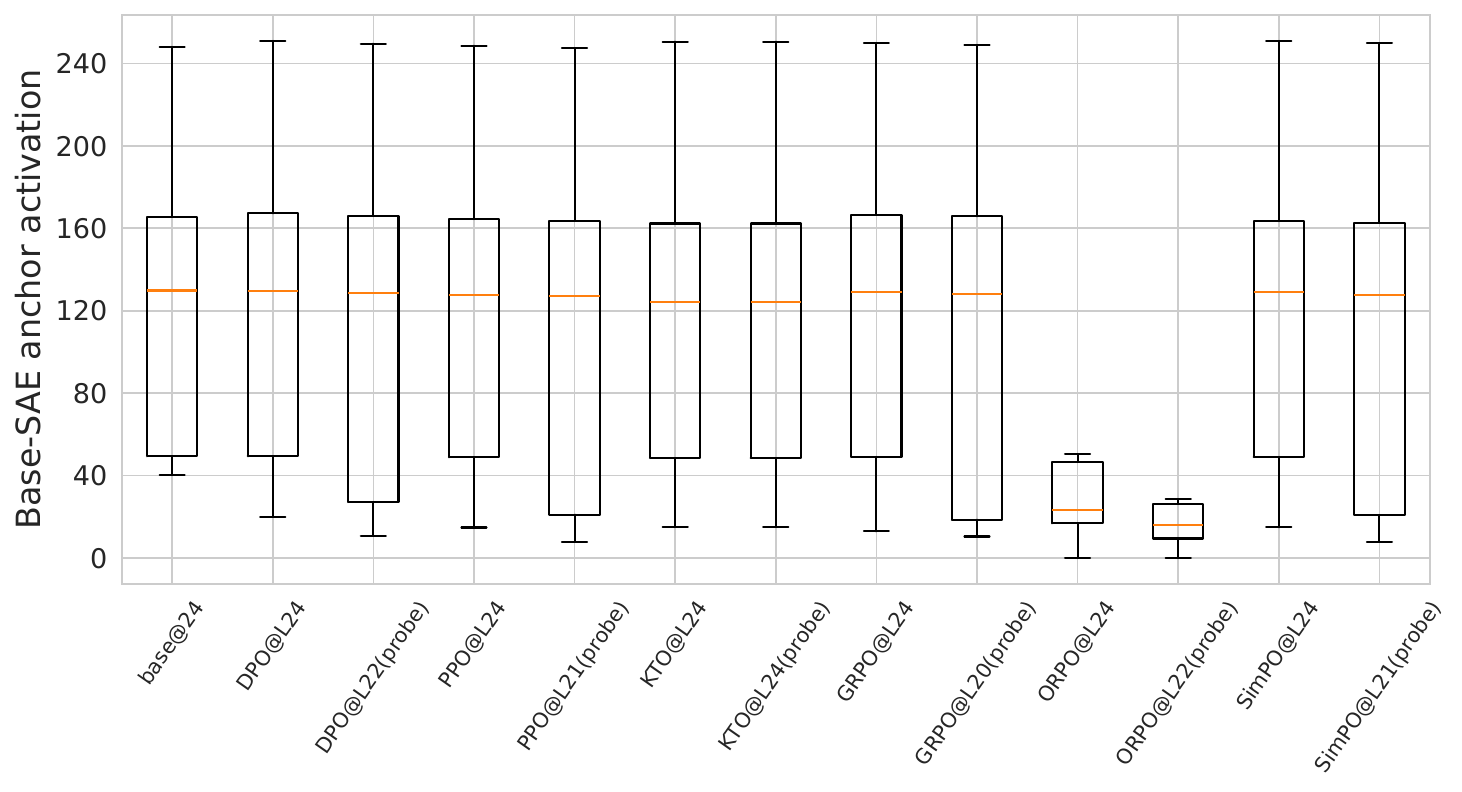}
\caption{Distribution of anchor feature activations for Qwen3-4B (Feature 6910) across alignment methods.}
\label{fig:anchor-qwen}
\end{figure}

\begin{figure}[!htbp]
\centering
\includegraphics[width=0.8\textwidth]{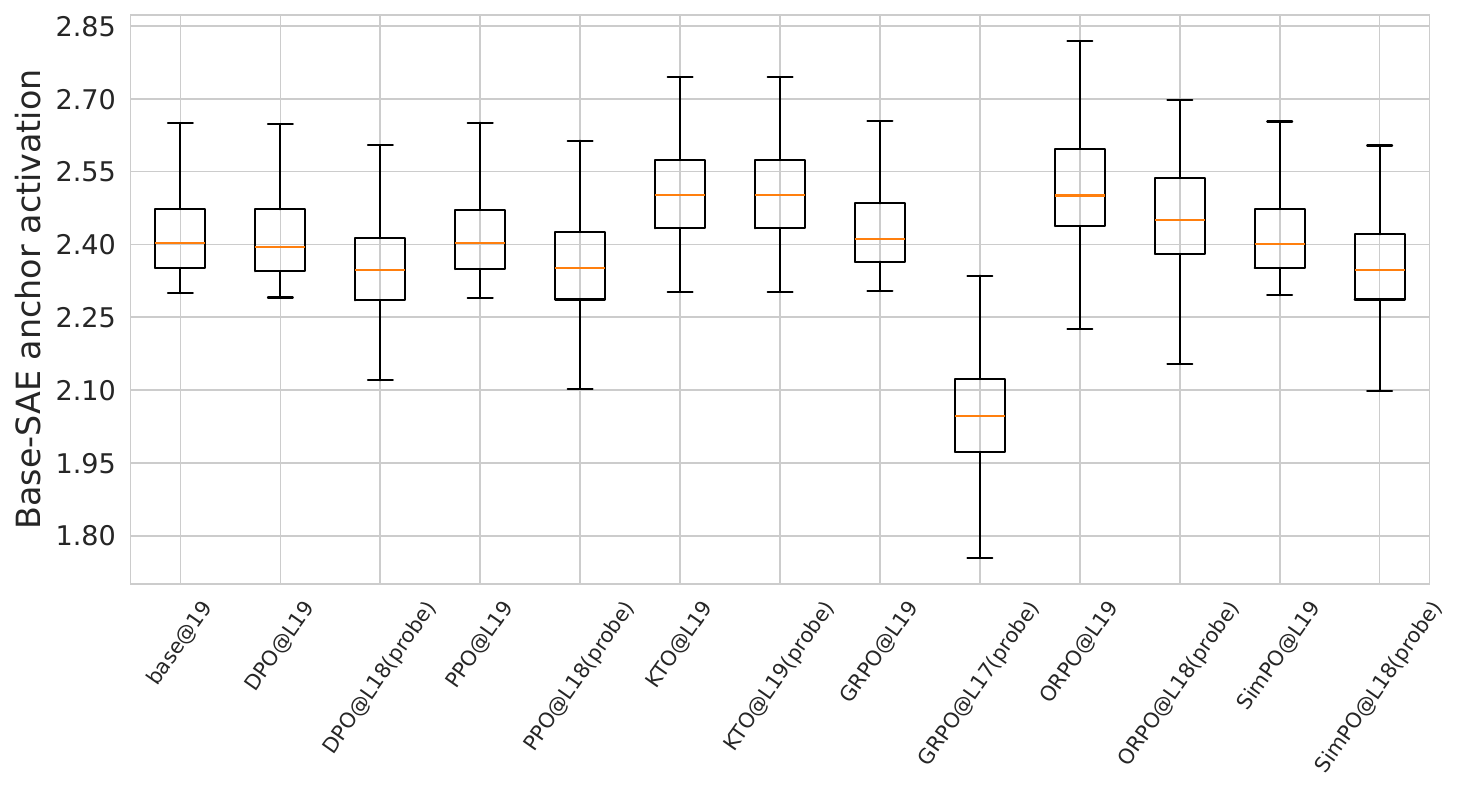}
\caption{Distribution of anchor feature activations for SmolLM3-3B (Feature 10154) across alignment methods.}
\label{fig:anchor-smollm}
\end{figure}

Figures~\ref{fig:anchor-llama}--\ref{fig:anchor-smollm} presents the activation distributions of the anchor feature for three model families. We also examine the maximally activated neurons to further probe whether the observed stability holds beyond average activations, with detailed results provided in the in Table \ref{tab:anchor_stability} in Appendix \ref{app:sae_metrics}.

\paragraph{ORPO suppression in Llama and Qwen.}
For Llama-3.2-3B (Feature 15132) and Qwen3-4B (Feature 6910), the ORPO method induces a marked reduction in anchor feature activation relative to the base model. In the Qwen family, this suppression is more pronounced, with median activations decreasing by 400\%. This suppression is consistent across both the SAE training layer and the linear probe best layer (e.g., ORPO, L22). This suggests that ORPO may aggressively attenuate certain representations that are active in the base model, potentially reflecting the method's direct optimization on generation odds.

\paragraph{Feature variance under KTO.}
The KTO method exhibits consistently higher variance in anchor feature activations across the model families. The boxplots for KTO show larger interquartile ranges and longer whiskers compared to other methods. This indicates that KTO introduces greater dependence on the context, their representations thereof, despite maintaining similar median activation levels to the base model.

\paragraph{Model-specific heterogeneity.}
The SmolLM3-3B results (Feature 10154) demonstrate that these effects are not uniform across architectures. In this family, ORPO maintains activation levels comparable to the base model. Instead, the GRPO method shows reduced activation specifically at the linear probe best layer (GRPO, L17), while activation at the training layer (GRPO, L19) remains elevated. Additionally, KTO results in a slight increase in median activation for SmolLM, contrasting with the preservation seen in other models. This heterogeneity implies that the impact of alignment algorithms on specific feature directions depends on the model architecture and the initial pretraining state.

\subsection{Crosscoder Feature Analysis} \label{sec:crosscoder}
\begin{figure*}[t!]
    \centering
    \includegraphics[width=\linewidth]{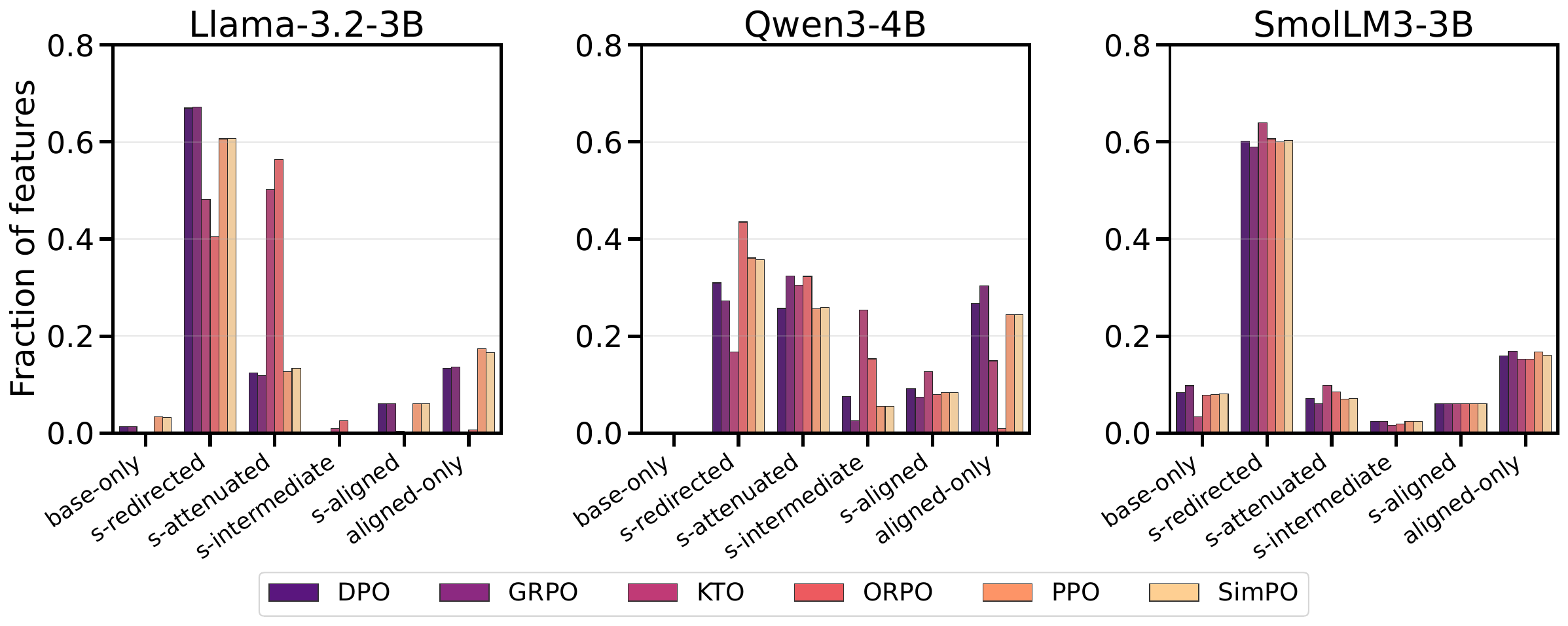}
    \caption{Crosscoder Feature Distribution.}
    \label{fig:crosscoder-main-feature-distribution}
\end{figure*}

We analyze how AFT modifies the shared latent space between base and aligned models using crosscoders trained at the probe-optimal layer (\cref{fig:crosscoder-main-feature-distribution}). All algorithms preserve a majority of base-model features, yet the geometric transformations differ substantially. Amongst the methods, KTO and ORPO share the most, yet we see a stark difference in their performance. Qwen3-4B's feature distribution is less peaky and has higher aligned-specific features, while SmolLM3-3B best preserves the original base features albeit high feature sharing.

\paragraph{KTO's feature sharing is constructive.} Post KTO, the feature distribution substantiates that the geometric transformation induced from alignment has benefited the (linear) encoding of the preference representations. The asymmetric preference optimization process modifies the features such that through sharing, the existing (linear) discriminability is not only borrowed (high feature sharing), but also further improved (\cref{fig:lp_layerwise_all_models}, column (d)) through the newer aligned-only features. 

\paragraph{DPO and ORPO degrade understanding of preferences, but through different ways.} In either case, feature sharing is high just like KTO. Yet, the higher rotation instead leads to non-constructive distortion of the original geometry causing degraded separability. This leads to the new representations occupying subspaces orthogonal to the original preference-encoding subspaces. On the other hand, after ORPO, many of the shared features are attenuated, dampening the original preference-relevant activations. This dampening affect is also more clearly visible in \cref{fig:anchor-llama}--\cref{fig:anchor-smollm}, where the activation of specific features (e.g., \texttt{f6910} in Qwen3-4B) are significantly smaller. Together, these results imply that alignment objectives that directly optimize pairwise log-probability differences as a standalone classification signal, rather than using them as part of iterative policy improvement as in PPO-style methods (e.g., SimPO, GRPO), hinder the overall (linear) encoding of preferences. This also substantiates the discussions in $\S$\ref{sec:linear_probing} and \ref{sec:sae-results}.

\paragraph{PPO and SimPO show similar behavior.} Both PPO and SimPO utilize the log-probability differences for iteratively improving the policies. Their subtle differences in the alignment objectives do not hinder the preference-encoding mechanisms. This is geometrically reflected in their feature distribution too, reifying the probing results in $\S$\ref{sec:linear_probing}.

\section{Conclusion}
This study provides the first comprehensive mechanistic comparison of six leading preference optimization algorithms across three model architectures, showing that alignment is not a uniform behavioral intervention but a set of qualitatively distinct representational transformations. Using linear probing, Sparse Autoencoders, and crosscoder analysis, we find that different objectives reshape internal feature geometry in systematically different ways. In particular, KTO and GRPO tend to enhance the linear decodability of preference representations through \textbf{constructive feature sharing} and \textbf{sparse recruitment of high-salience features}, whereas DPO and ORPO often reduce separability through non-constructive \textit{geometric distortion}, including rotation and feature attenuation. By contrast, PPO and SimPO largely \textbf{preserve baseline geometry}. Notably, these effects are somewhat architecture-dependent, indicating that alignment outcomes cannot be fully understood without accounting for model initialization and inductive biases. Taken together, our results provide a more white-box analysis of alignment fine-tuning moving the field beyond output-level benchmarking. We further motivate the readers to utilize such analyses for designing \textbf{targeted alignment objectives}, e.g., safety-training, that also improve the geometric fidelity for better transparency and verifiability of the post-training protocols.

\section{Limitations}

While this work provides a systematic mechanistic comparison of six preference-optimization algorithms across three model families, we acknowledge the following limitations. Our experiments are constrained to 3B--4B parameter models. Alignment-induced representational changes, particularly feature superposition density and feature redundancy, may scale non-linearly; thus, the geometric patterns observed here may not fully generalize to 70B+ production models or architectures with fundamentally different inductive biases. Second, our analysis is diagnostic rather than causal at the component level. Linear probes identify decodable preference information, SAEs decompose activations into sparse features, and crosscoders compare feature geometry across base and aligned models. We do not perform exact causal ablations, activation patching, path patching, or attention-head/MLP-level interventions to establish which components are necessary or sufficient for the observed behavior. As a result, our claims should be read as feature-level evidence about representational change, not as complete causal circuit identification.

\section*{Acknowledgment}

The authors acknowledge Llambda and CloudRift for providing compute credits. The authors also acknowledge the Safety and Alignment Research India organization for providing the platform to conduct this research.

\bibliography{references}
\bibliographystyle{plain}


\appendix
\clearpage

\section*{Appendix}

\noindent\begin{minipage}{\linewidth}
\begin{tcolorbox}[
    colback=white,
    colframe=blue,
    title={\textbf{Table of Contents}},
    fonttitle=\bfseries,
    boxrule=0.8pt,
    left=6pt, right=6pt, top=4pt, bottom=4pt
]
\begin{description}[leftmargin=1.5em, labelwidth=1.2em, labelsep=0.3em]
    \item[A] Impact Statement \dotfill \pageref{app:hyperparams}
    \item[B] Hyperparameters for Alignment Fine Tuning \dotfill \pageref{app:hyperparams}
    \item[C] Linear Probes \dotfill \pageref{app:linear_probes}
    \item[D] Sparse Autoencoders \dotfill \pageref{app:sae_metrics}
    \item[E] Crosscoder Details \dotfill \pageref{app:crosscoder-details}
    \item[F] LLM Usage \dotfill \pageref{app:llm_usage}
\end{description}
\end{tcolorbox}
\end{minipage}
\vspace{1em}

\section{Impact Statement}\label{app:impact}

This work addresses a critical transparency gap in the post-training of large language models by moving alignment evaluation beyond opaque behavioral benchmarks toward feature-level diagnostics. Using linear probing, sparse autoencoders, and crosscoder analysis, we show that commonly used preference-optimization methods can induce qualitatively different internal changes: DPO and ORPO may reduce the linear separability of preference representations through geometric distortion or feature attenuation, whereas KTO better preserves or enhances internal discriminability, making the resulting models more amenable to mechanistic interpretability and safety auditing. These findings have important implications for AI safety, since a model can appear behaviorally aligned while internally obscuring preference structure, potentially masking vulnerabilities to adversarial prompts, distribution shifts, or hidden capability degradation. At the same time, this mechanistic transparency introduces dual-use considerations, as detailed knowledge of preference-feature structure could be misused to circumvent safeguards or reverse-engineer post-training pipelines. For this reason, our results motivate mechanism-aware alignment objectives, standardized internal auditing protocols, and controlled disclosure practices that support trustworthy deployment while preserving public oversight and scientific progress.

\section{Hyperparameters for Alignment Fine Tuning} 
\label{app:hyperparams}
In this section, we detail the exact hyperparameters utilized across all alignment scripts, linear probes, Sparse Autoencoders (SAEs), and Crosscoders.

\paragraph{LoRA Configuration (All Alignment Methods)} 
Rank $r=16$, $\alpha=32$, dropout $0.05$, targeting \texttt{q\_proj, k\_proj, v\_proj, o\_proj}.

\paragraph{DPO} 
Learning rate $5 \times 10^{-7}$, batch size $4$, gradient accumulation steps $8$, $1$ epoch, max length $512$, AdamW optimizer.

\paragraph{PPO} 
Learning rate $3 \times 10^{-6}$, batch size $2$, gradient accumulation steps $4$, $1$ epoch, max prompt tokens $256$, response length $256$, KL coefficient $0.05$ (k1 estimator), PPO epochs $1$, mini-batches $1$, temperature $0.7$, AdamW optimizer.

\paragraph{GRPO} 
Learning rate $1 \times 10^{-6}$, batch size $4$, gradient accumulation steps $8$, $1$ epoch, max prompt tokens $512$, max completion length $256$, $4$ generations per prompt, chosen weight $1.0$, rejected weight $0.25$, length penalty start $1.35$, length penalty scale $0.05$, $\beta=0.0$, temperature $0.8$, top-$p$ $0.9$, 8-bit AdamW optimizer.

\paragraph{SimPO} 
Learning rate $5 \times 10^{-7}$, batch size $4$, gradient accumulation steps $8$, $1$ epoch, max length $1024$, max prompt length $128$, $\beta=2.0$, $\gamma=1.0$, CPO $\alpha=0.0$, warmup ratio $0.03$, max gradient norm $1.0$, AdamW optimizer.

\paragraph{ORPO}
Learning rate $5 \times 10^{-6}$, batch size $16$, gradient accumulation steps $2$, $1$ epoch, max length $512$, max prompt length $256$, $\beta=0.1$, bf16 precision, AdamW optimizer.

\paragraph{KTO} 
Learning rate $5 \times 10^{-7}$, batch size $4$, gradient accumulation steps $8$, $1$ epoch, max length $512$, $\beta=0.1$, desirable weight $1.0$, undesirable weight $1.0$, AdamW optimizer.

\section{Linear Probes}
\label{app:linear_probes}

\subsection{Methodology}
Linear probing is used as a targeted diagnostic to assess the degree to which a given alignment algorithm induces linear separability in the model's residual-stream  representations. For each (base model, algorithm) configuration, hidden-state activations are extracted at every layer using $10{,}000$ chosen and rejected response pairs sampled from the corresponding UltraFeedback data split. Activations are pooled at the final prompt token, with a maximum context length of $1024$ tokens. A logistic regression classifier with balanced class weights is trained via 5-fold cross-validation and Adam optimizer (learning rate $0.05$, $1500$ iterations) to predict the preference label from the hidden states at each layer. The \textbf{peak probe layer} is the layer achieving the highest held-out accuracy.


\subsection{Summary of Best-Layer Results}

The best-layer results show that ORPO produces weaker preference separability than the strongest alignment methods, especially on Llama-3.2-3B and Qwen3-4B. On Llama-3.2-3B, ORPO falls below the base model across accuracy, F1, AUROC, and AUPRC; on Qwen3-4B, it similarly remains below the base model. This degraded separability motivates the feature-geometry analysis in the main text, where ORPO is associated with attenuation of shared preference-relevant features rather than improved alignment of the original preference direction.

\begin{table}[!ht]
\centering
\caption{%
    Linear probe results at the peak layer for each (base model, algorithm) pair.
    Accuracy and F1 are macro-averaged over 5-fold cross-validation. AUROC and AUPRC
    are computed on held-out probability estimates.
    Coeff.~norm is the $\ell_2$ norm of the probe weight vector at the peak layer.
    Bold entries denote the best value within each base-model family.%
}
\label{tab:app_lp_full}
\small
\begin{tabular}{llrrrrrr}
\toprule
Base & Algorithm & Layer & Acc & F1 & AUROC & AUPRC & Coeff.~norm \\
\midrule
SmolLM3-3B & Baseline & 19 & 0.766 & 0.771 & 0.851 & 0.848 &  5.29 \\
SmolLM3-3B & DPO      & 18 & 0.767 & 0.768 & 0.853 & 0.853 &  5.09 \\
SmolLM3-3B & GRPO     & 19 & 0.818 & 0.820 & 0.904 & 0.904 & 10.77 \\
SmolLM3-3B & KTO      & 22 & {0.922} & {0.921} & {0.971} & {0.962} & 14.61 \\
SmolLM3-3B & ORPO     & 18 & 0.769 & 0.768 & 0.855 & 0.854 &  5.63 \\
SmolLM3-3B & PPO      & 18 & 0.766 & 0.767 & 0.852 & 0.851 &  5.10 \\
SmolLM3-3B & SimPO    & 18 & 0.766 & 0.768 & 0.852 & 0.852 &  5.11 \\
\midrule
Llama-3.2-3B & Baseline & 11 & 0.844 & 0.843 & 0.906 & 0.896 & 24.86 \\
Llama-3.2-3B & DPO      & 13 & 0.803 & 0.802 & 0.876 & 0.870 & 14.32 \\
Llama-3.2-3B & GRPO     & 13 & 0.946 & 0.946 & 0.959 & 0.948 & {47.98} \\
Llama-3.2-3B & KTO      & 24 & {0.958} & {0.958} & {0.980} & {0.975} & 14.46 \\
Llama-3.2-3B & ORPO     & 25 & 0.786 & 0.787 & 0.860 & 0.849 & 13.67 \\
Llama-3.2-3B & PPO      & 11 & 0.847 & 0.846 & 0.907 & 0.897 & 24.77 \\
Llama-3.2-3B & SimPO    & 11 & 0.846 & 0.845 & 0.907 & 0.896 & 24.75 \\
\midrule
Qwen3-4B & Baseline & 24 & 0.890 & 0.890 & 0.938 & 0.929 & 25.04 \\
Qwen3-4B & DPO      & 20 & 0.847 & 0.848 & 0.908 & 0.898 & 15.51 \\
Qwen3-4B & GRPO     & 20 & 0.948 & 0.948 & 0.967 & 0.959 & 43.18 \\
Qwen3-4B & KTO      & 25 & {0.970} & {0.970} & {0.988} & {0.984} & 14.22 \\
Qwen3-4B & ORPO     & 22 & 0.854 & 0.854 & 0.912 & 0.900 & 19.08 \\
Qwen3-4B & PPO      & 22 & 0.890 & 0.890 & 0.938 & 0.930 & 24.29 \\
Qwen3-4B & SimPO    & 24 & 0.891 & 0.892 & 0.938 & 0.928 & 25.16 \\
\bottomrule
\end{tabular}
\end{table}

\subsection{Combined Best-Layer ROC Curves}
\begin{figure}[!h]
    \centering
    \begin{subfigure}[b]{0.32\linewidth}
        \includegraphics[width=\linewidth]{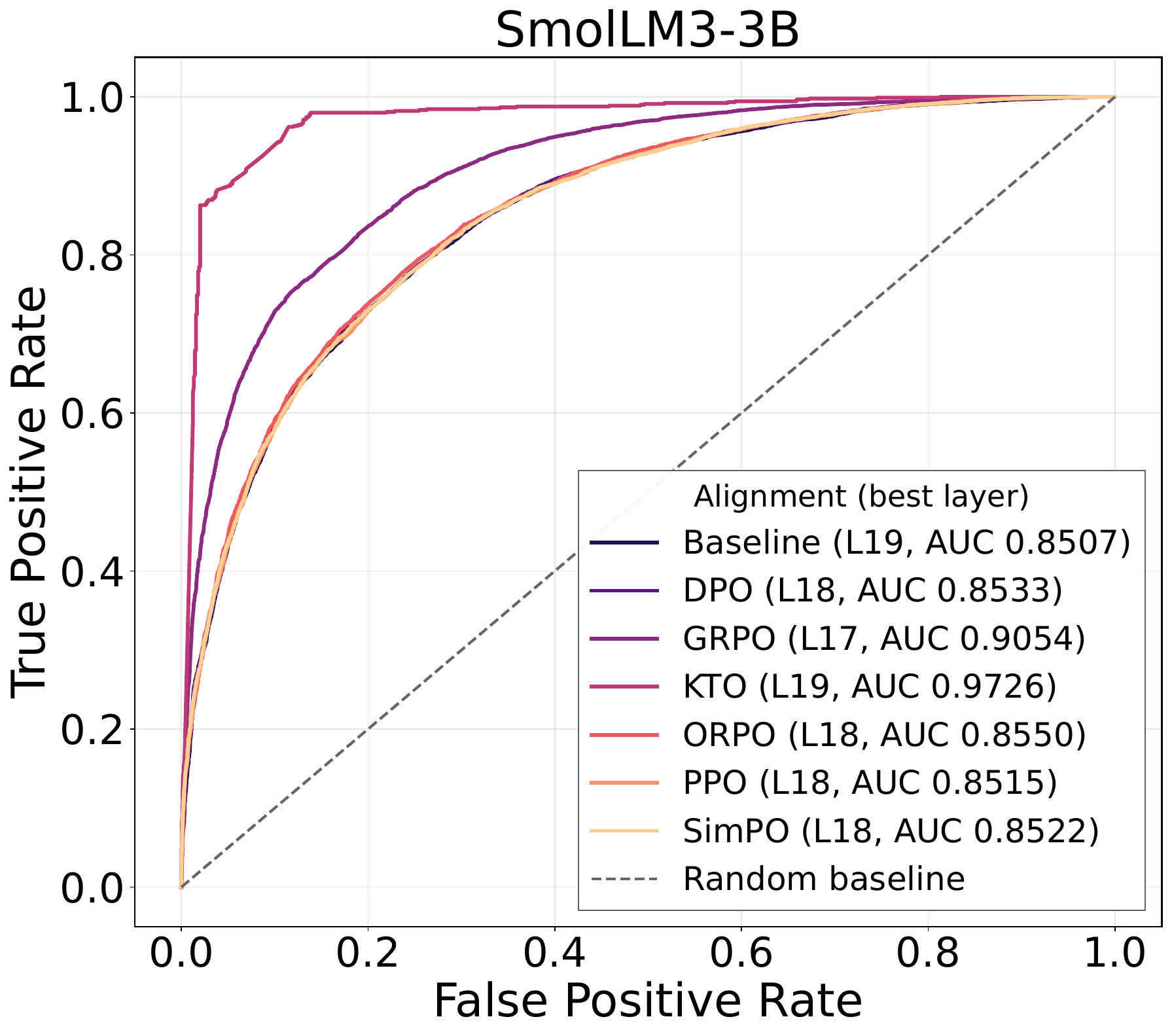}
        \caption{SmolLM3-3B}
    \end{subfigure}
    \hfill
    \begin{subfigure}[b]{0.32\linewidth}
        \includegraphics[width=\linewidth]{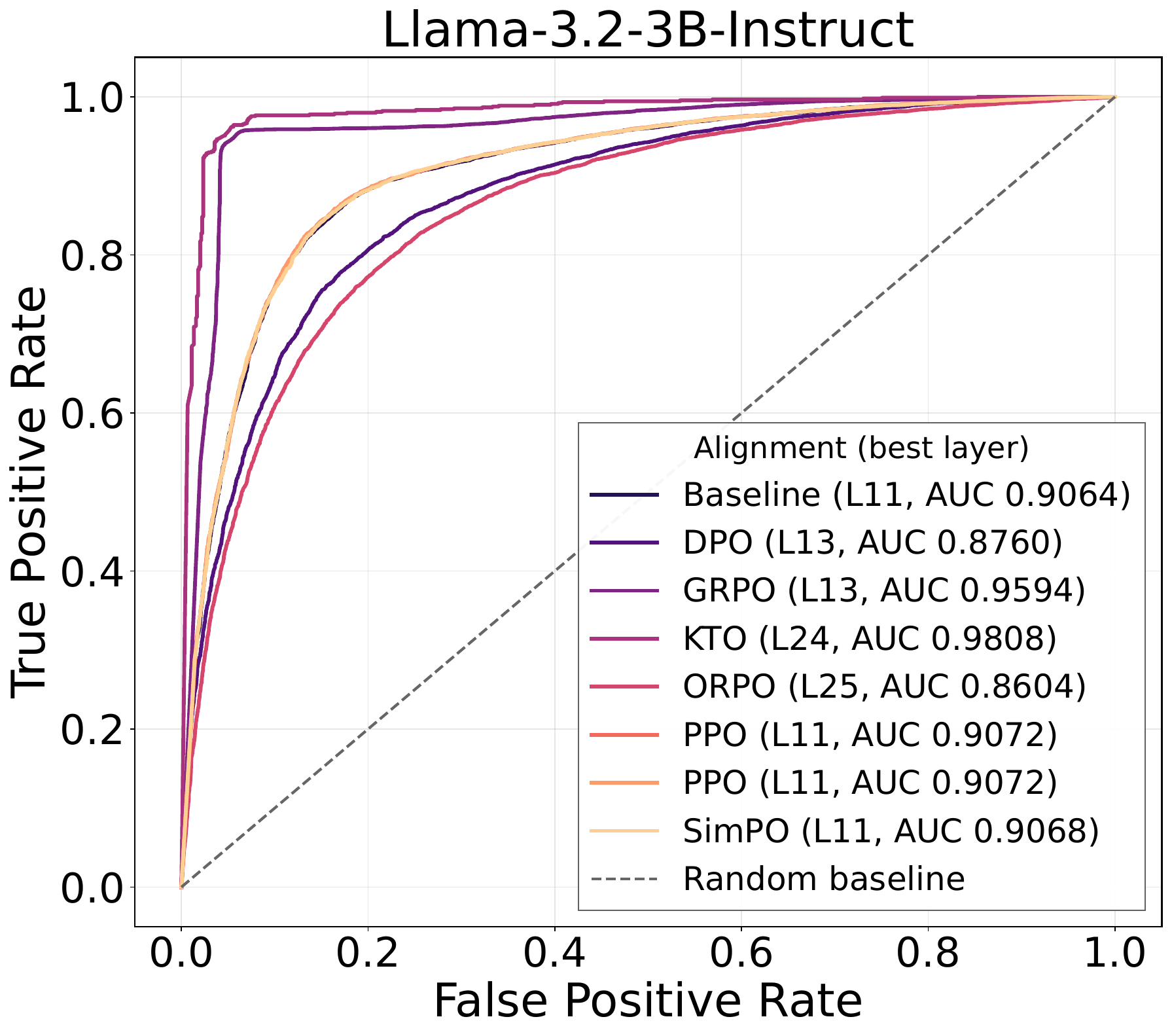}
        \caption{Llama-3.2-3B}
    \end{subfigure}
    \hfill
    \begin{subfigure}[b]{0.32\linewidth}
        \includegraphics[width=\linewidth]{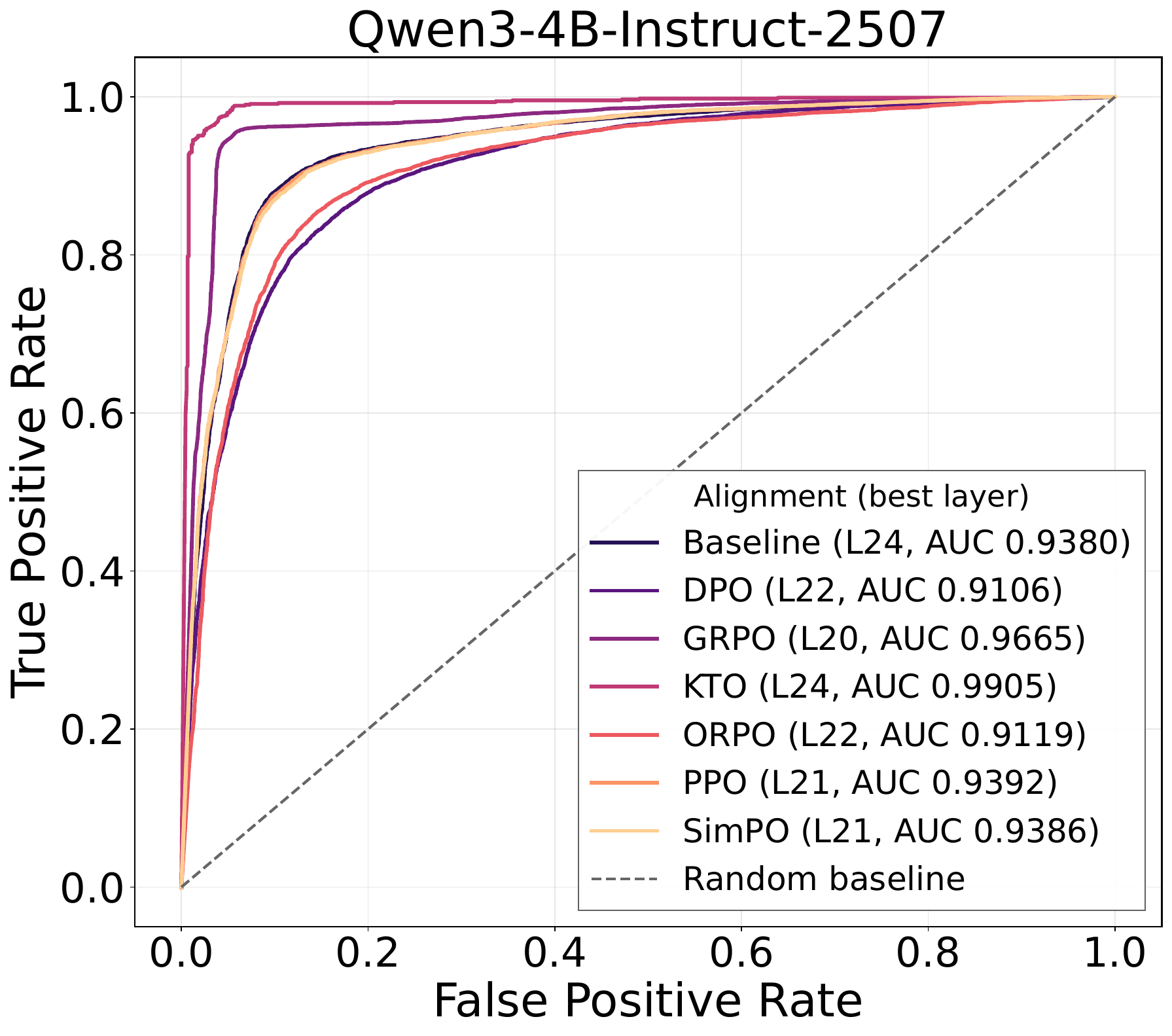}
        \caption{Qwen3-4B}
    \end{subfigure}
    \caption{%
        Combined Receiver Operating Characteristic (ROC) curves at the peak probe layer, one panel per base model. KTO consistently achieves the highest AUROC within each family, while ORPO exhibits the lowest.%
    }
    \label{fig:lp_combined_roc}
\end{figure}

\subsection{Per-Model Diagnostic Figures}
\label{subsec-appendix:linear-probe=diagnostics}
We firstly present the layer-wise $\ell_2$ co-efficient norm across each method. This helps to identify the relative prominence of the preference-specific separability within each AFT method.

\begin{figure*}[ht!]
    \centering
    \scriptsize

    \begin{minipage}{\textwidth}
        \centering
        \setcounter{subfigure}{0}
        \renewcommand{\thesubfigure}{\alph{subfigure}}
        \begin{subfigure}[!ht]{0.125\textwidth}
            \includegraphics[width=\linewidth]{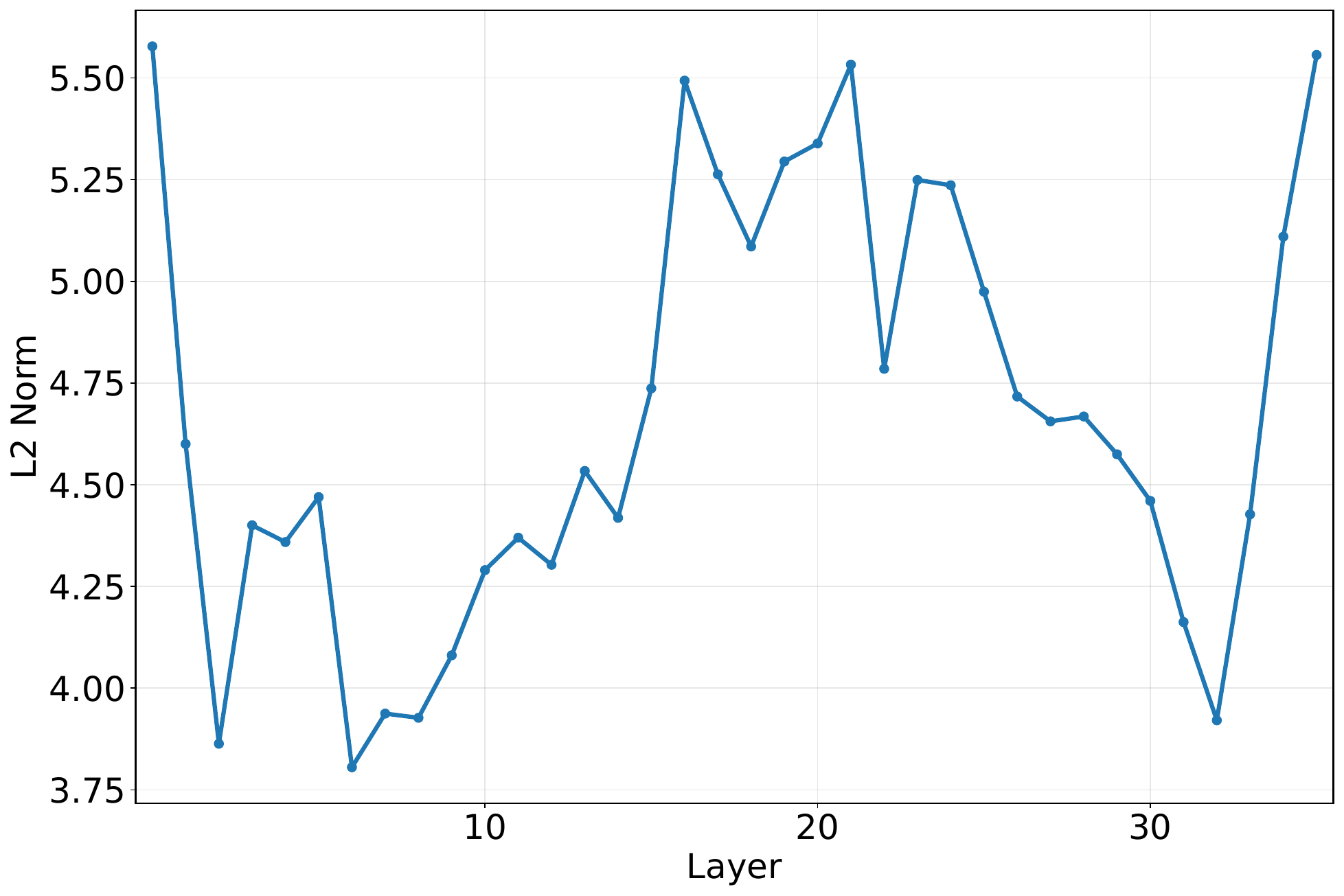}
            \caption{Base}
        \end{subfigure}\hfill
        \begin{subfigure}[!ht]{0.125\textwidth}
            \includegraphics[width=\linewidth]{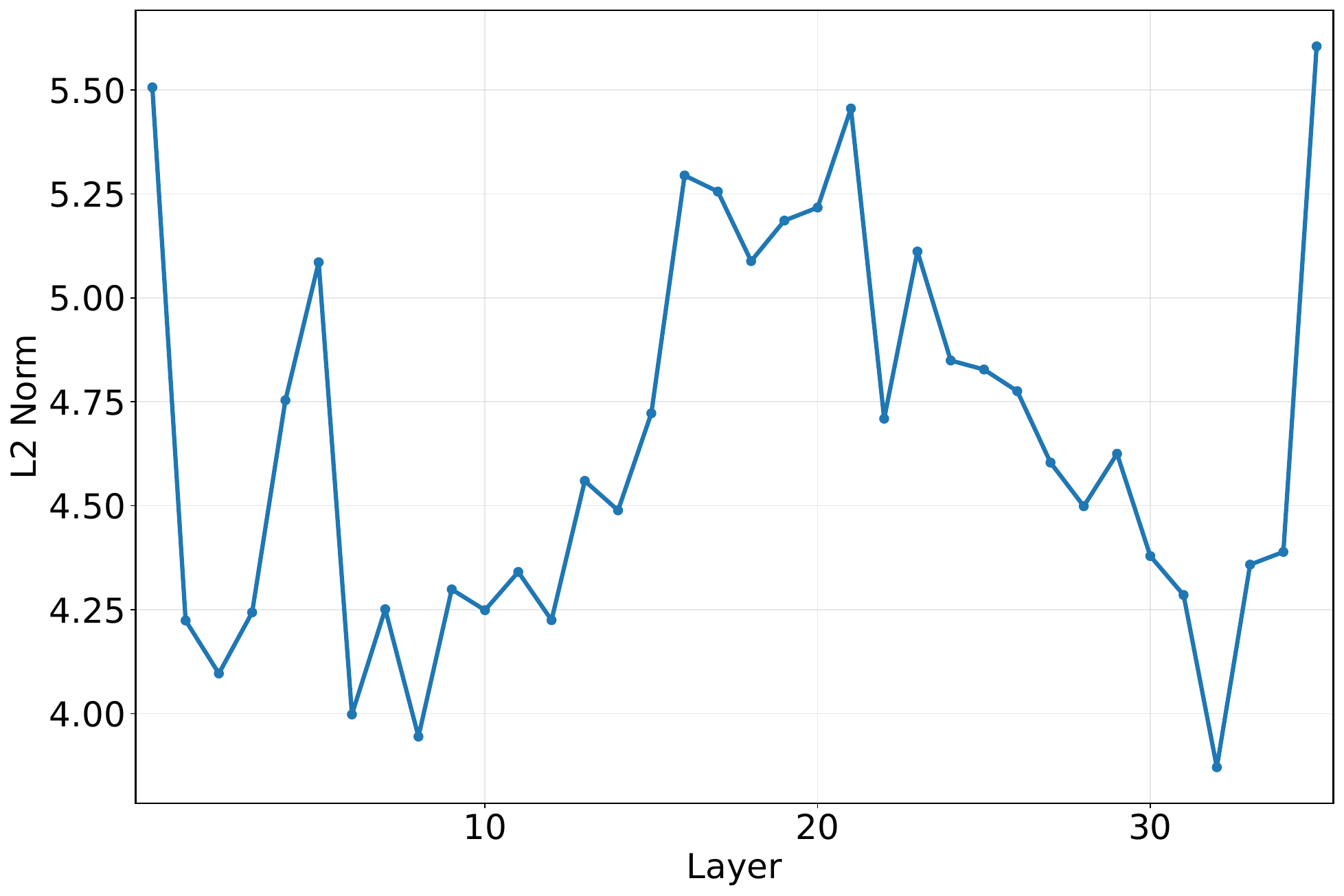}
            \caption{DPO}
        \end{subfigure}\hfill
        \begin{subfigure}[!ht]{0.125\textwidth}
            \includegraphics[width=\linewidth]{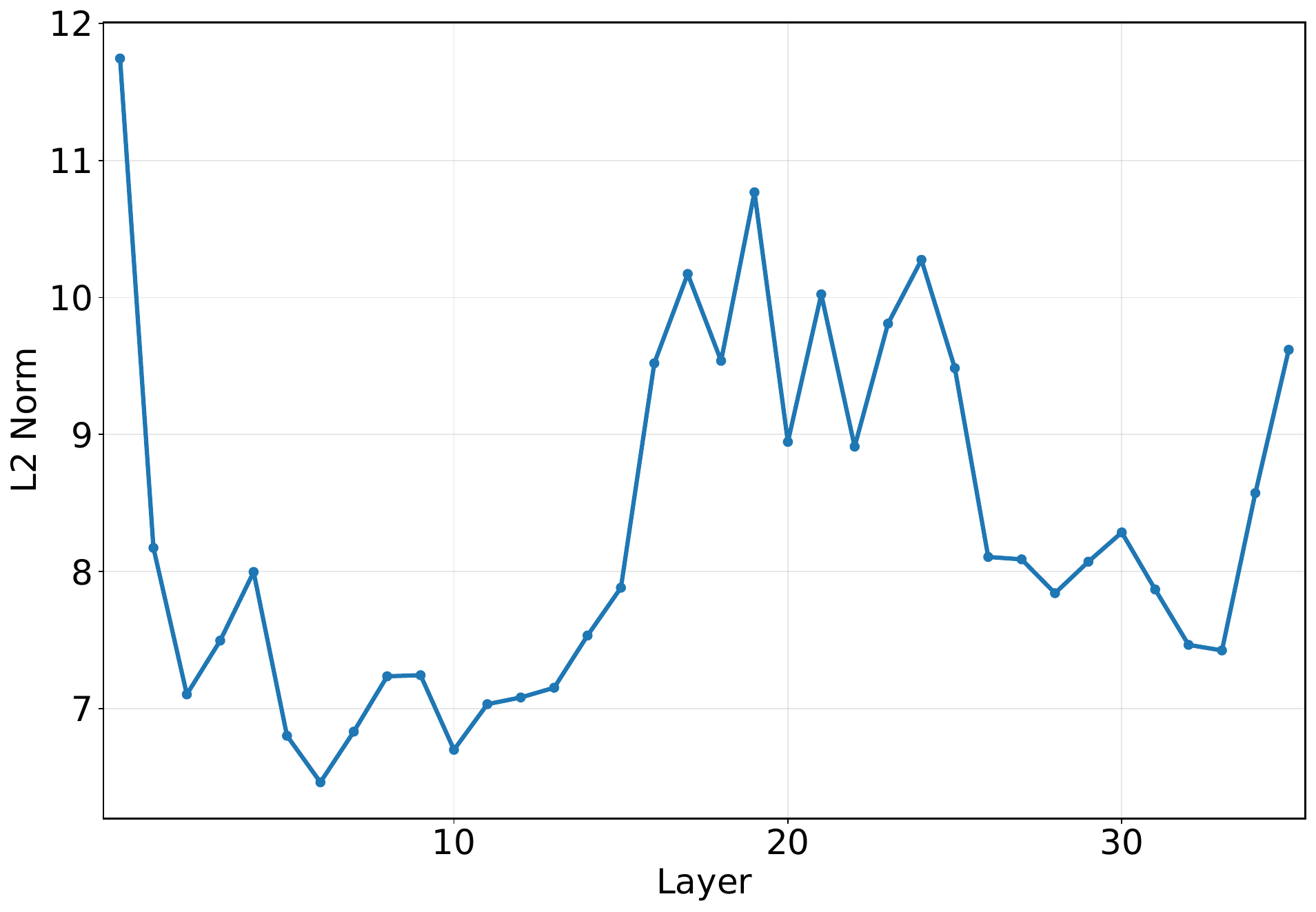}
            \caption{GRPO}
        \end{subfigure}\hfill
        \begin{subfigure}[!ht]{0.125\textwidth}
            \includegraphics[width=\linewidth]{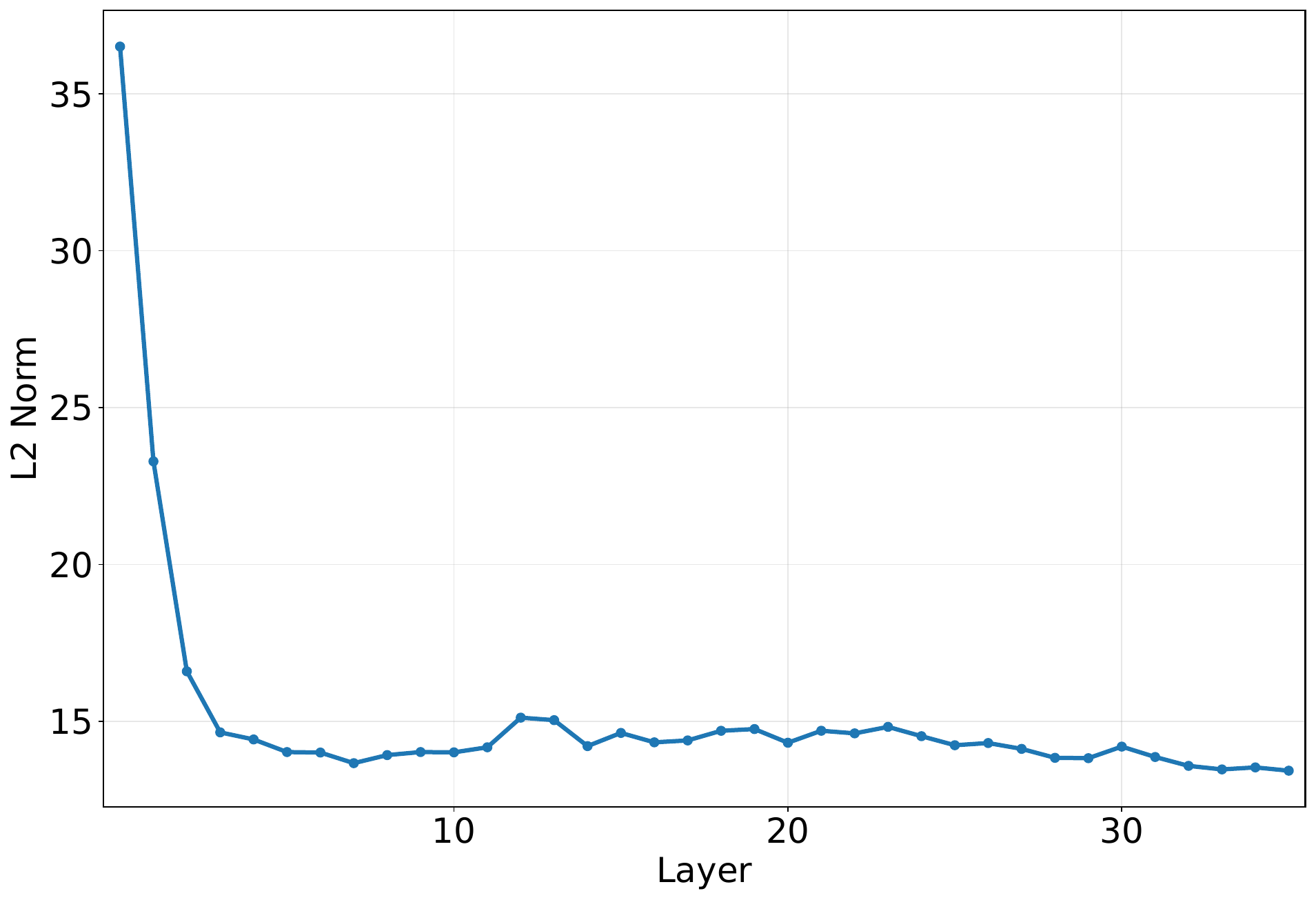}
            \caption{KTO}
        \end{subfigure}\hfill
        \begin{subfigure}[!ht]{0.125\textwidth}
            \includegraphics[width=\linewidth]{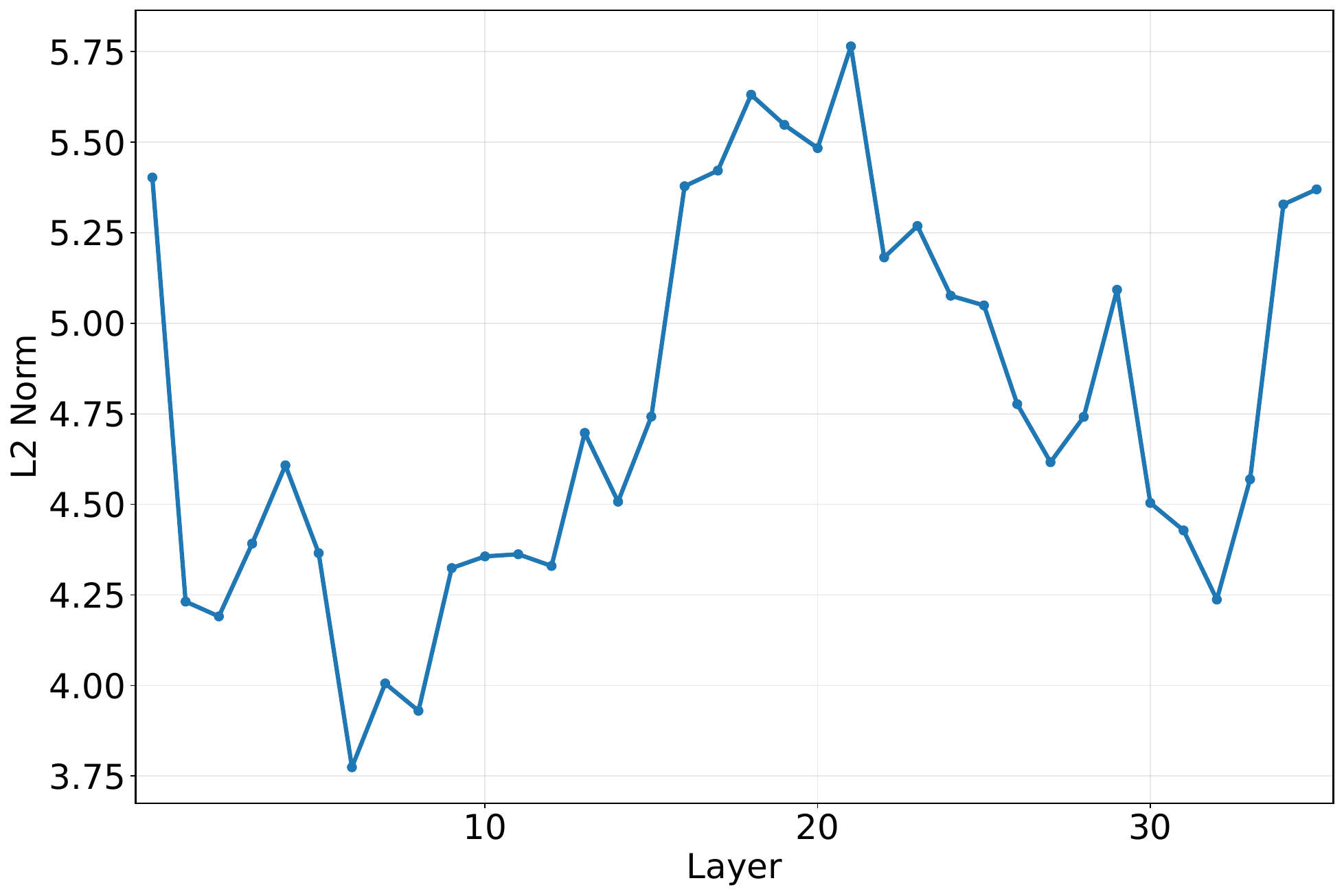}
            \caption{ORPO}
        \end{subfigure}\hfill
        \begin{subfigure}[!ht]{0.125\textwidth}
            \includegraphics[width=\linewidth]{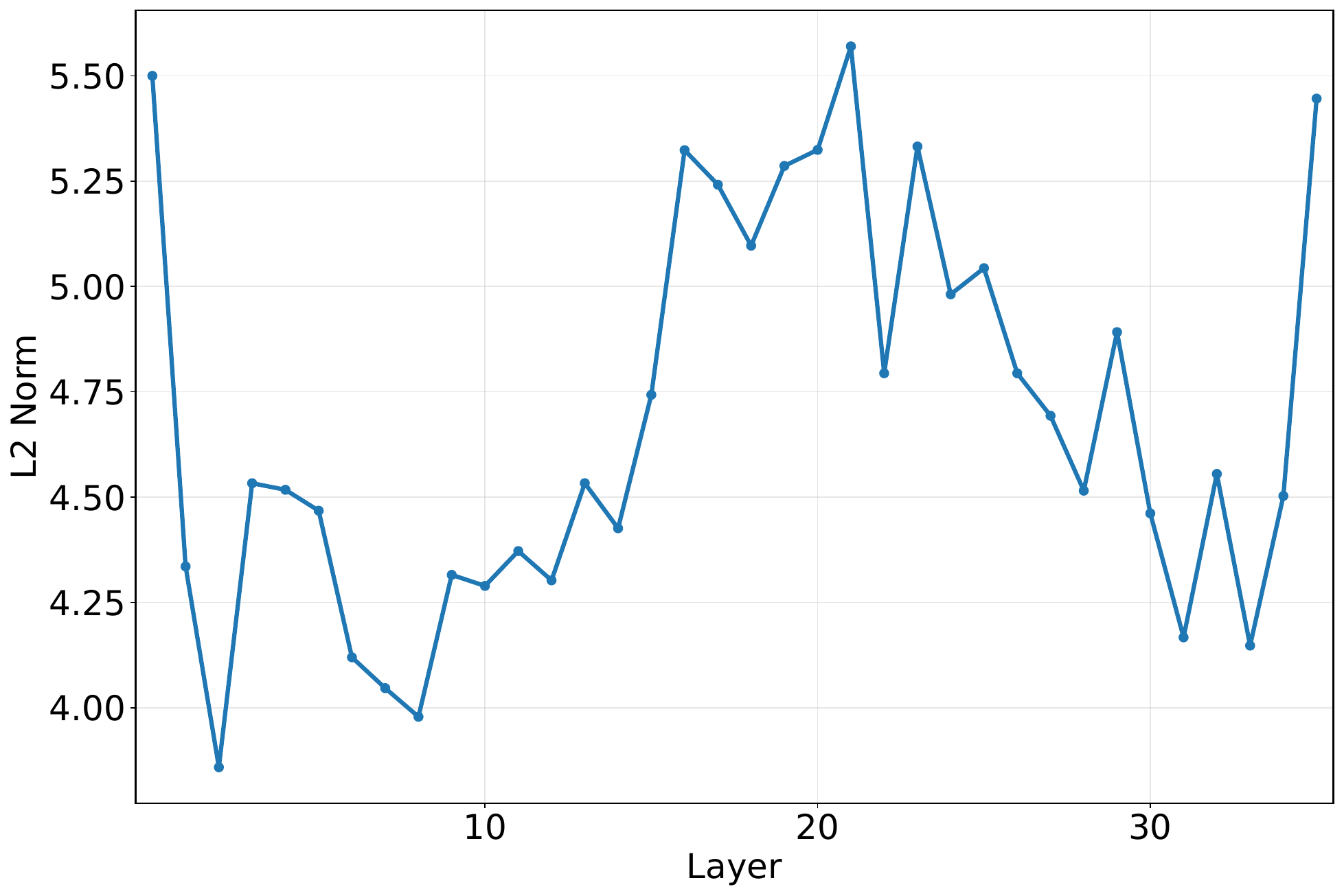}
            \caption{PPO}
        \end{subfigure}\hfill
        \begin{subfigure}[!ht]{0.125\textwidth}
            \includegraphics[width=\linewidth]{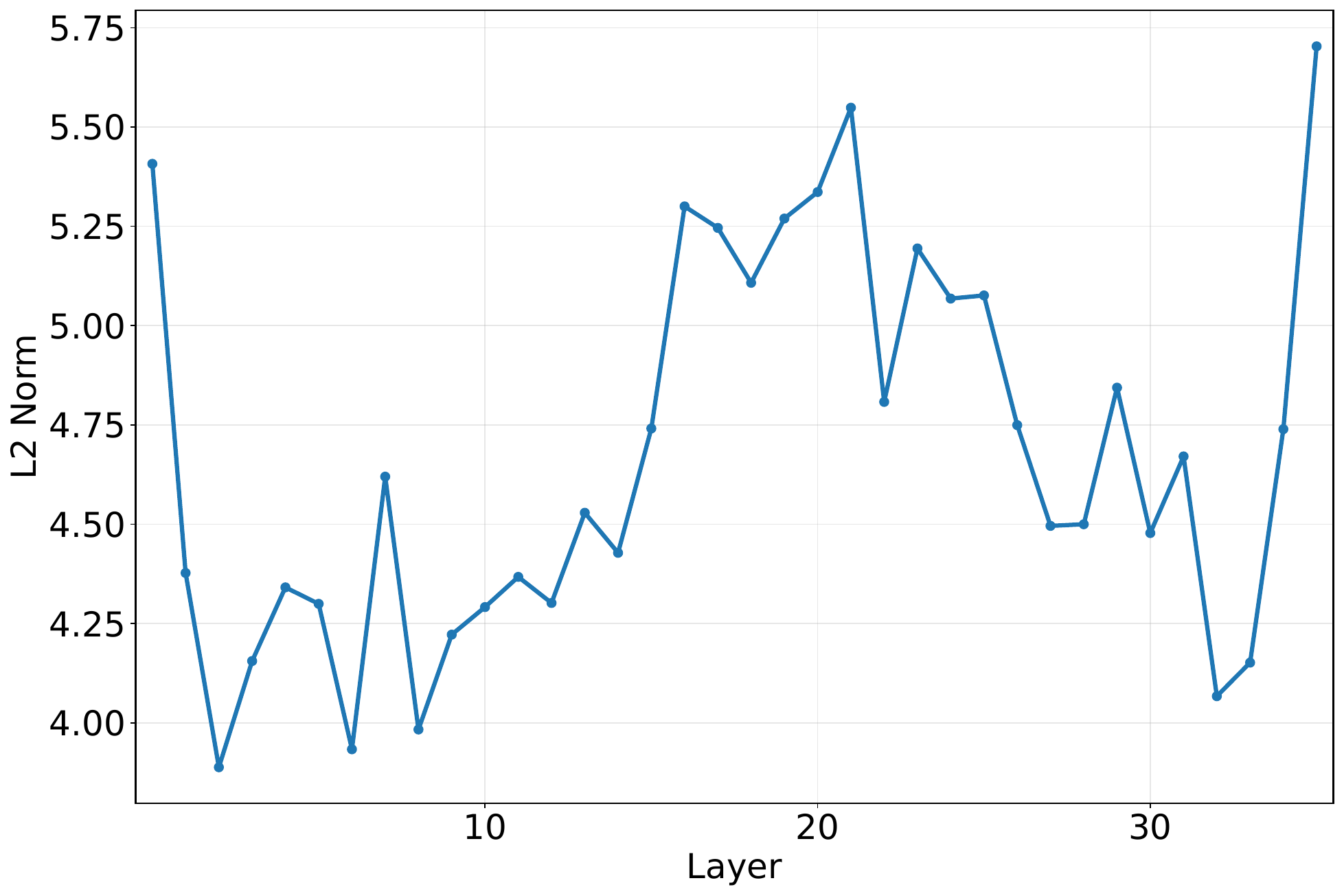}
            \caption{SimPO}
        \end{subfigure}

        \vspace{0.3em}
        \textbf{(i) SmolLM3-3B}
    \end{minipage}

    \vspace{0.6em}

    \begin{minipage}{\textwidth}
        \centering
        \setcounter{subfigure}{0}
        \renewcommand{\thesubfigure}{\alph{subfigure}}
        \begin{subfigure}[!ht]{0.125\textwidth}
            \includegraphics[width=\linewidth]{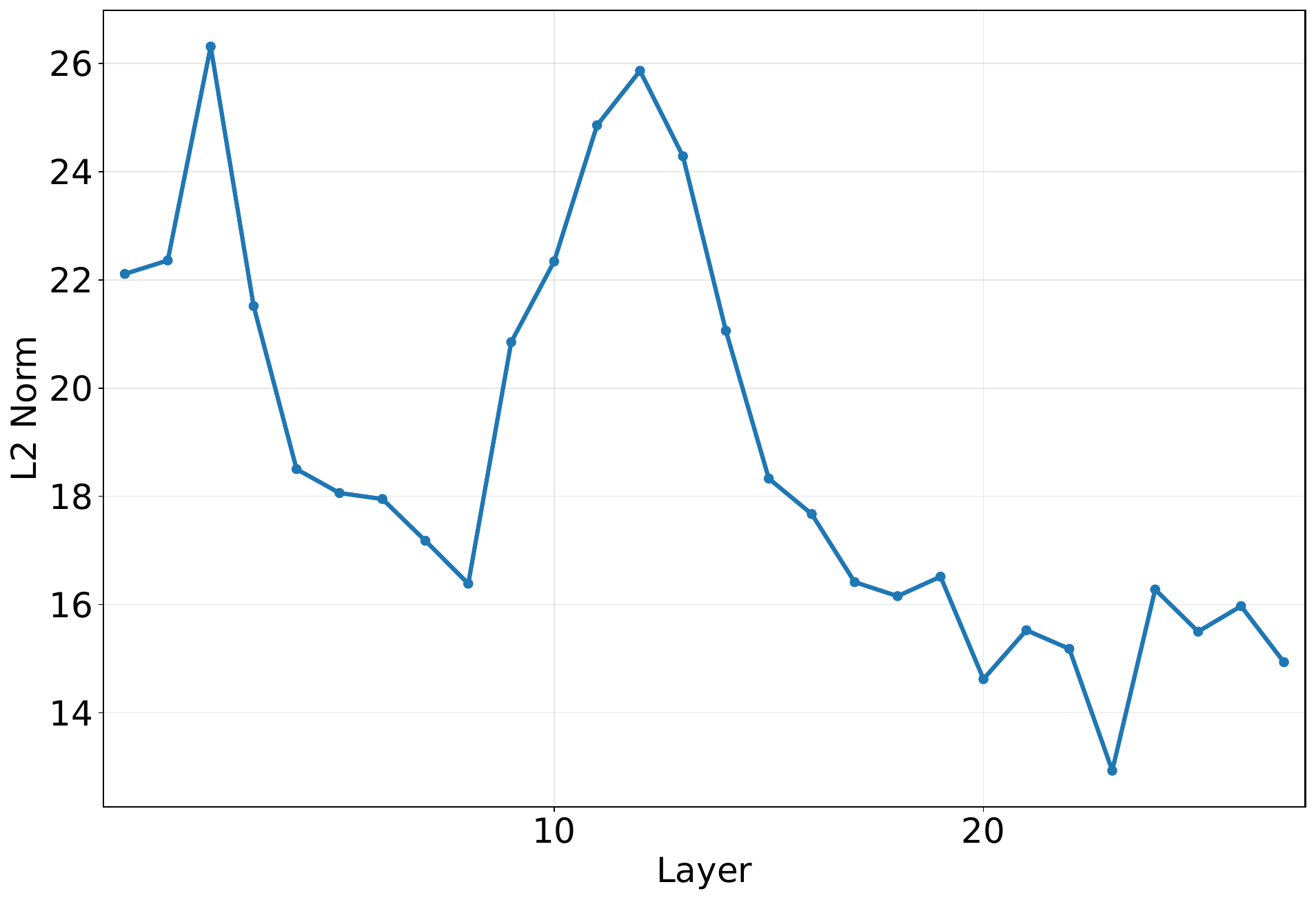}
            \caption{Base}
        \end{subfigure}\hfill
        \begin{subfigure}[!ht]{0.125\textwidth}
            \includegraphics[width=\linewidth]{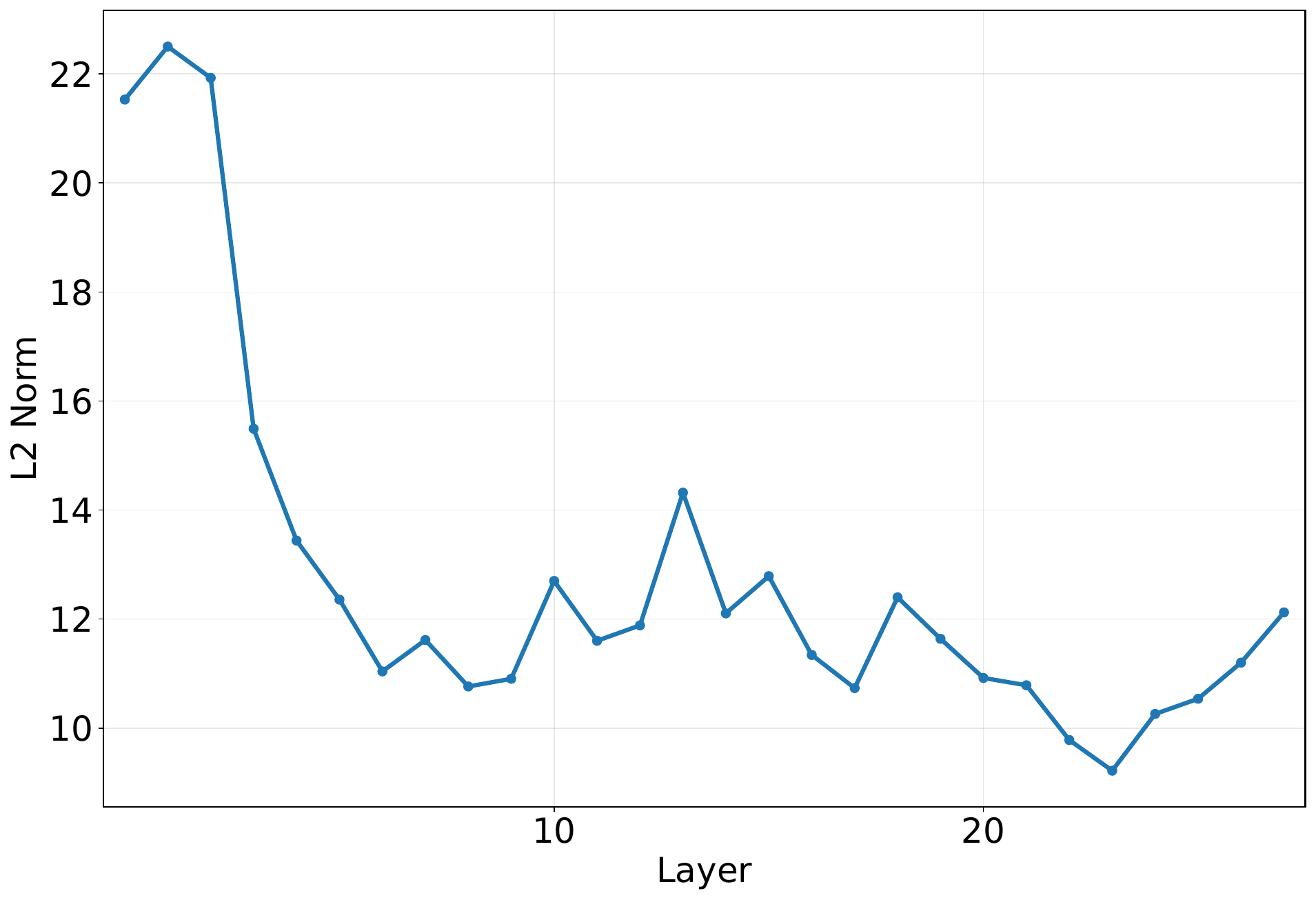}
            \caption{DPO}
        \end{subfigure}\hfill
        \begin{subfigure}[!ht]{0.125\textwidth}
            \includegraphics[width=\linewidth]{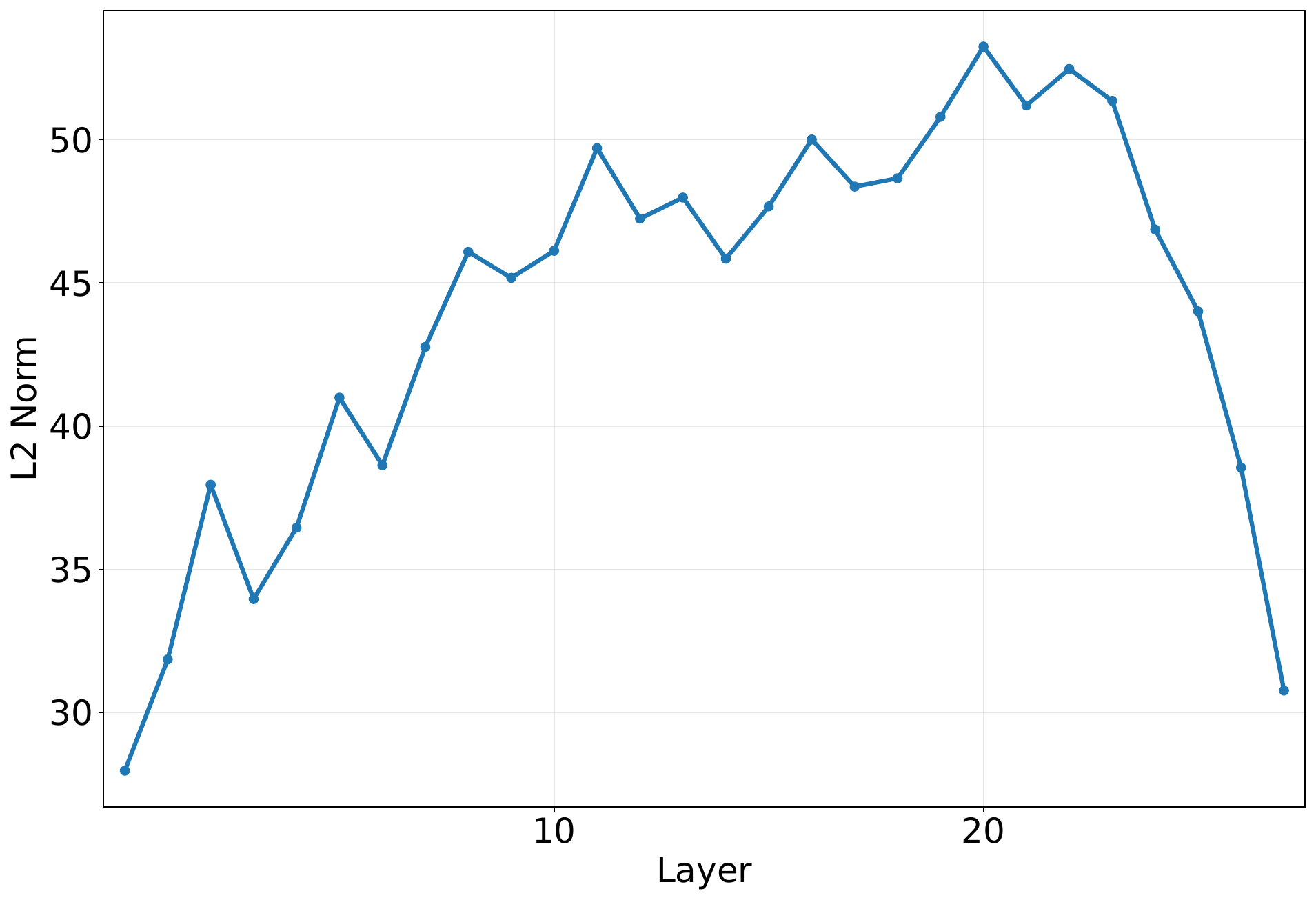}
            \caption{GRPO}
        \end{subfigure}\hfill
        \begin{subfigure}[!ht]{0.125\textwidth}
            \includegraphics[width=\linewidth]{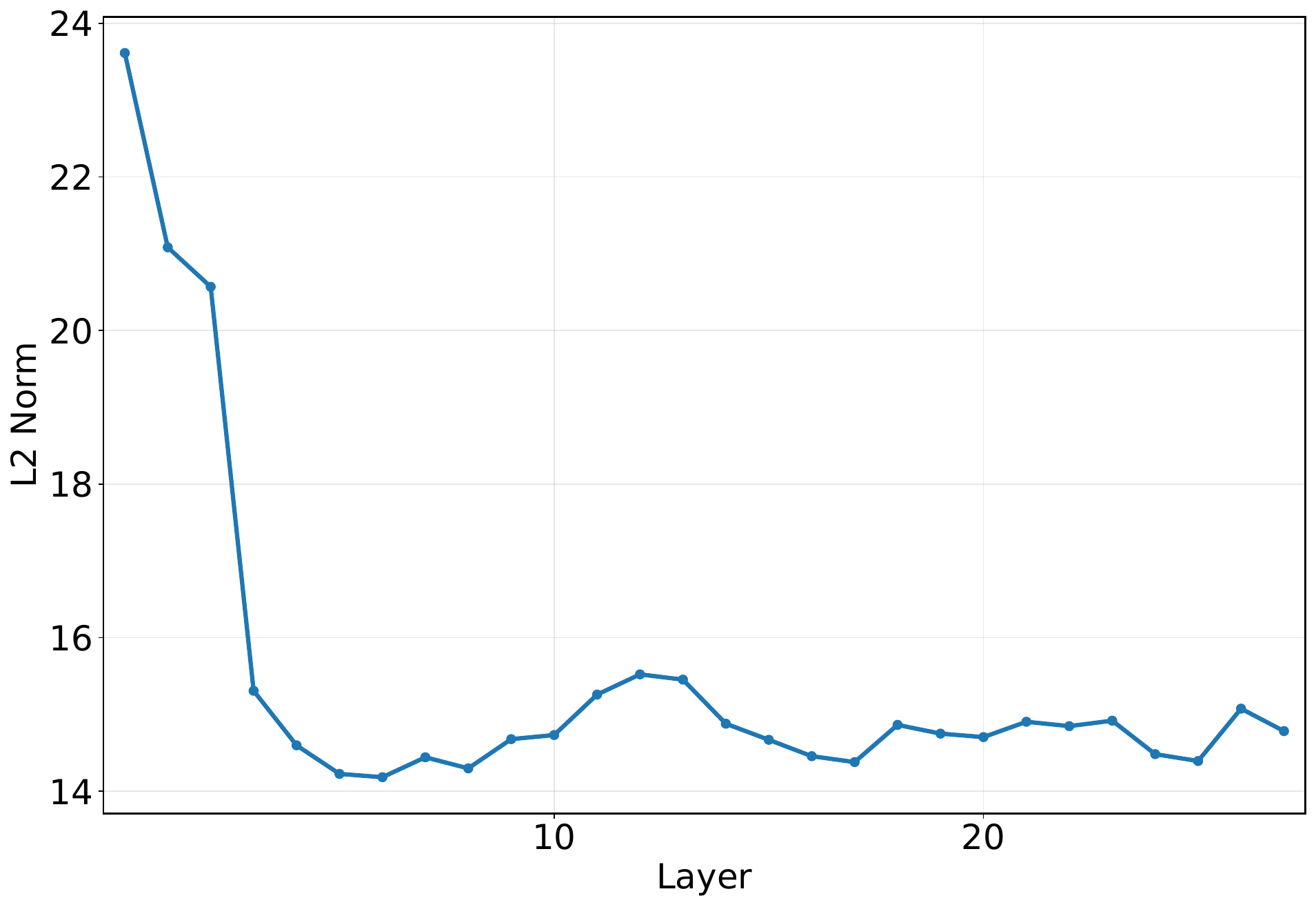}
            \caption{KTO}
        \end{subfigure}\hfill
        \begin{subfigure}[!ht]{0.125\textwidth}
            \includegraphics[width=\linewidth]{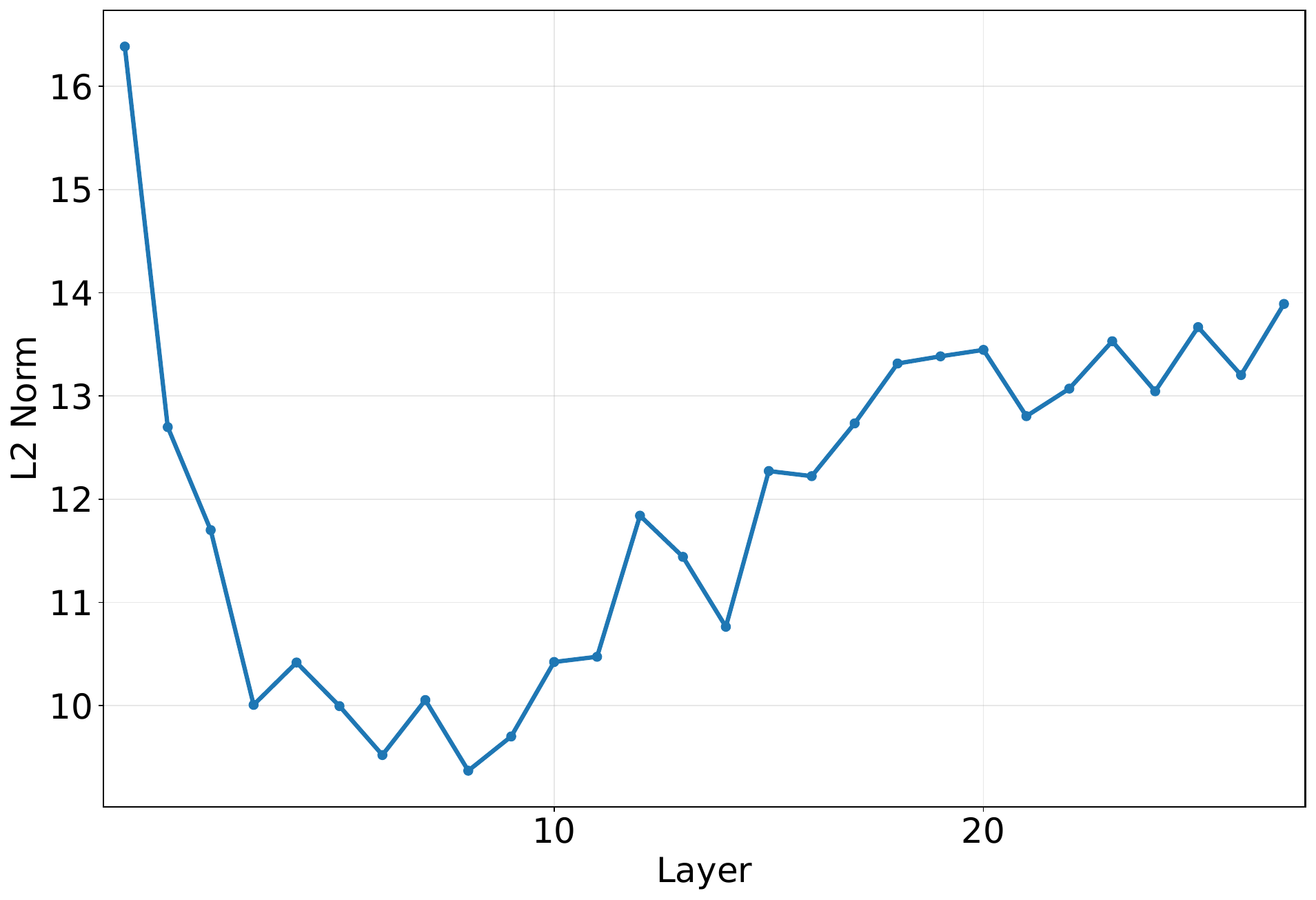}
            \caption{ORPO}
        \end{subfigure}\hfill
        \begin{subfigure}[!ht]{0.125\textwidth}
            \includegraphics[width=\linewidth]{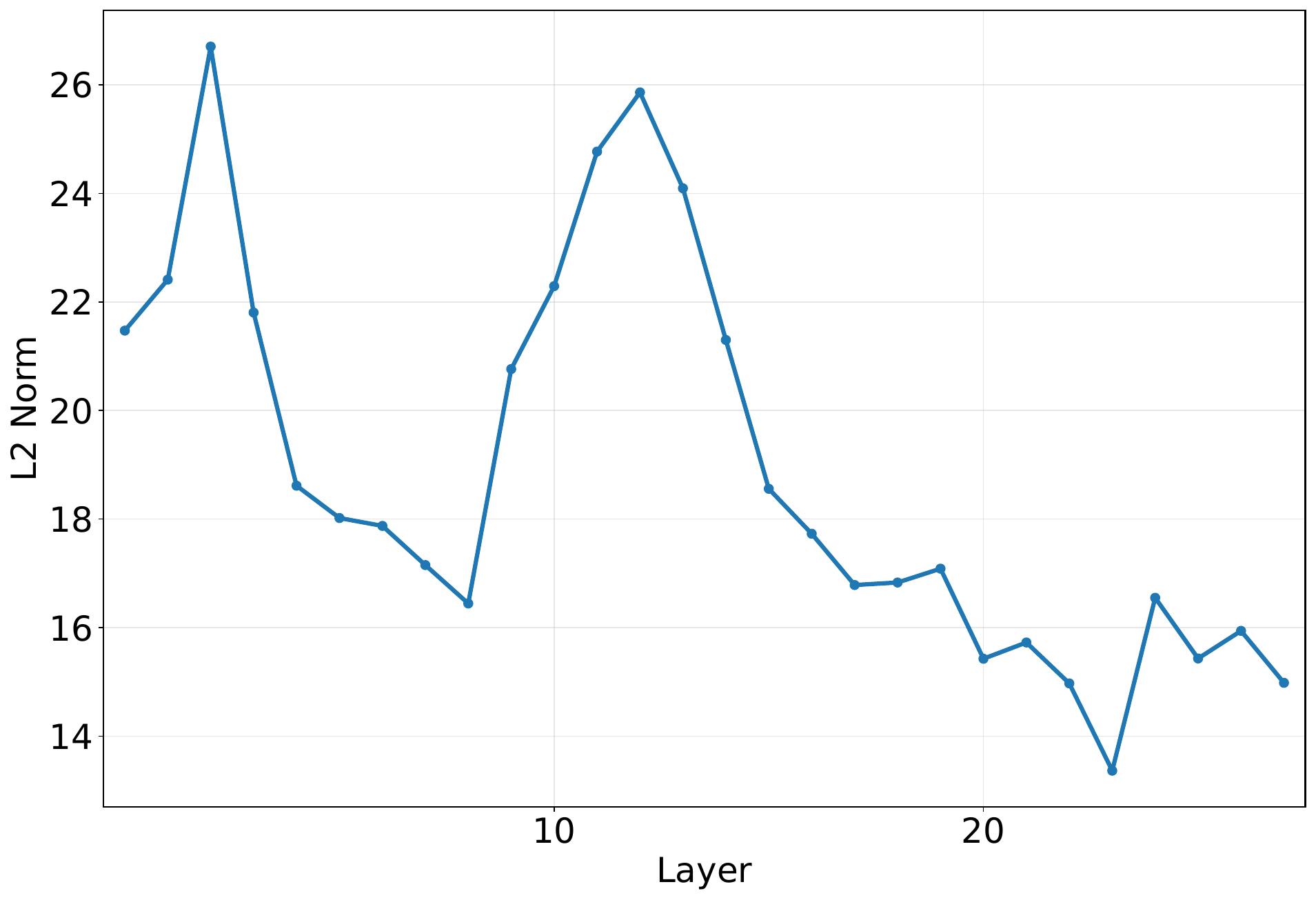}
            \caption{PPO}
        \end{subfigure}\hfill
        \begin{subfigure}[!ht]{0.125\textwidth}
            \includegraphics[width=\linewidth]{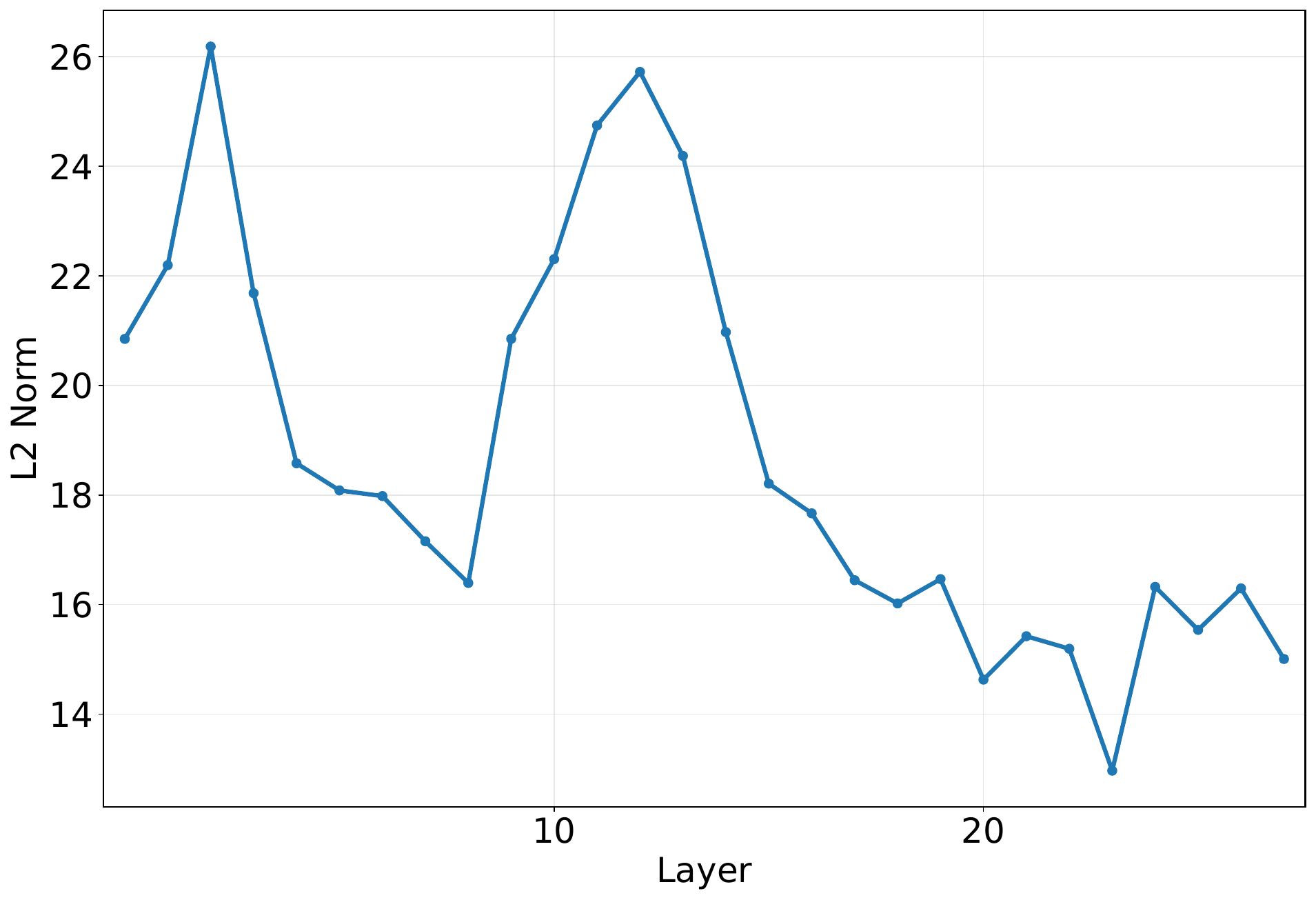}
            \caption{SimPO}
        \end{subfigure}

        \vspace{0.3em}
        \textbf{(ii) Llama-3.2-3B}
    \end{minipage}

    \vspace{0.6em}

    \begin{minipage}{\textwidth}
        \centering
        \setcounter{subfigure}{0}
        \renewcommand{\thesubfigure}{\alph{subfigure}}
        \begin{subfigure}[!ht]{0.125\textwidth}
            \includegraphics[width=\linewidth]{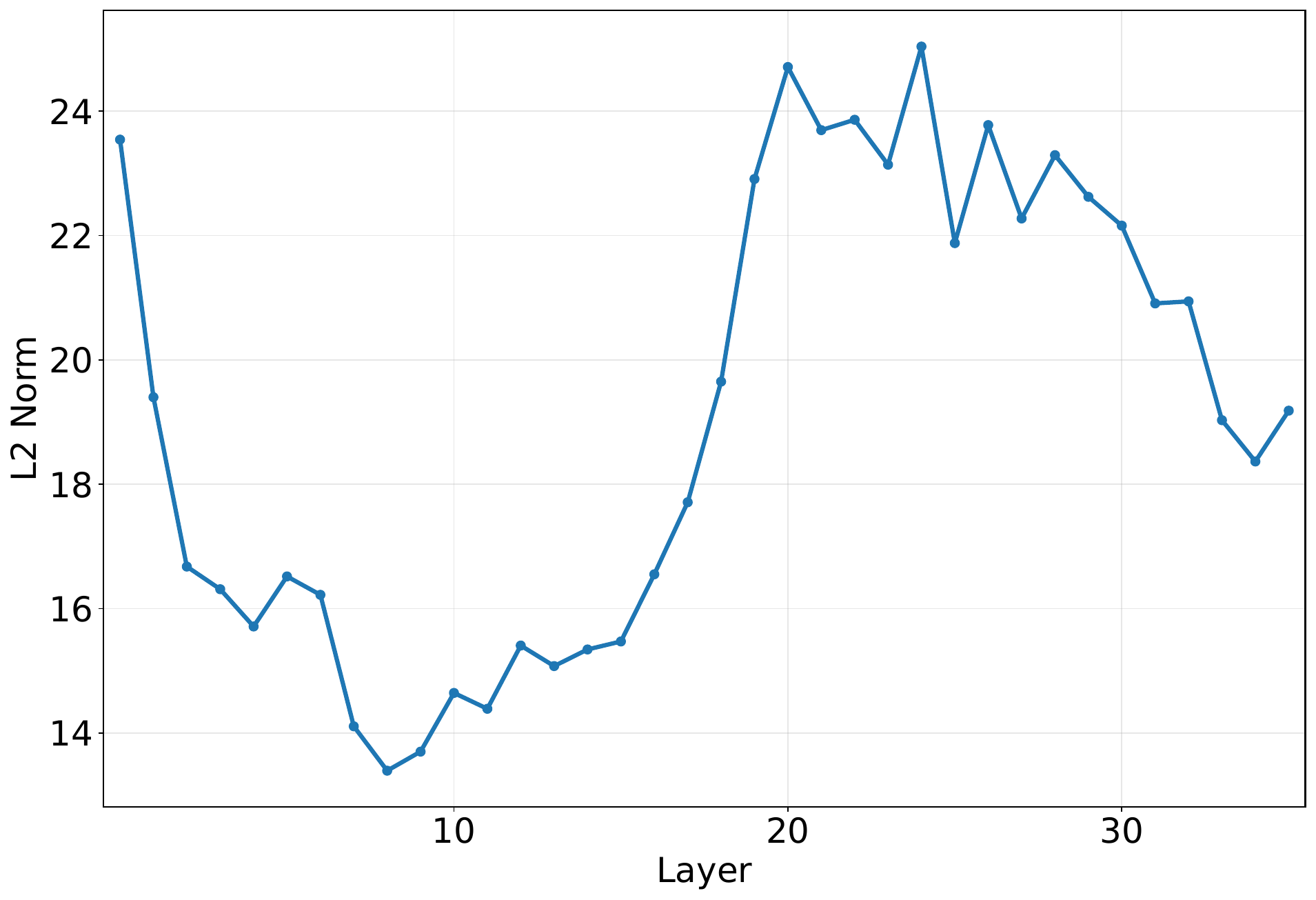}
            \caption{Base}
        \end{subfigure}\hfill
        \begin{subfigure}[!ht]{0.125\textwidth}
            \includegraphics[width=\linewidth]{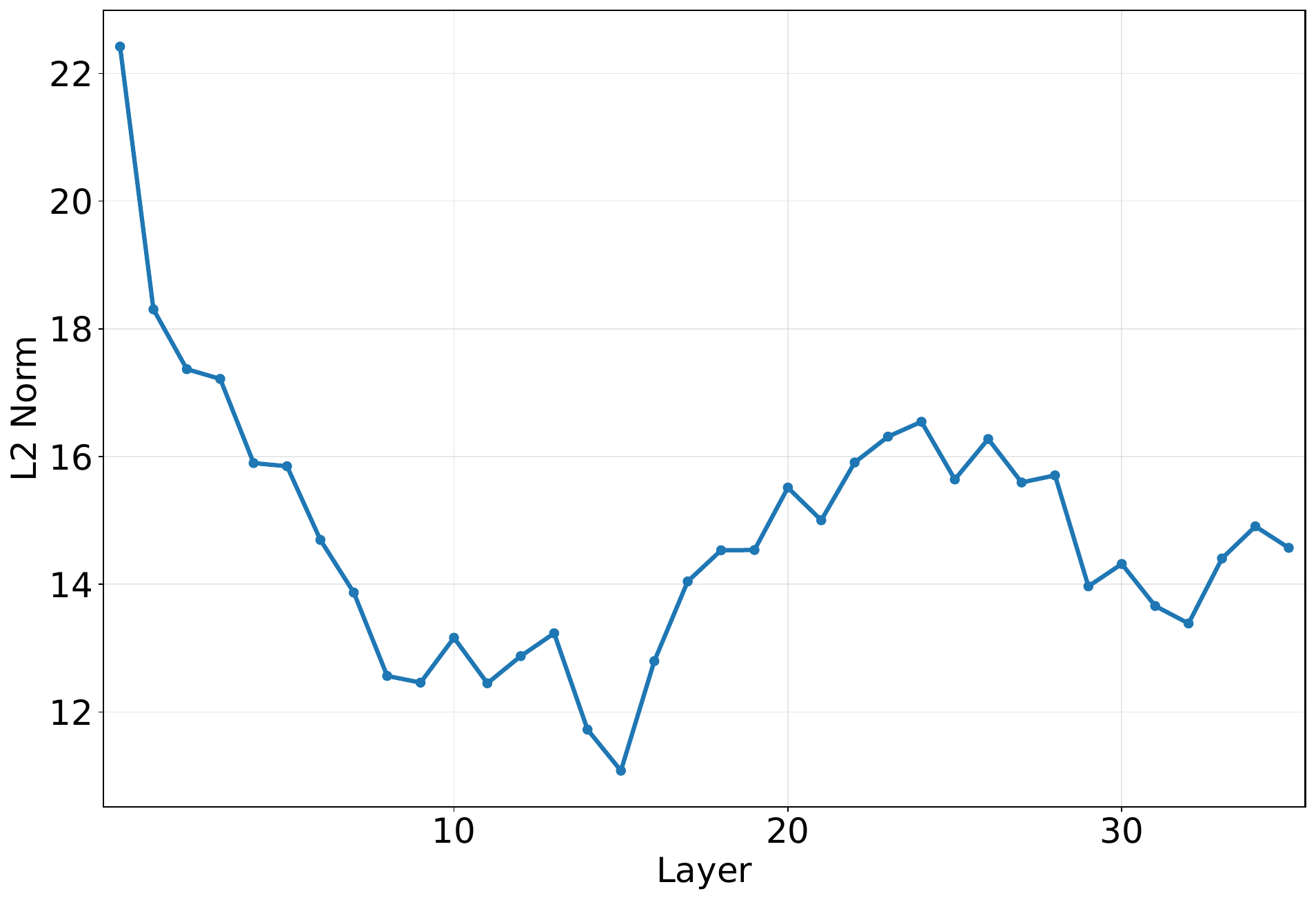}
            \caption{DPO}
        \end{subfigure}\hfill
        \begin{subfigure}[!ht]{0.125\textwidth}
            \includegraphics[width=\linewidth]{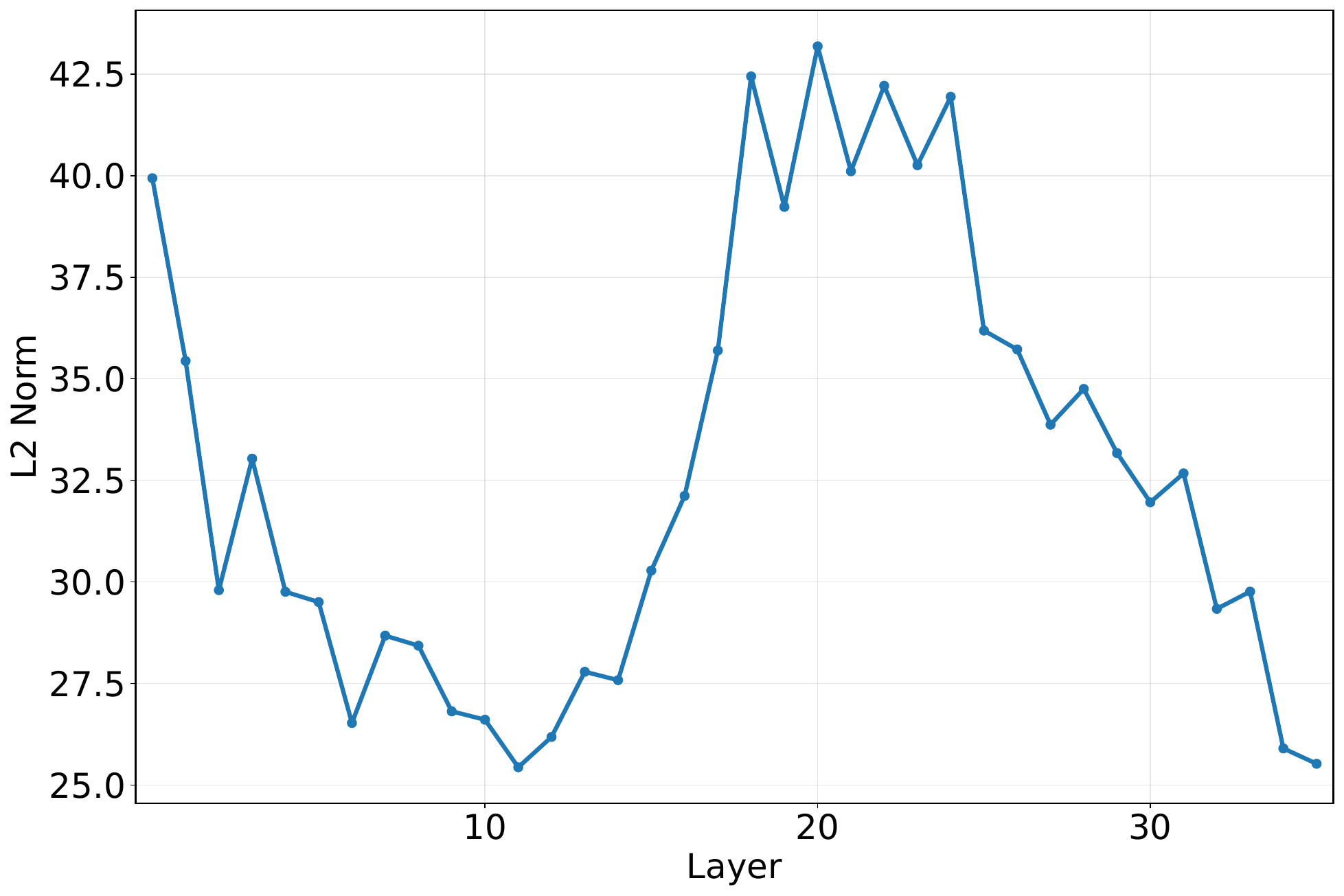}
            \caption{GRPO}
        \end{subfigure}\hfill
        \begin{subfigure}[!ht]{0.125\textwidth}
            \includegraphics[width=\linewidth]{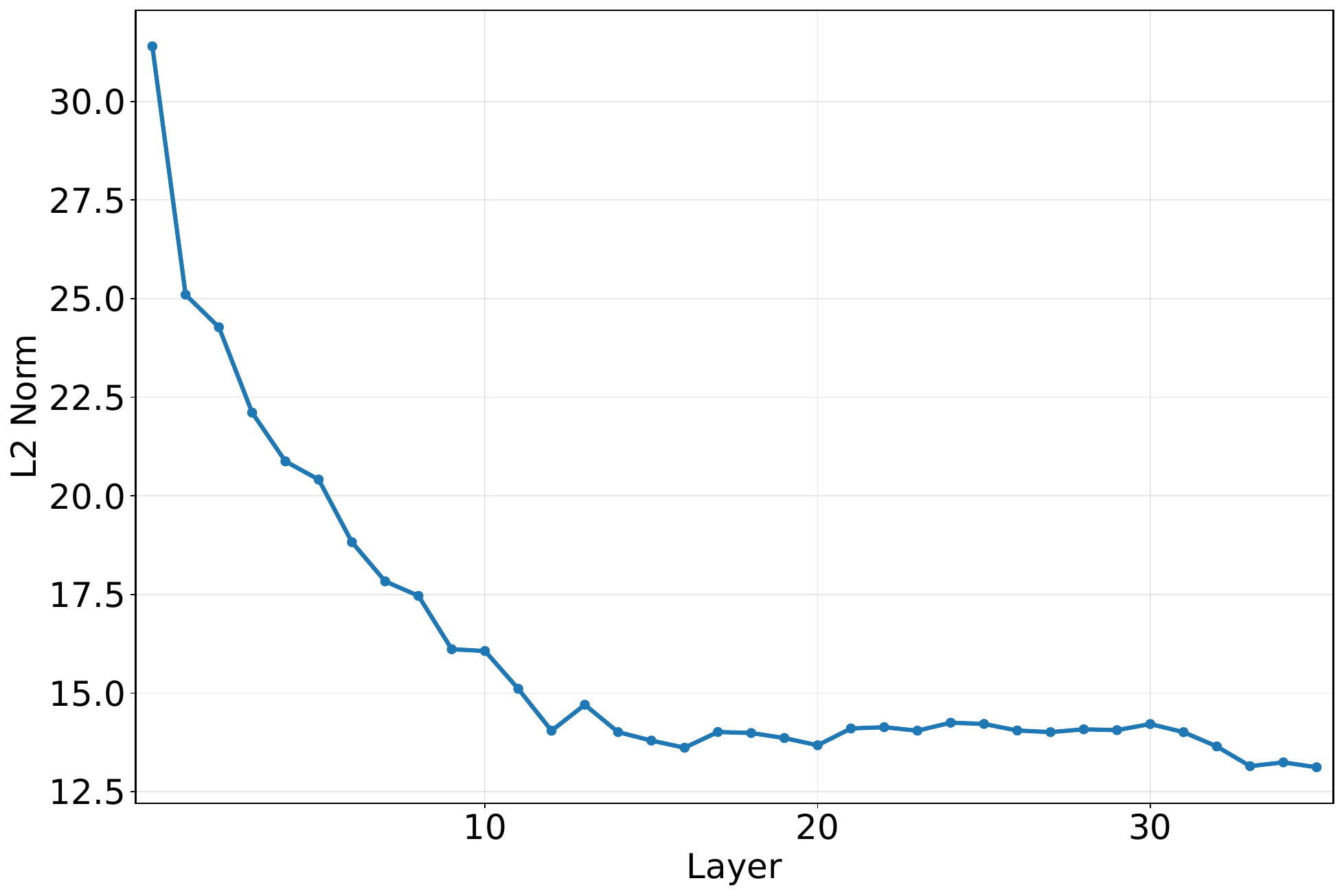}
            \caption{KTO}
        \end{subfigure}\hfill
        \begin{subfigure}[!ht]{0.125\textwidth}
            \includegraphics[width=\linewidth]{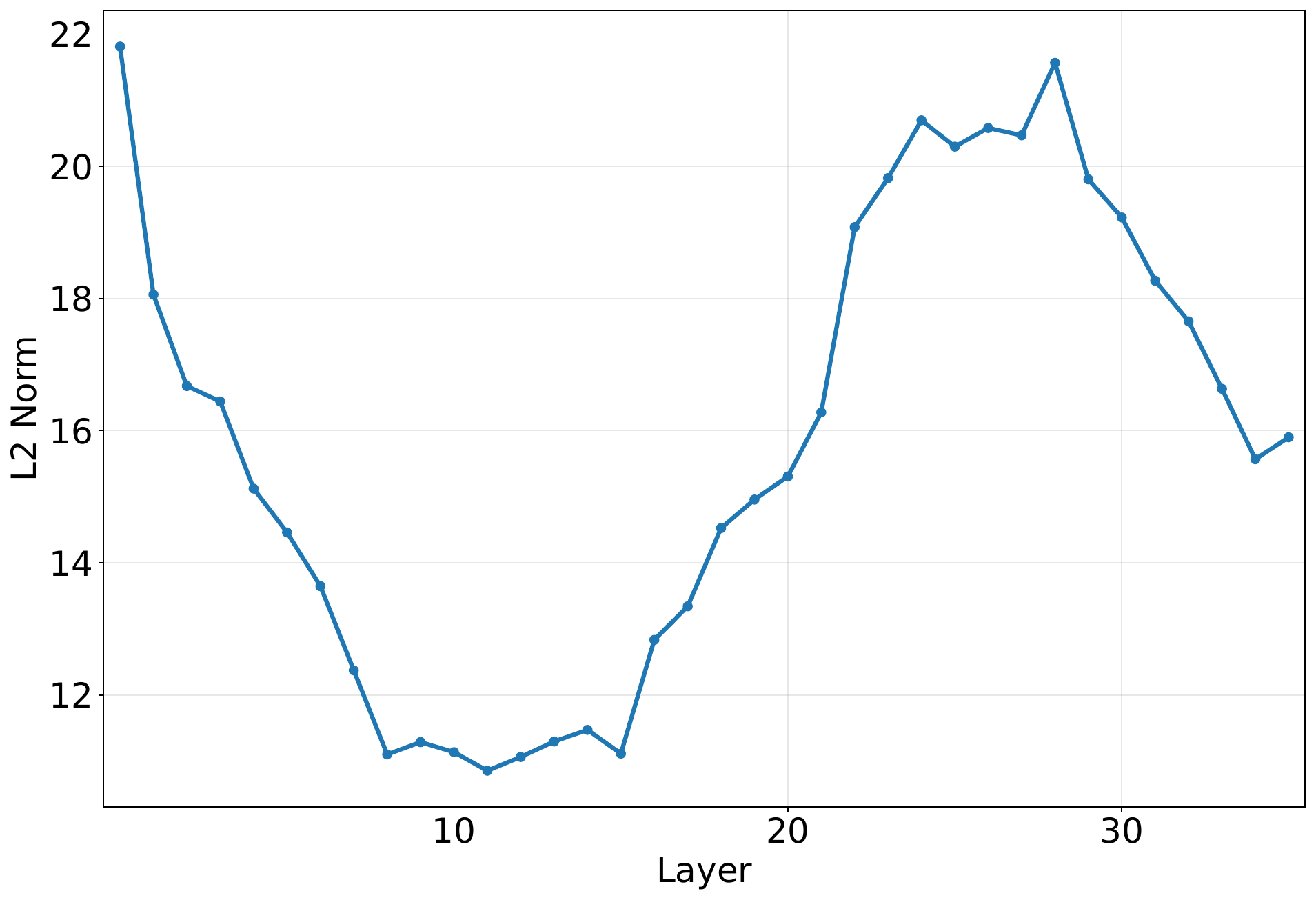}
            \caption{ORPO}
        \end{subfigure}\hfill
        \begin{subfigure}[!ht]{0.125\textwidth}
            \includegraphics[width=\linewidth]{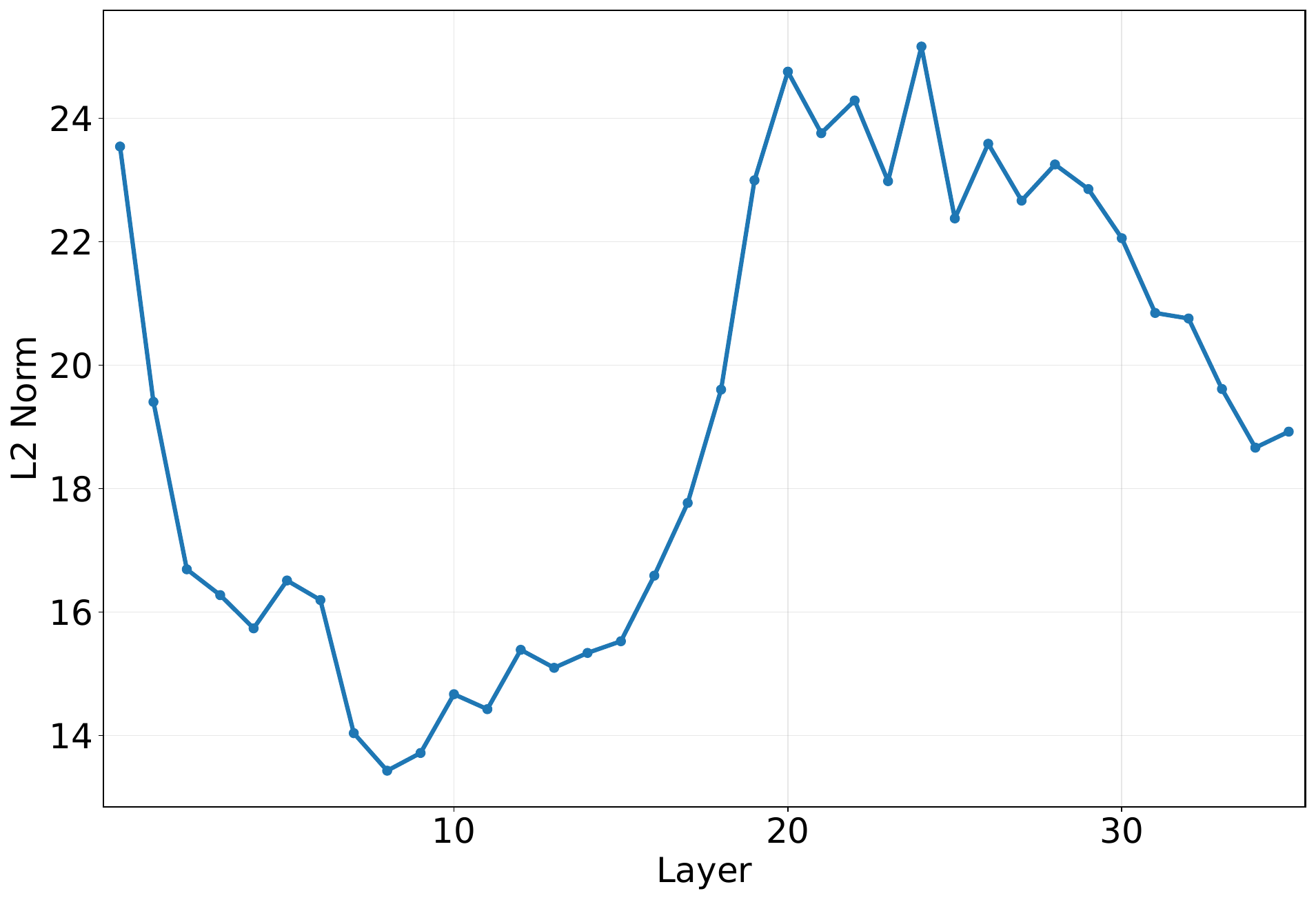}
            \caption{PPO}
        \end{subfigure}\hfill
        \begin{subfigure}[!ht]{0.125\textwidth}
            \includegraphics[width=\linewidth]{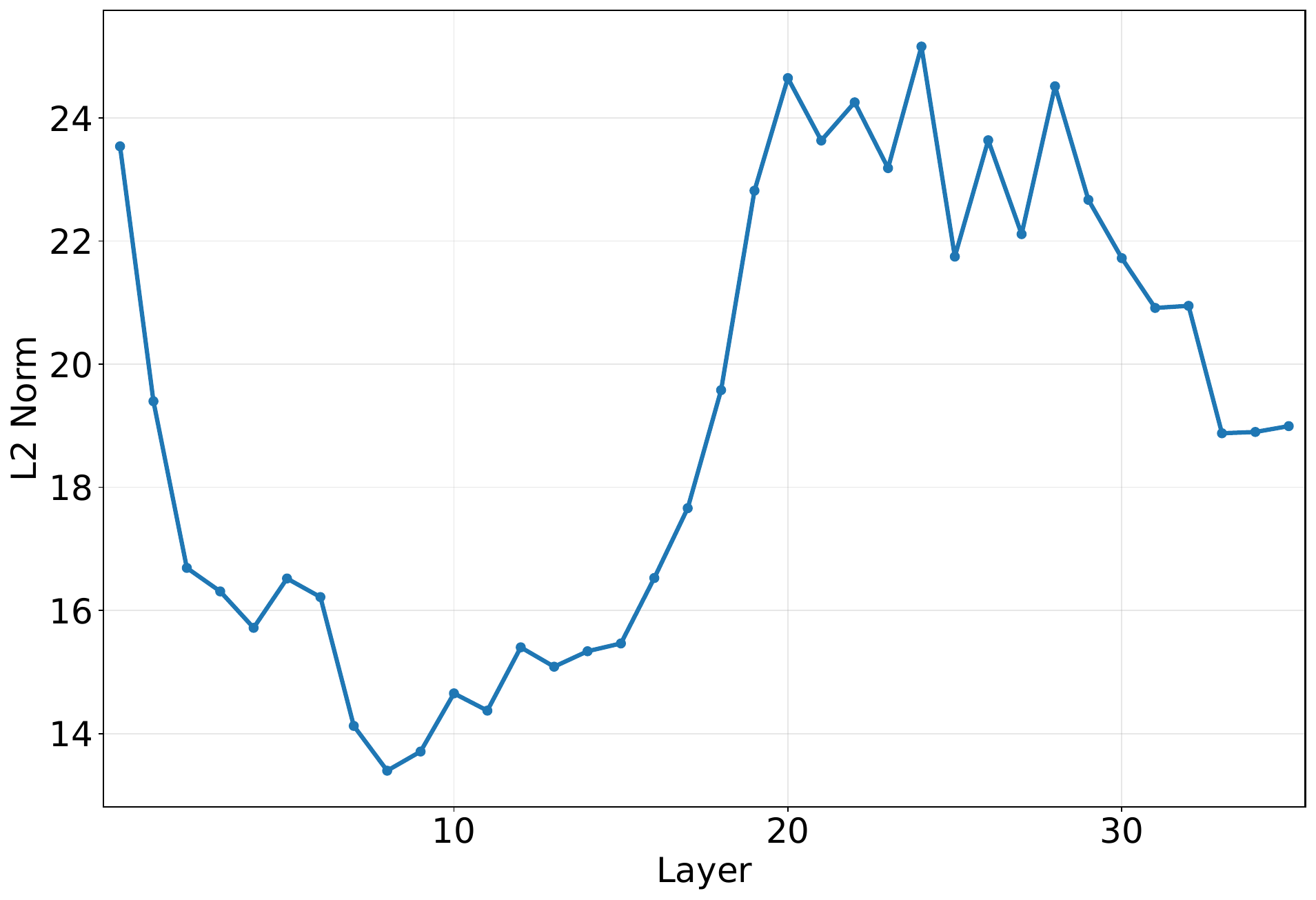}
            \caption{SimPO}
        \end{subfigure}

        \vspace{0.3em}
        \textbf{(iii) Qwen3-4B}
    \end{minipage}

    \caption{Layer-wise $\ell_2$ norm of the linear probe weight vector for each base model and AFT method (mean over 5-fold cross-validation at each layer).}
    \label{fig:lp_layerwise_coef_norm_all_models}
\end{figure*}

\begin{figure}[!ht]
    \centering
    \begin{subfigure}[t]{0.32\linewidth}
        \includegraphics[width=\linewidth]{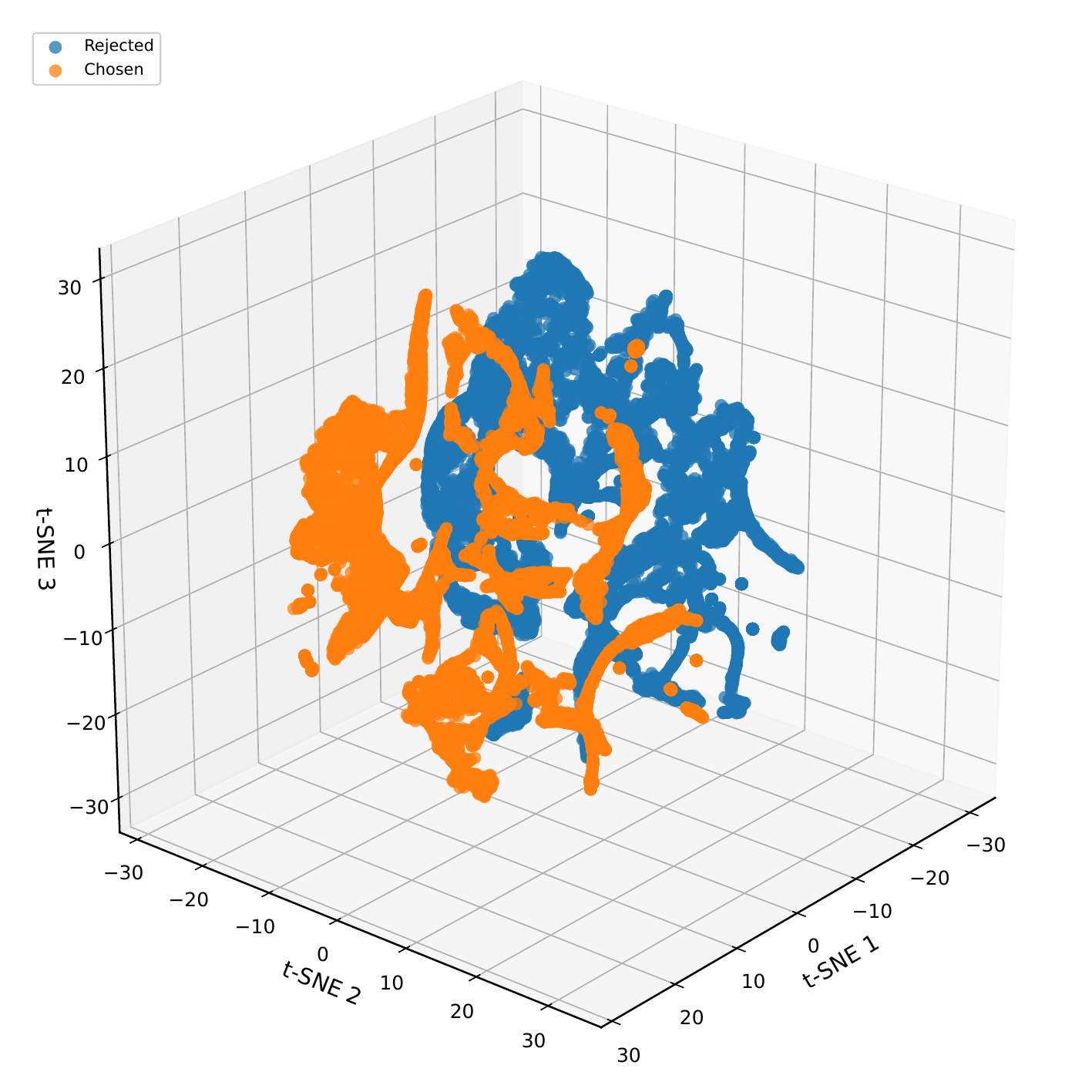}
        \caption{DPO}
    \end{subfigure}
    \hfill
    \begin{subfigure}[t]{0.32\linewidth}
        \includegraphics[width=\linewidth]{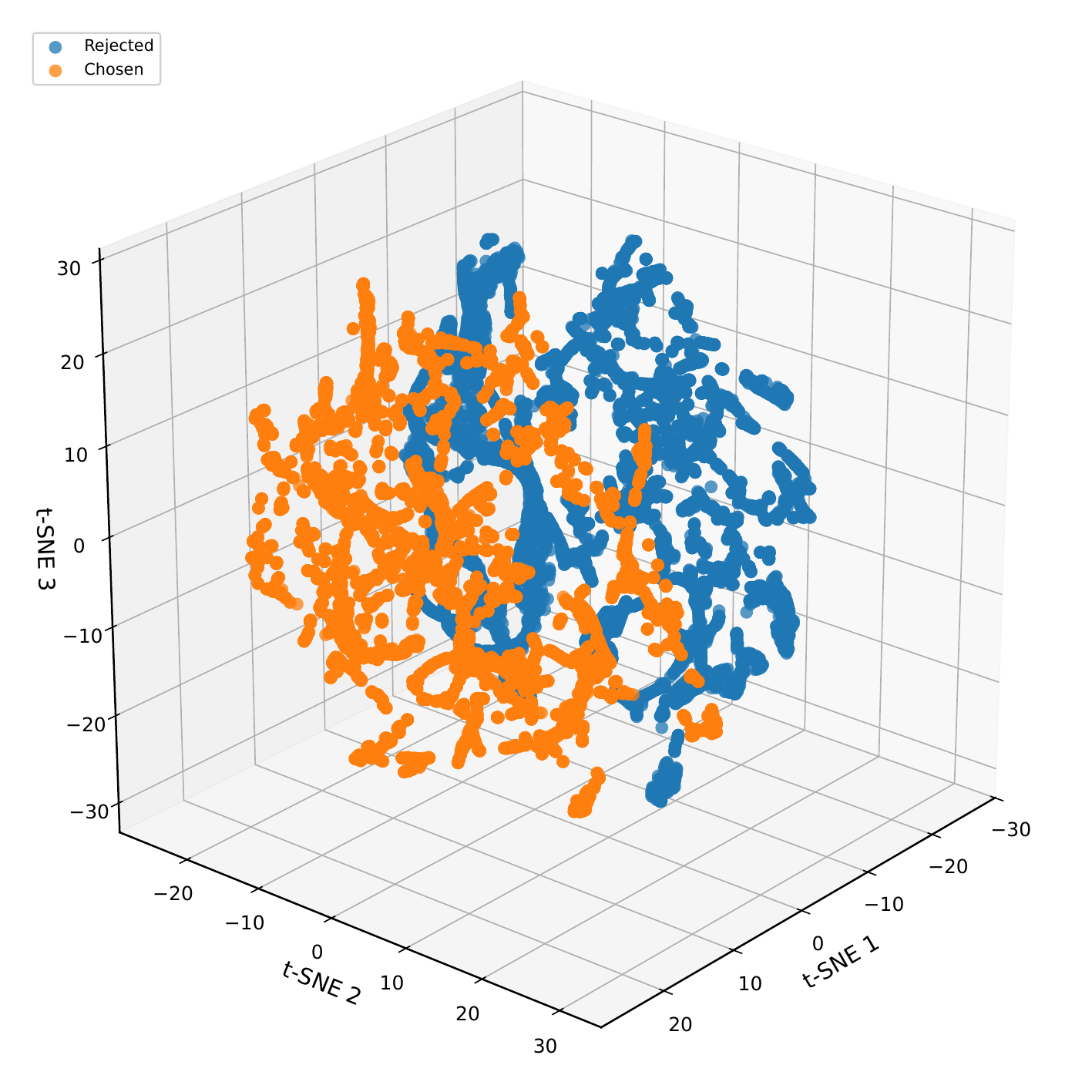}
        \caption{GRPO}
    \end{subfigure}
    \hfill
    \begin{subfigure}[t]{0.32\linewidth}
        \includegraphics[width=\linewidth]{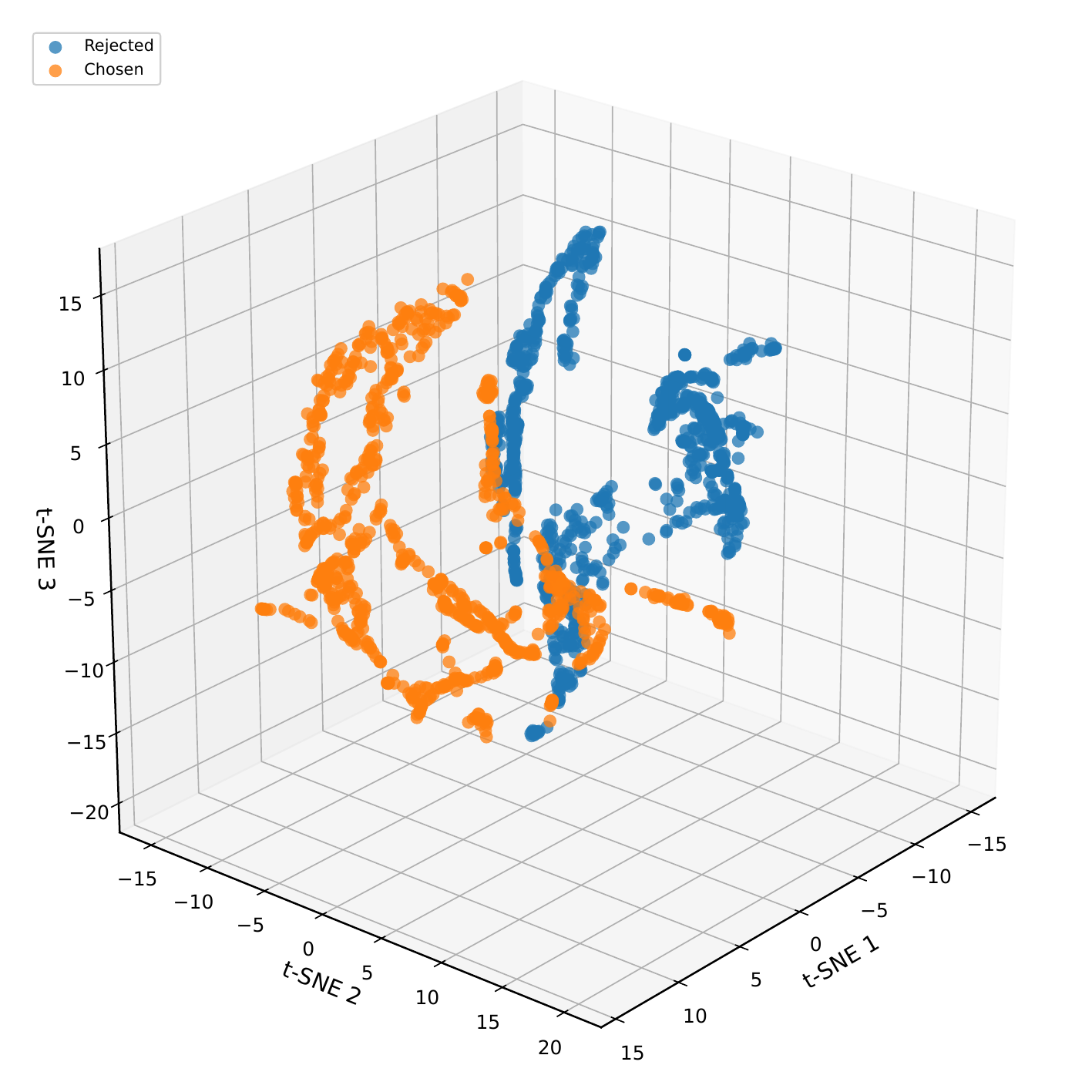}
        \caption{KTO}
    \end{subfigure}

    \vspace{0.9em}
    \begin{subfigure}[t]{0.32\linewidth}
        \includegraphics[width=\linewidth]{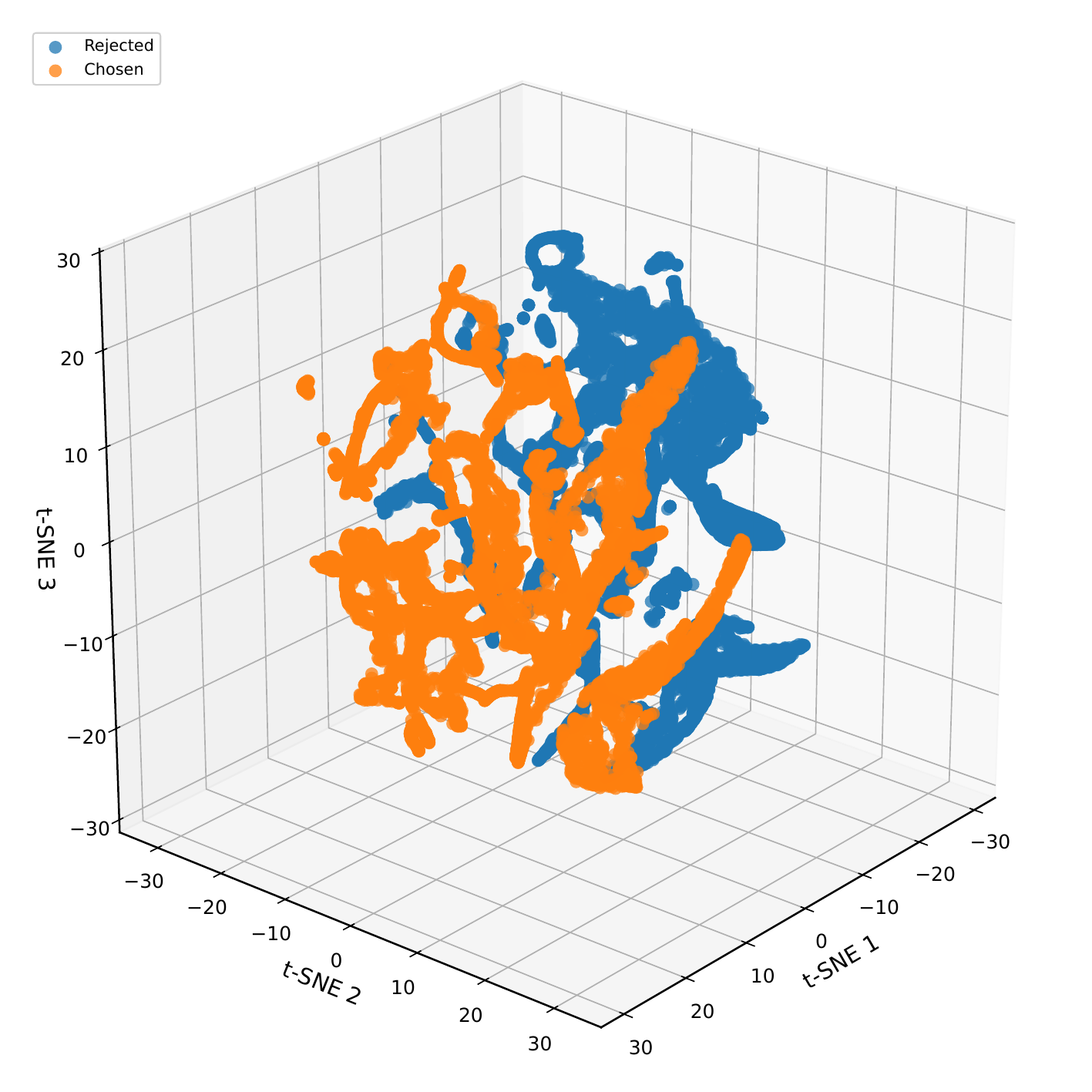}
        \caption{ORPO}
    \end{subfigure}
    \hfill
    \begin{subfigure}[t]{0.32\linewidth}
        \includegraphics[width=\linewidth]{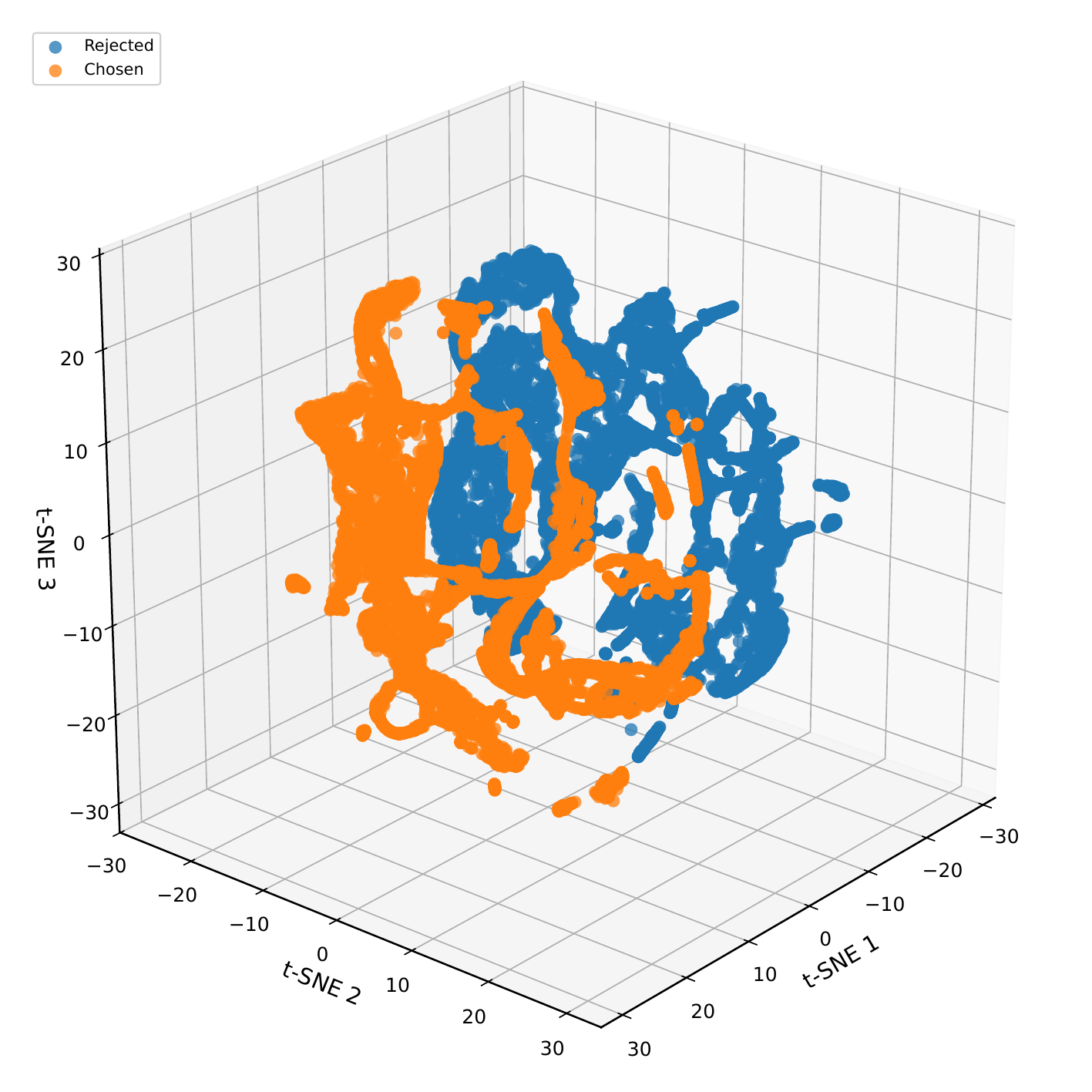}
        \caption{PPO}
    \end{subfigure}
    \hfill
    \begin{subfigure}[t]{0.32\linewidth}
        \includegraphics[width=\linewidth]{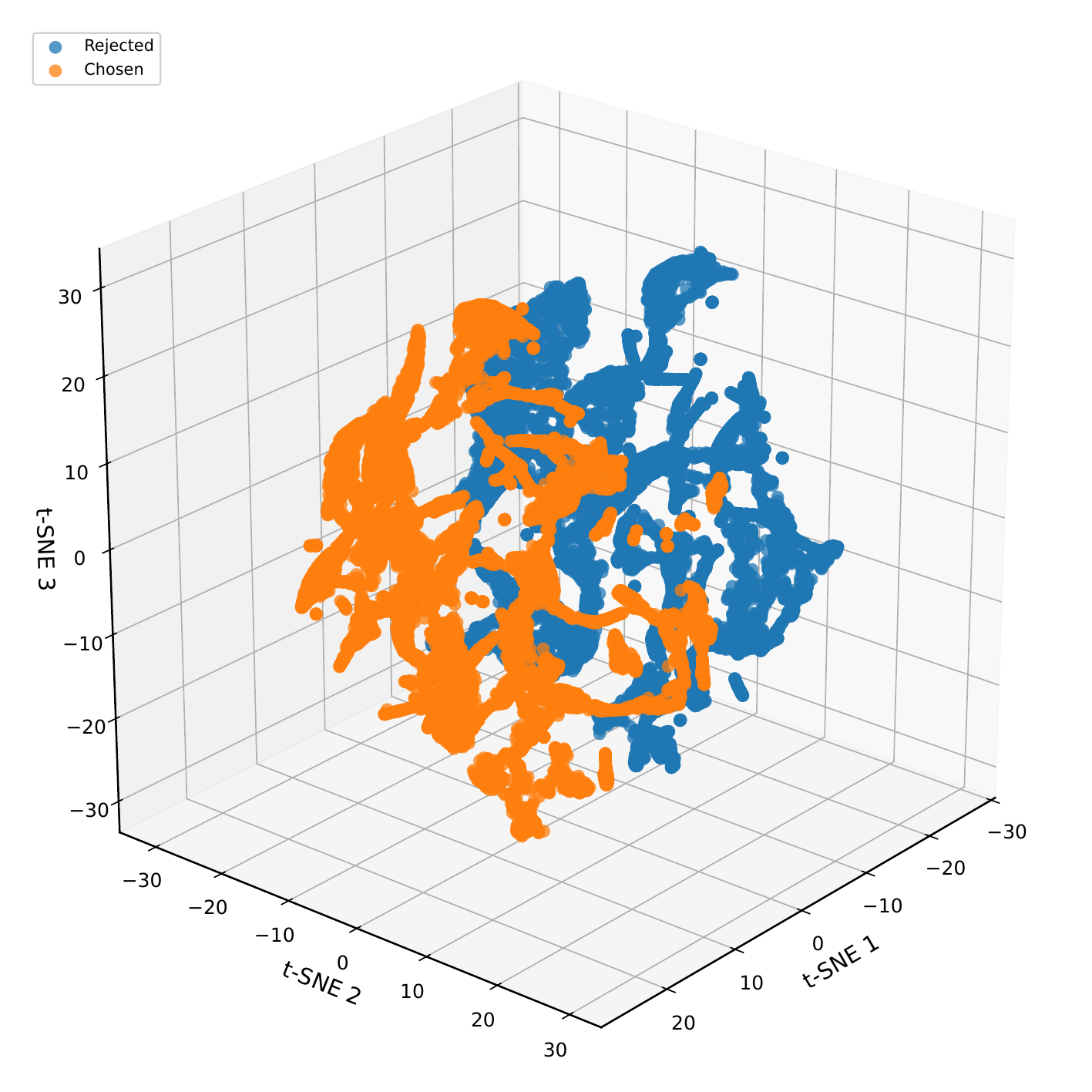}
        \caption{SimPO}
    \end{subfigure}
    \caption{%
        \textbf{SmolLM3-3B.} PCA of residual-stream activations at each method's best probe layer..
    }
    \label{fig:lp_pca_grid_smollm3}
\end{figure}

\begin{figure}[!ht]
    \centering
    \begin{subfigure}[t]{0.32\linewidth}
        \includegraphics[width=\linewidth]{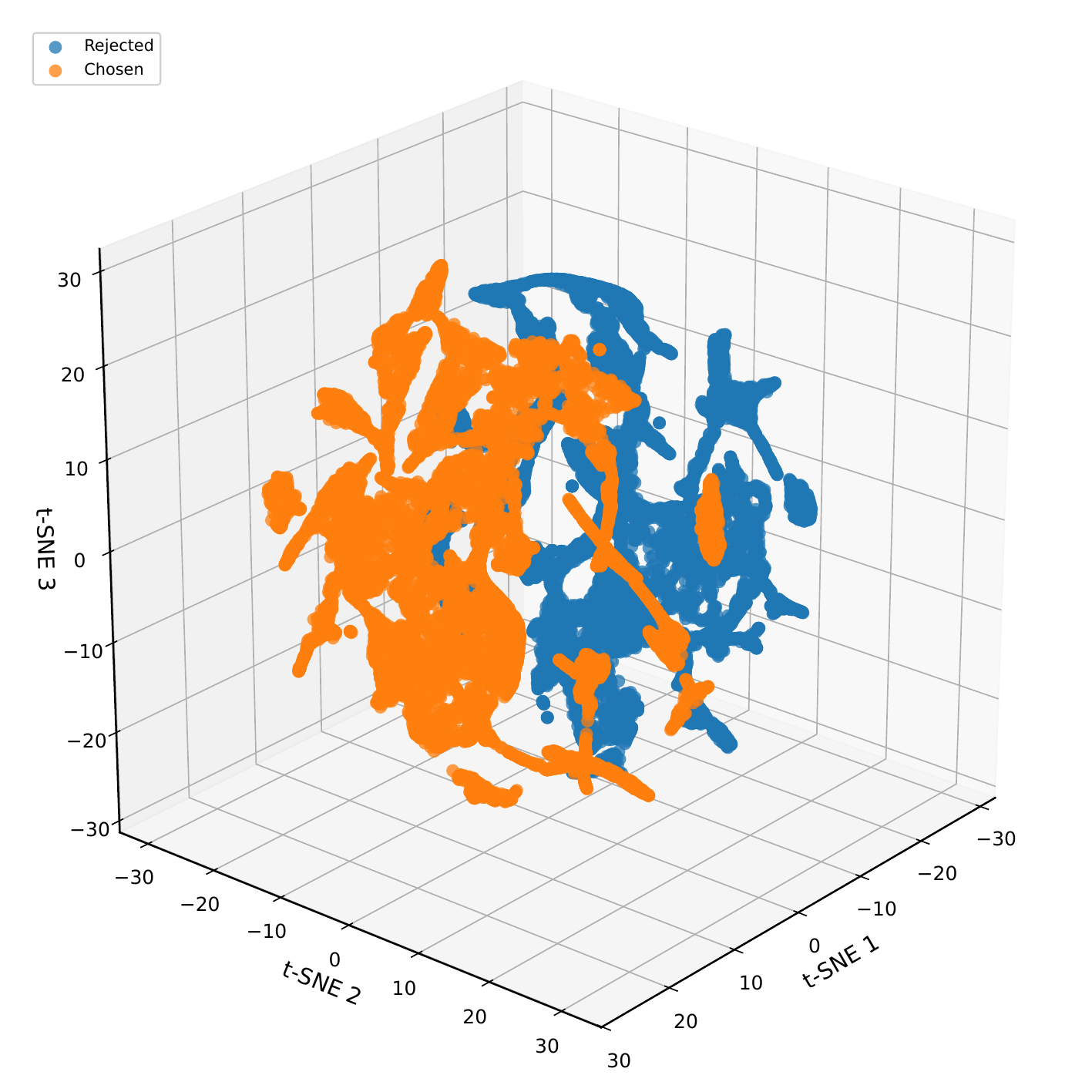}
        \caption{DPO}
    \end{subfigure}
    \hfill
    \begin{subfigure}[t]{0.32\linewidth}
        \includegraphics[width=\linewidth]{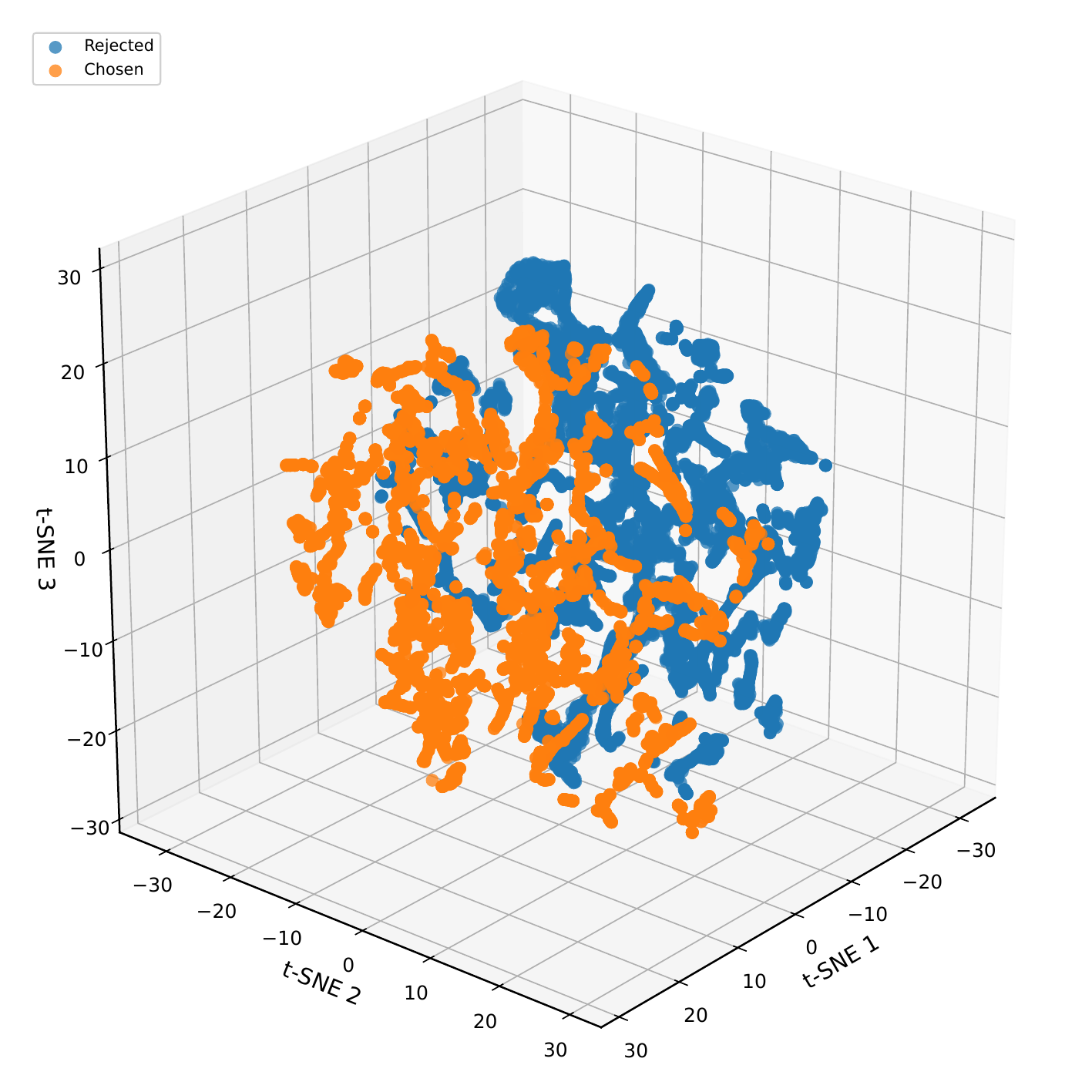}
        \caption{GRPO}
    \end{subfigure}
    \hfill
    \begin{subfigure}[t]{0.32\linewidth}
        \includegraphics[width=\linewidth]{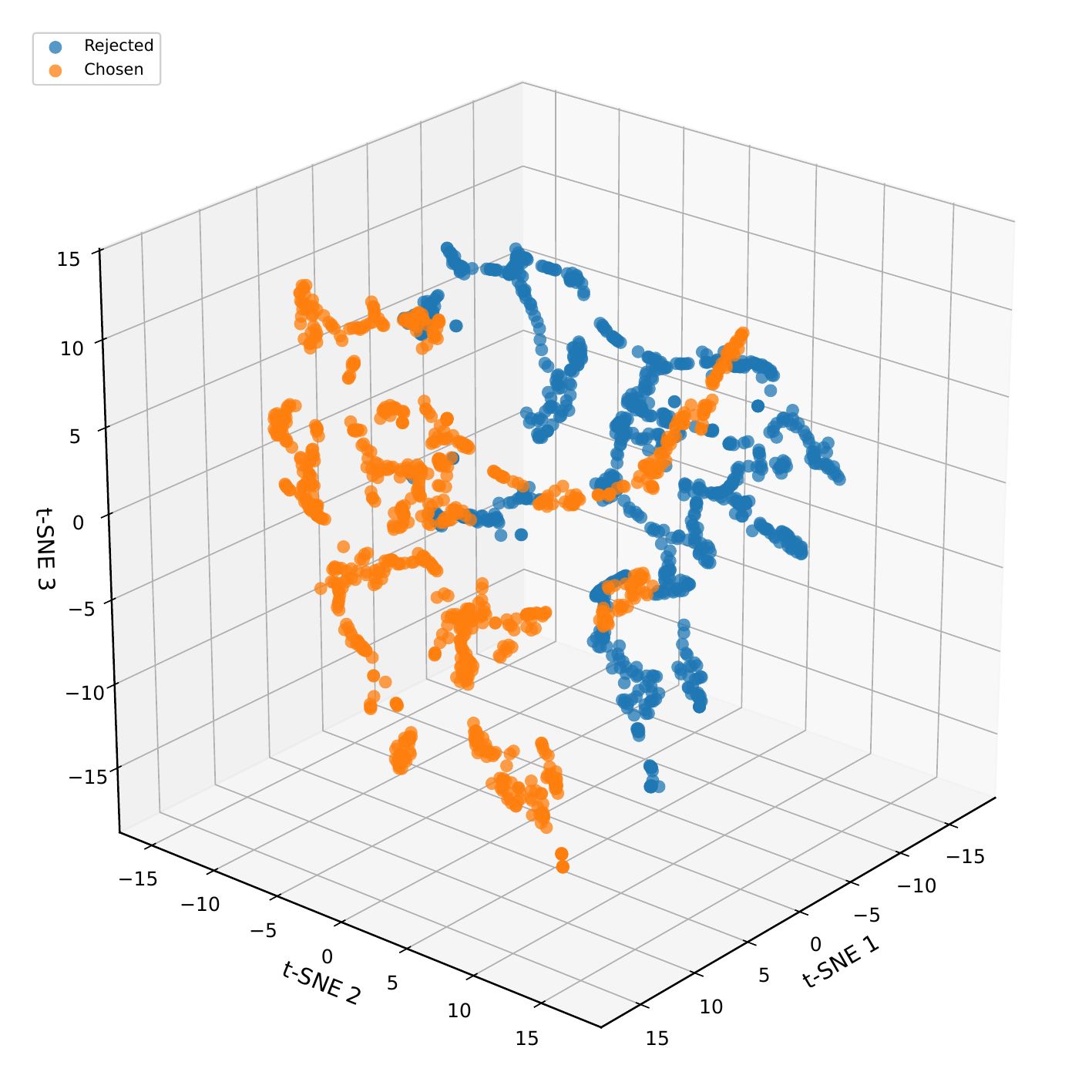}
        \caption{KTO}
    \end{subfigure}

    \vspace{0.9em}
    \begin{subfigure}[t]{0.32\linewidth}
        \includegraphics[width=\linewidth]{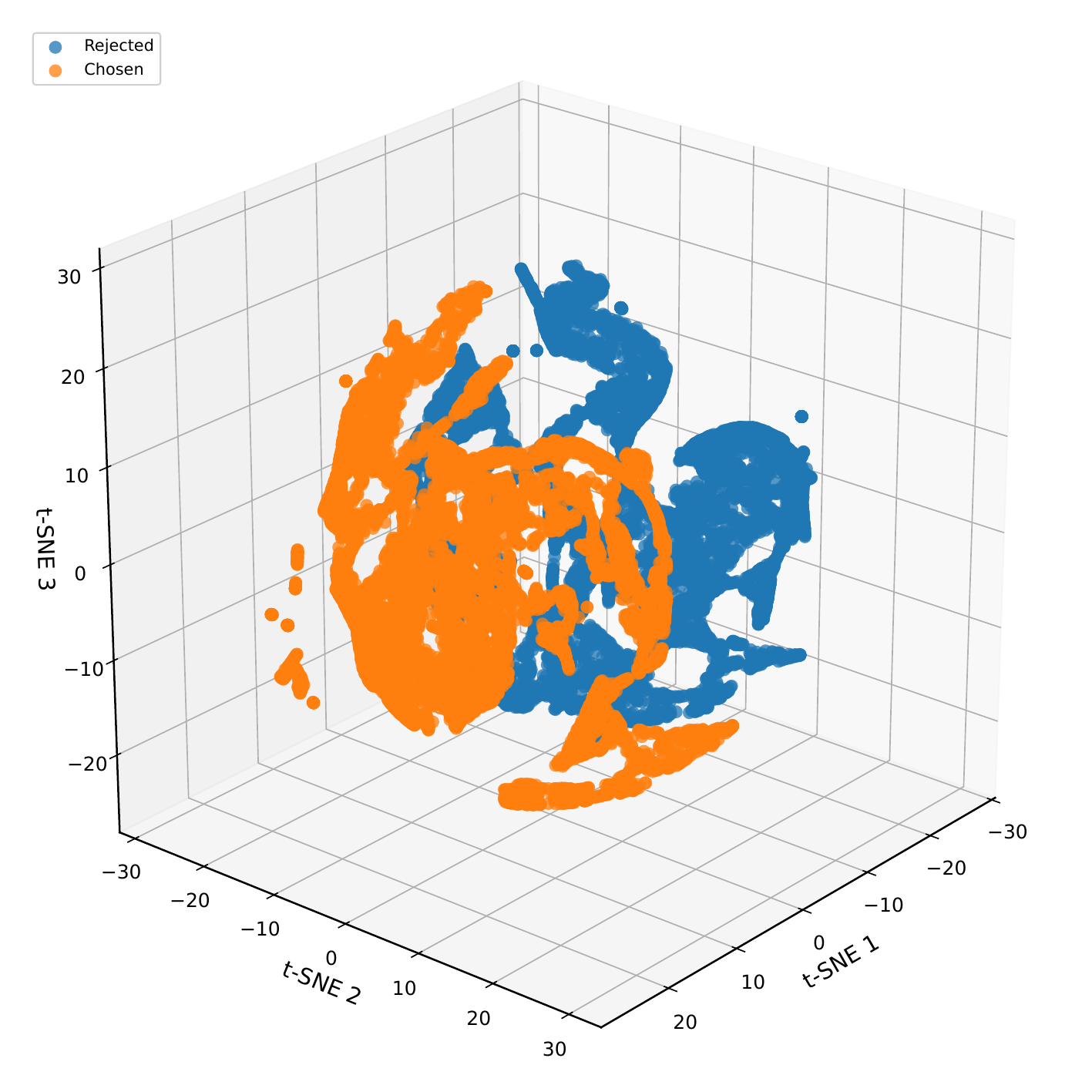}
        \caption{ORPO}
    \end{subfigure}
    \hfill
    \begin{subfigure}[t]{0.32\linewidth}
        \includegraphics[width=\linewidth]{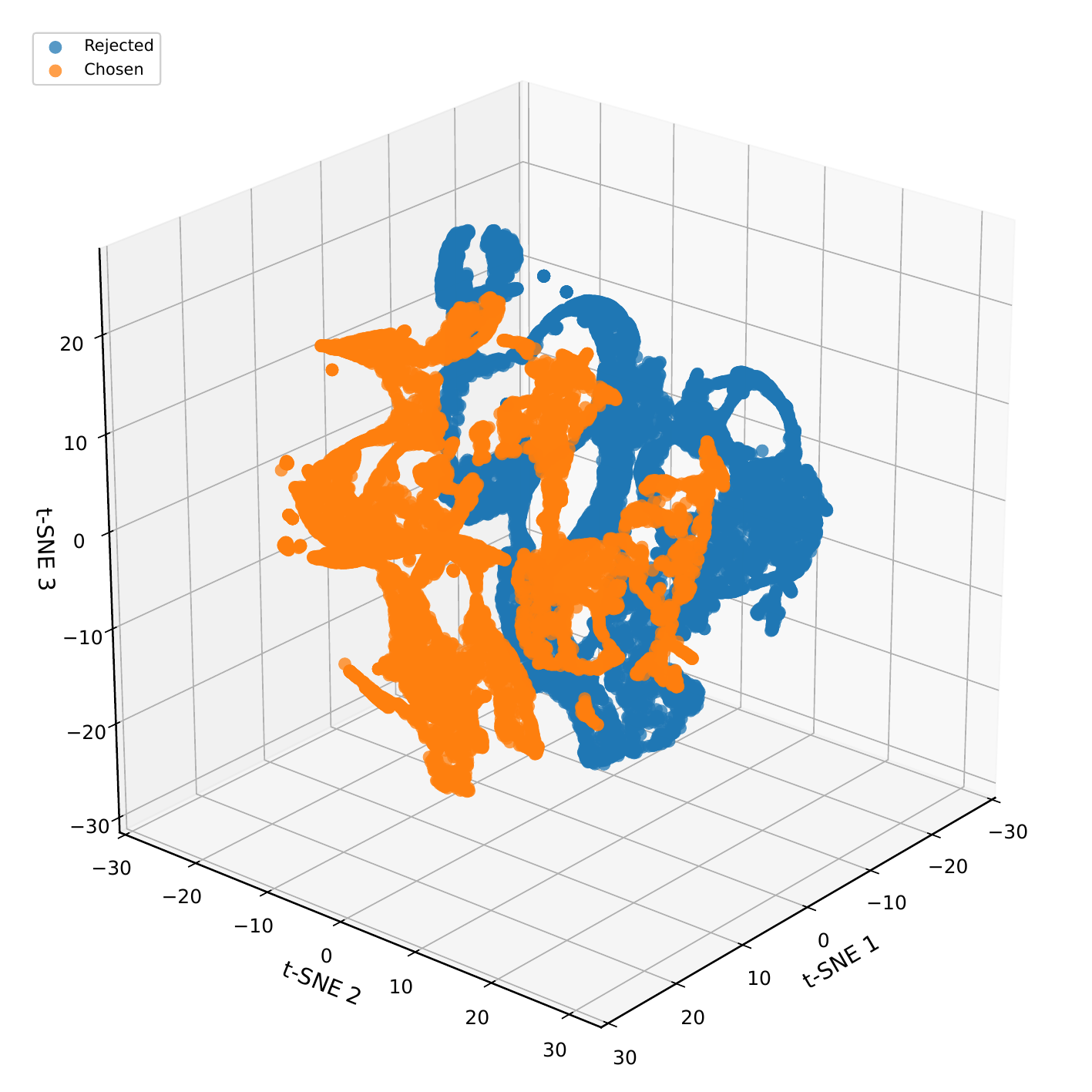}
        \caption{PPO}
    \end{subfigure}
    \hfill
    \begin{subfigure}[t]{0.32\linewidth}
        \includegraphics[width=\linewidth]{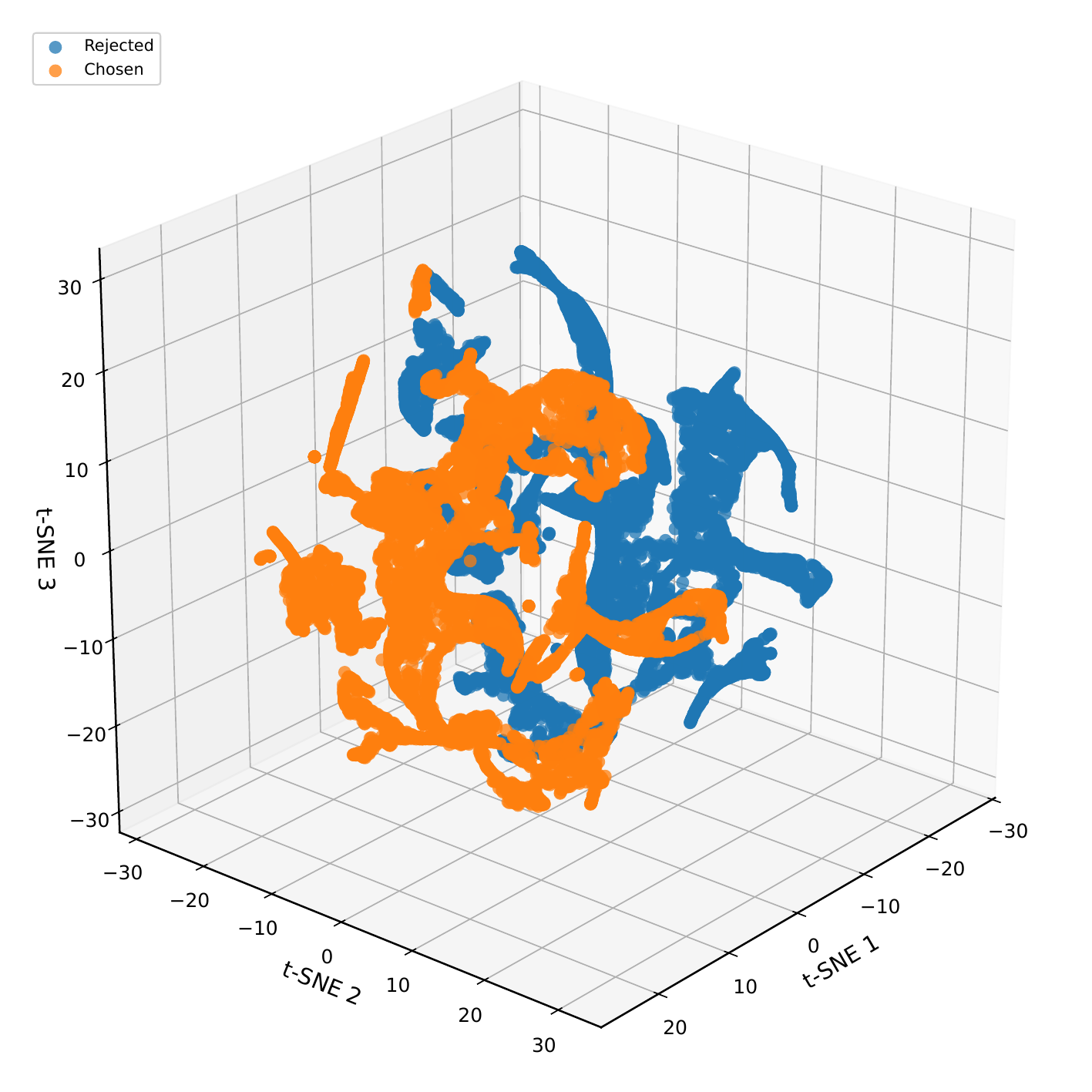}
        \caption{SimPO}
    \end{subfigure}
    \caption{%
        \textbf{Llama-3.2-3B.} PCA of residual-stream activations at each method's best probe layer..
    }
    \label{fig:lp_pca_grid_llama}
\end{figure}

\begin{figure}[!ht]
    \centering
    \begin{subfigure}[t]{0.32\linewidth}
        \includegraphics[width=\linewidth]{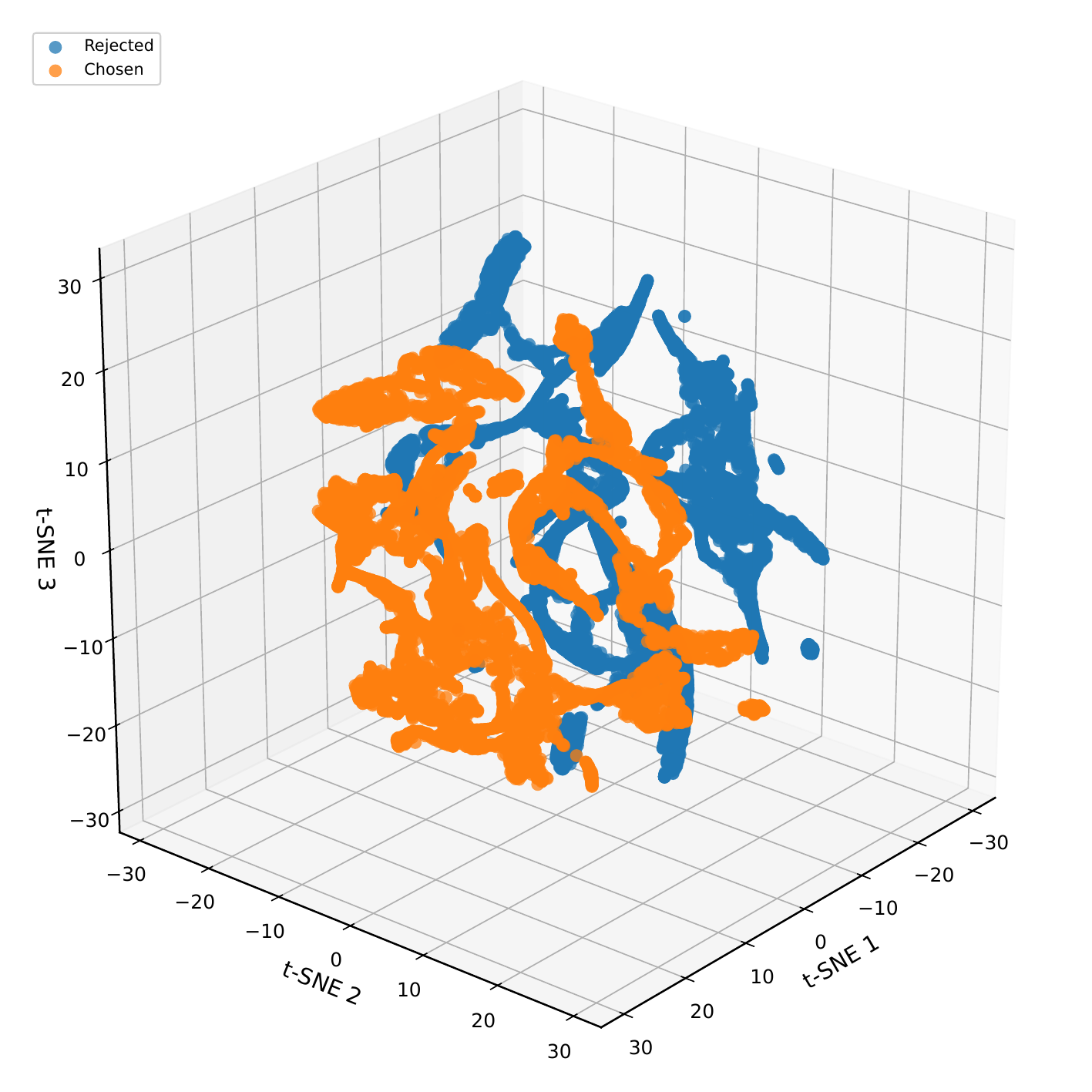}
        \caption{DPO}
    \end{subfigure}
    \hfill
    \begin{subfigure}[t]{0.32\linewidth}
        \includegraphics[width=\linewidth]{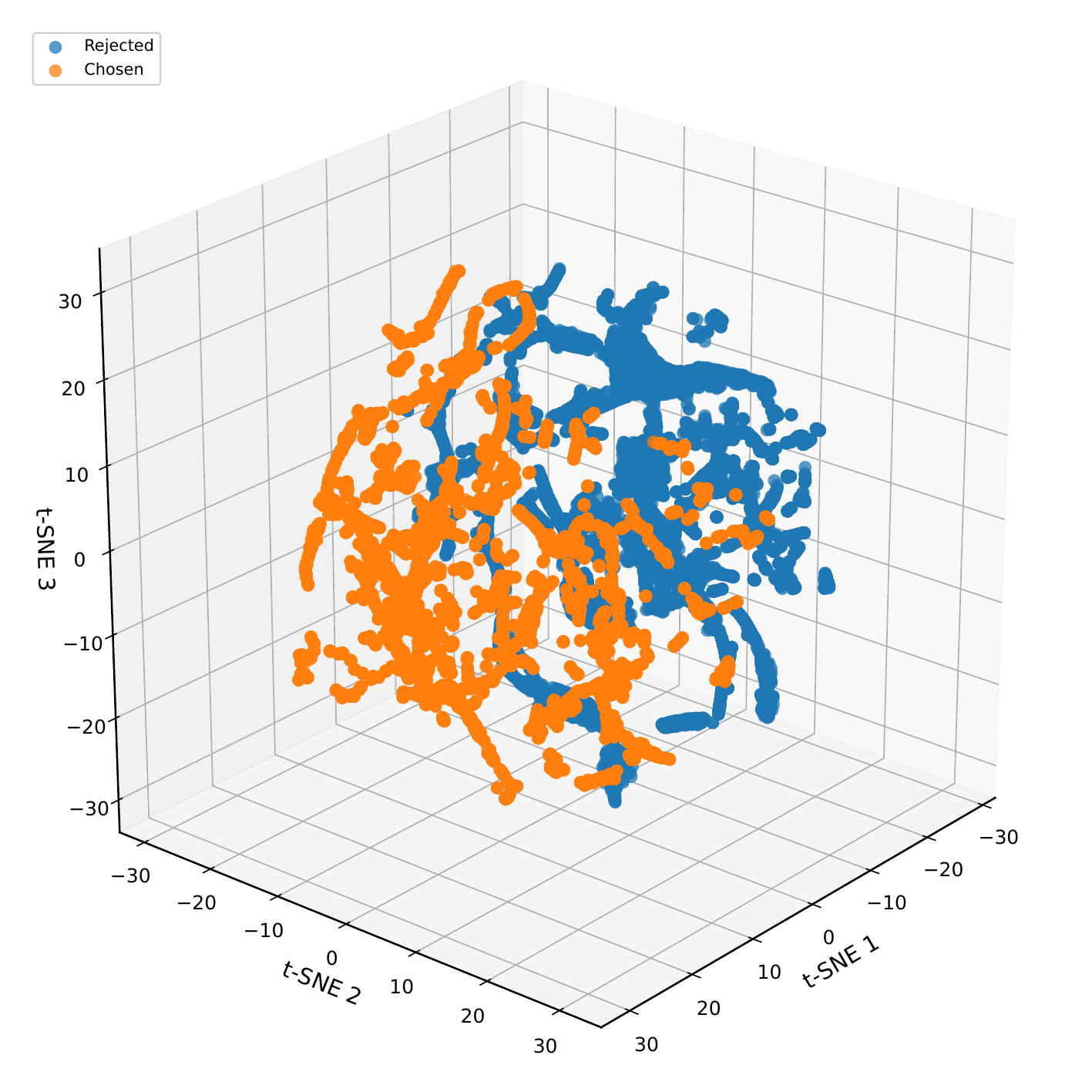}
        \caption{GRPO}
    \end{subfigure}
    \hfill
    \begin{subfigure}[t]{0.32\linewidth}
        \includegraphics[width=\linewidth]{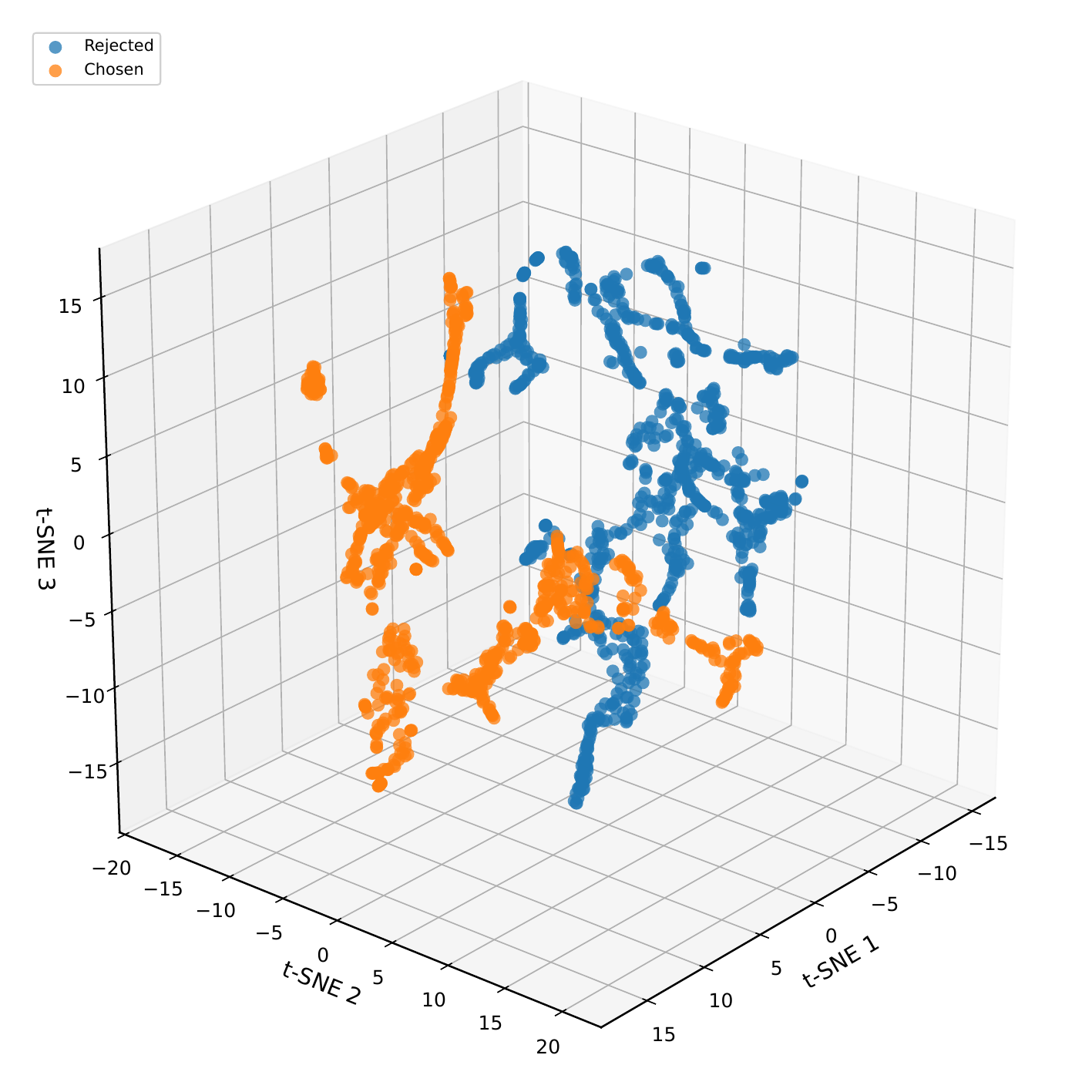}
        \caption{KTO}
    \end{subfigure}

    \vspace{0.9em}
    \begin{subfigure}[t]{0.32\linewidth}
        \includegraphics[width=\linewidth]{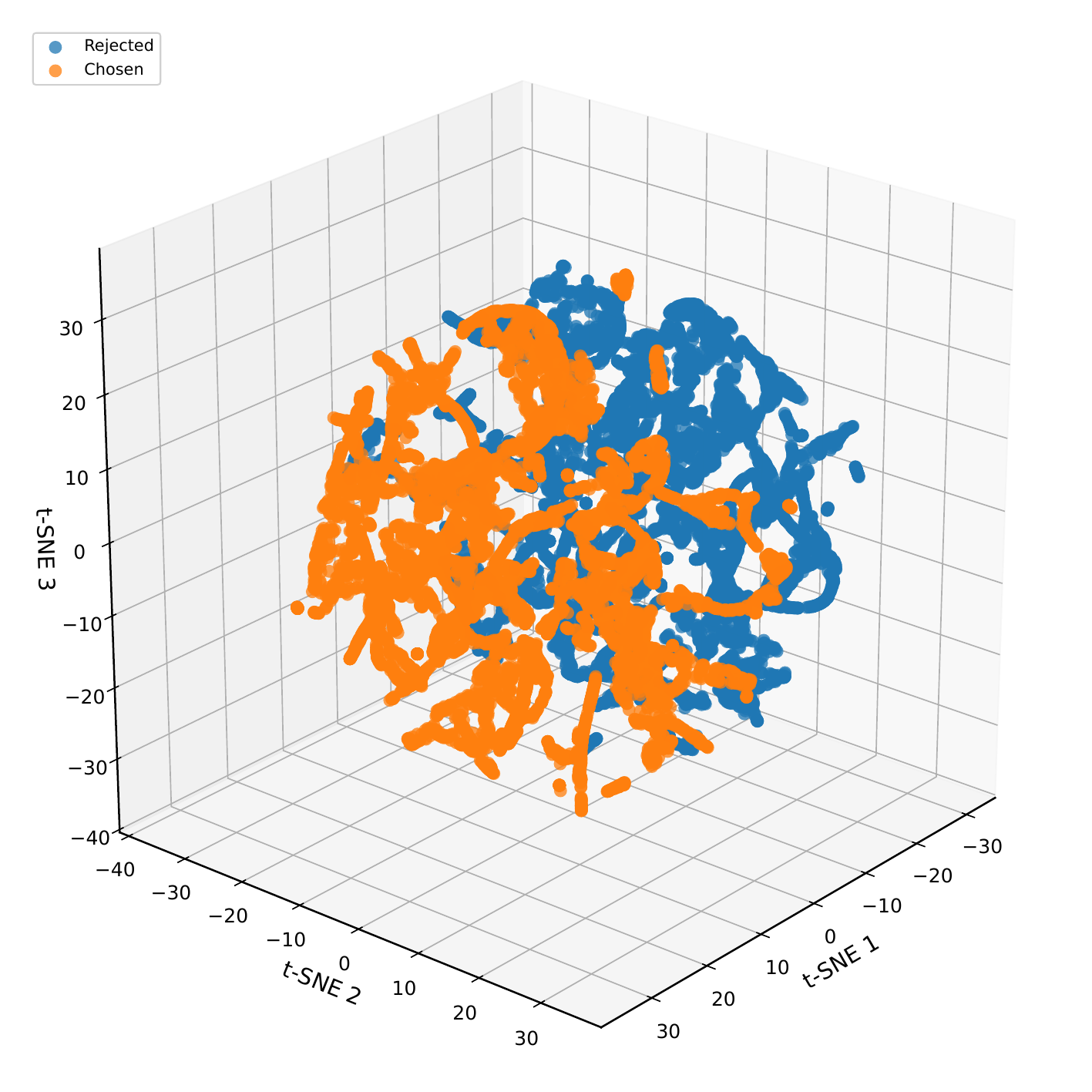}
        \caption{ORPO}
    \end{subfigure}
    \hfill
    \begin{subfigure}[t]{0.32\linewidth}
        \includegraphics[width=\linewidth]{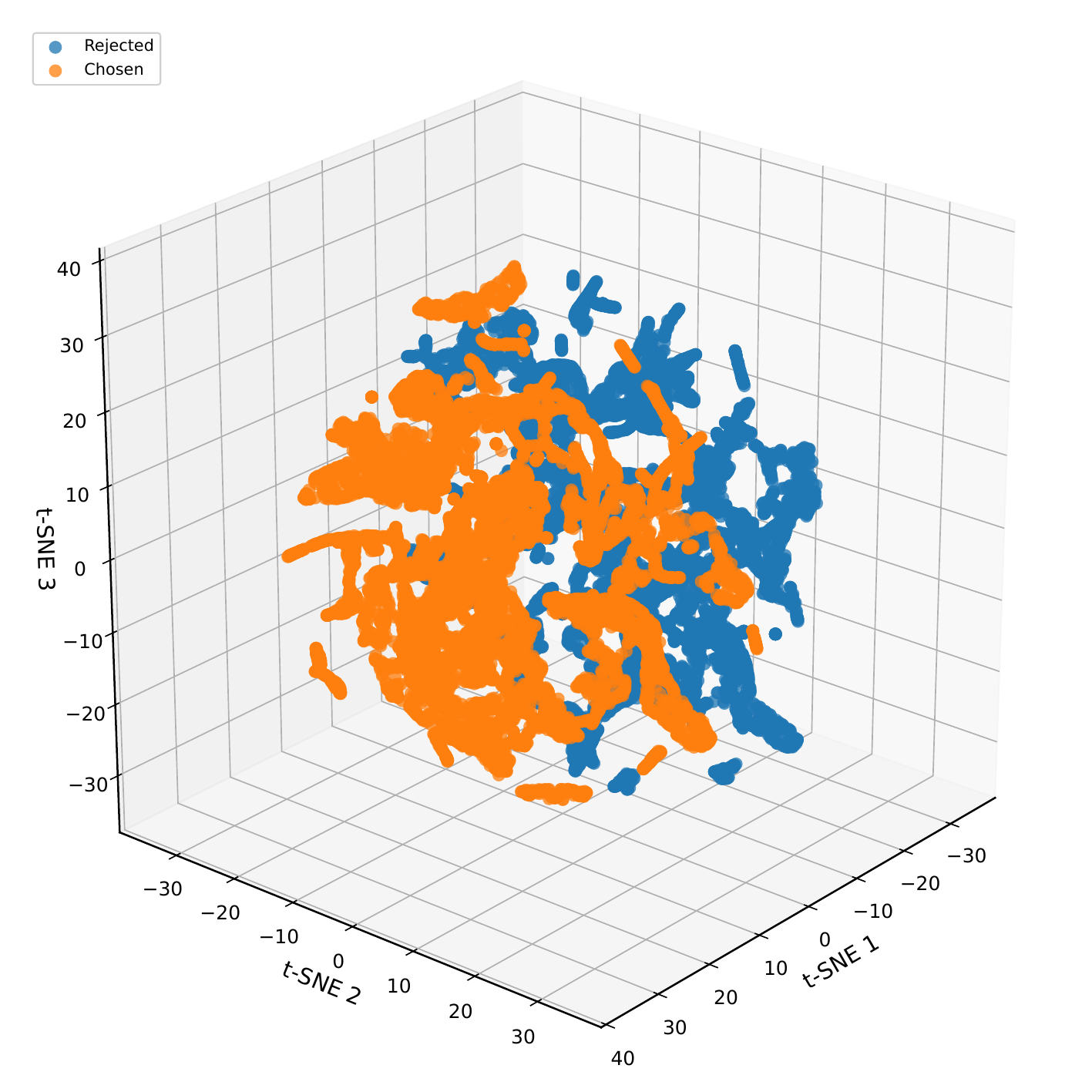}
        \caption{PPO}
    \end{subfigure}
    \hfill
    \begin{subfigure}[t]{0.32\linewidth}
        \includegraphics[width=\linewidth]{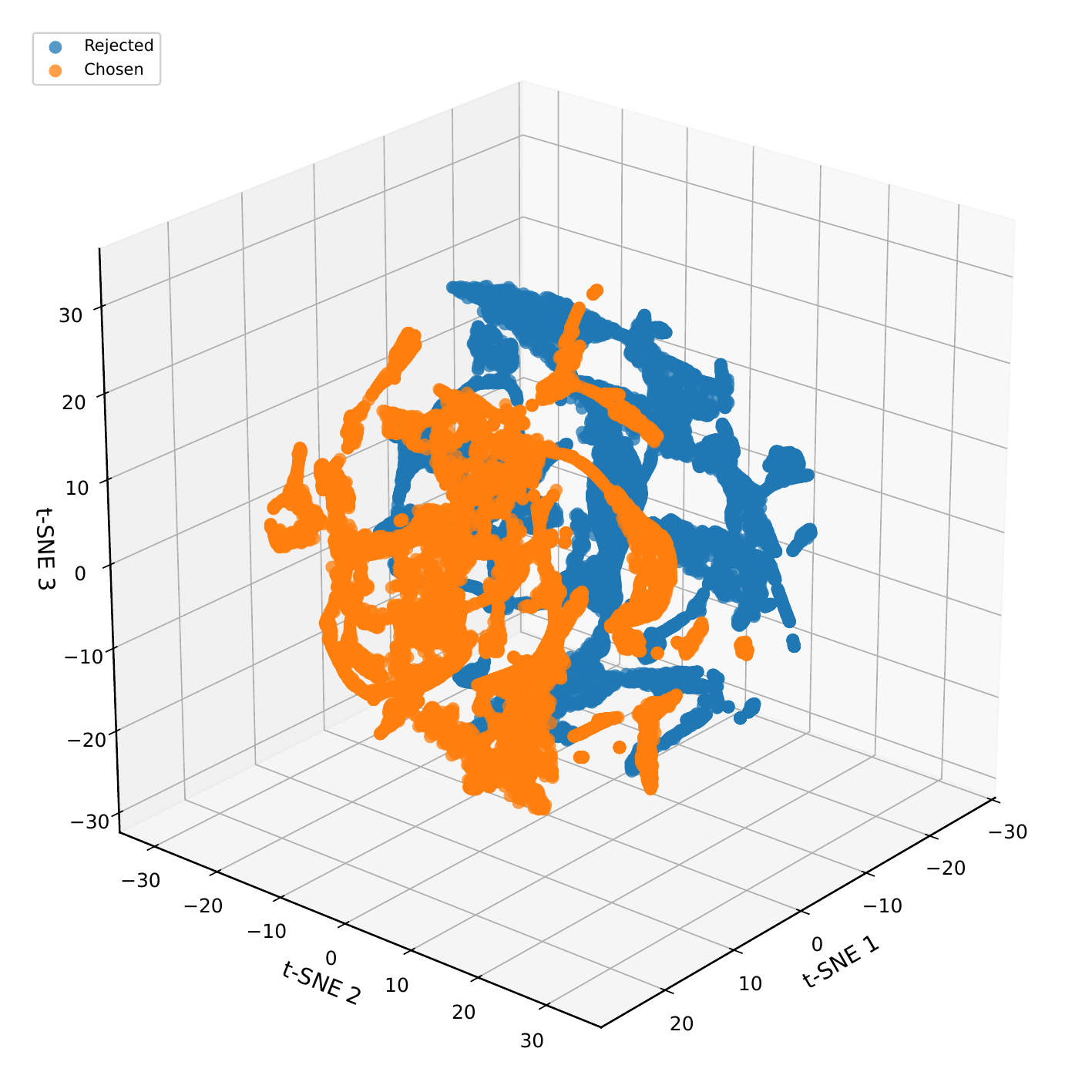}
        \caption{SimPO}
    \end{subfigure}
    \caption{%
        \textbf{Qwen3-4B.} PCA of residual-stream activations at each method's best probe layer.
    }
    \label{fig:lp_pca_grid_qwen}
\end{figure}

Next, Figures~\ref{fig:lp_pca_grid_smollm3}--\ref{fig:lp_pca_grid_qwen} show the PCA projections of the residual stream's activations (at ``best'' layer).

\clearpage
\section{Sparse Autoencoders}
\label{app:sae_metrics}

\subsection{Hyperparameters for Training} 
Dictionary size $d_{\text{SAE}}=4096$, sparsity $K=64$, training tokens $200,000$, batch size $2048$ tokens, Adam optimizer ($\beta_1=0.9, \beta_2=0.999$), peak learning rate $3 \times 10^{-4}$, linear warmup $100$ steps, auxiliary loss coefficient $1.0$, decoder initialization norm $0.1$, Top-$K$ threshold learning rate $0.01$.

\subsection{Training Dynamics and Extended Analysis}
Because all SAEs use the same hard-sparsity constraint ($K = 64$), the $L_0$ norm is identical across every model by construction and is therefore not a comparable axis. The informative sparsity metric in this configuration is $L_1$, which varies by more than an order of magnitude depending on the alignment algorithm used (as highlighted in the main text). 

\begin{figure}[!ht]
    \centering
    \includegraphics[width=\linewidth]{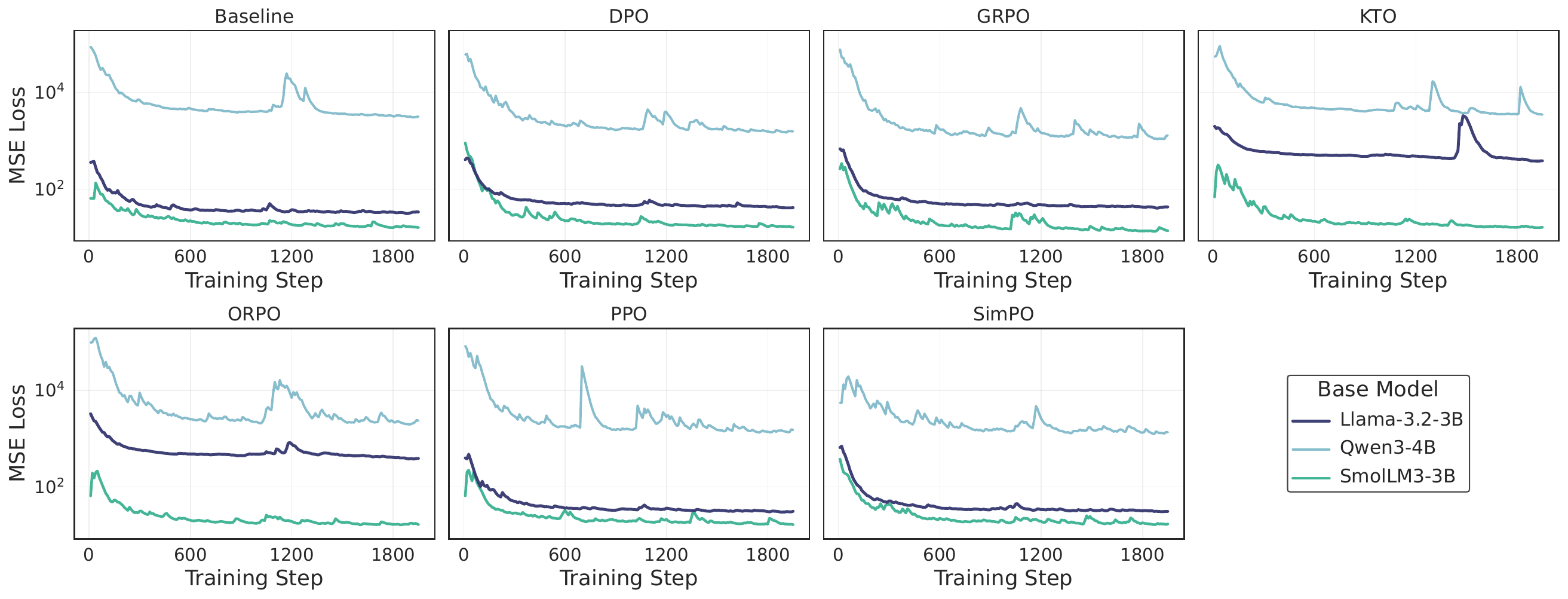}
    \caption{SAE reconstruction error (MSE) over training steps for each (base model, alignment) pair. All models use Batch Top-$K$ SAEs ($K=64$) trained at the probe-optimal layer. Alignment consistently increases asymptotic MSE and slows convergence, with KTO and ORPO exhibiting the largest penalties on Llama-3.2-3B and Qwen3-4B. Shaded regions denote $\pm 1$ SD over 3 random seeds.}
    \label{fig:sae_mse_loss}
\end{figure}

\paragraph{Alignment increases latent space complexity.}
As shown in \cref{fig:sae_mse_loss}, alignment consistently degrades SAE trainability: post-alignment models exhibit slower convergence and higher asymptotic MSE reconstruction error compared to their baselines. Across all three architectures, aligned models require $\sim$2--3$\times$ more training steps to reach plateau, and final MSE values are elevated by 0.3--0.8 log-units. This pattern holds uniformly across DPO, GRPO, PPO, and SimPO, indicating that preference optimization introduces non-stationarities that expand the effective dimensionality of the activation manifold. Notably, KTO and ORPO show the largest MSE penalties on Llama-3.2-3B and Qwen3-4B, suggesting that their more aggressive objective formulations induce greater representational distortion.

\begin{figure}[!ht]
    \centering
    \begin{subfigure}[b]{0.48\linewidth}
        \includegraphics[width=\linewidth]{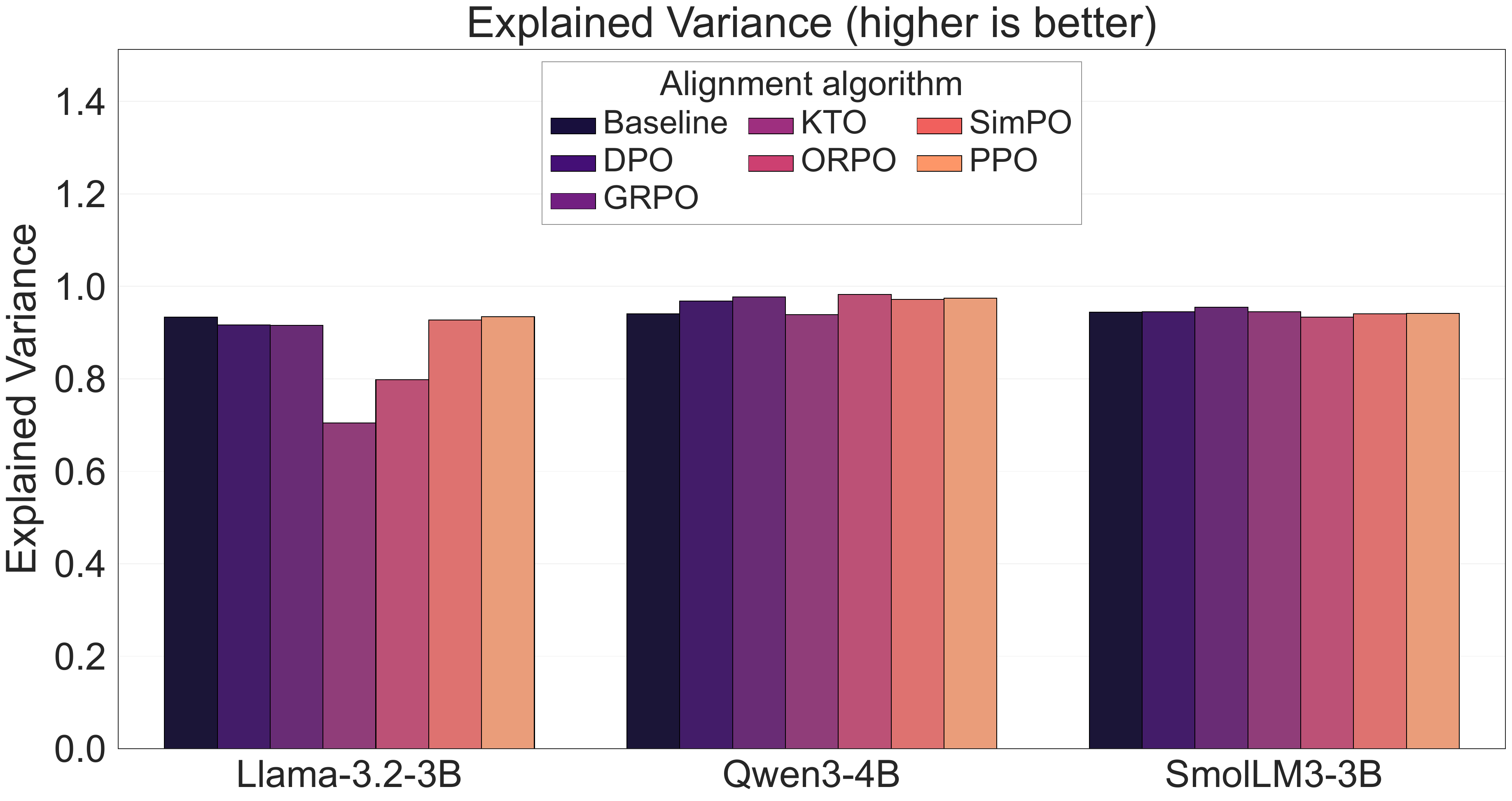}
        \caption{Explained Variance (Higher is better)}
        \label{fig:sae_bar_ev}
    \end{subfigure}
    \hfill
    \begin{subfigure}[b]{0.48\linewidth}
        \includegraphics[width=\linewidth]{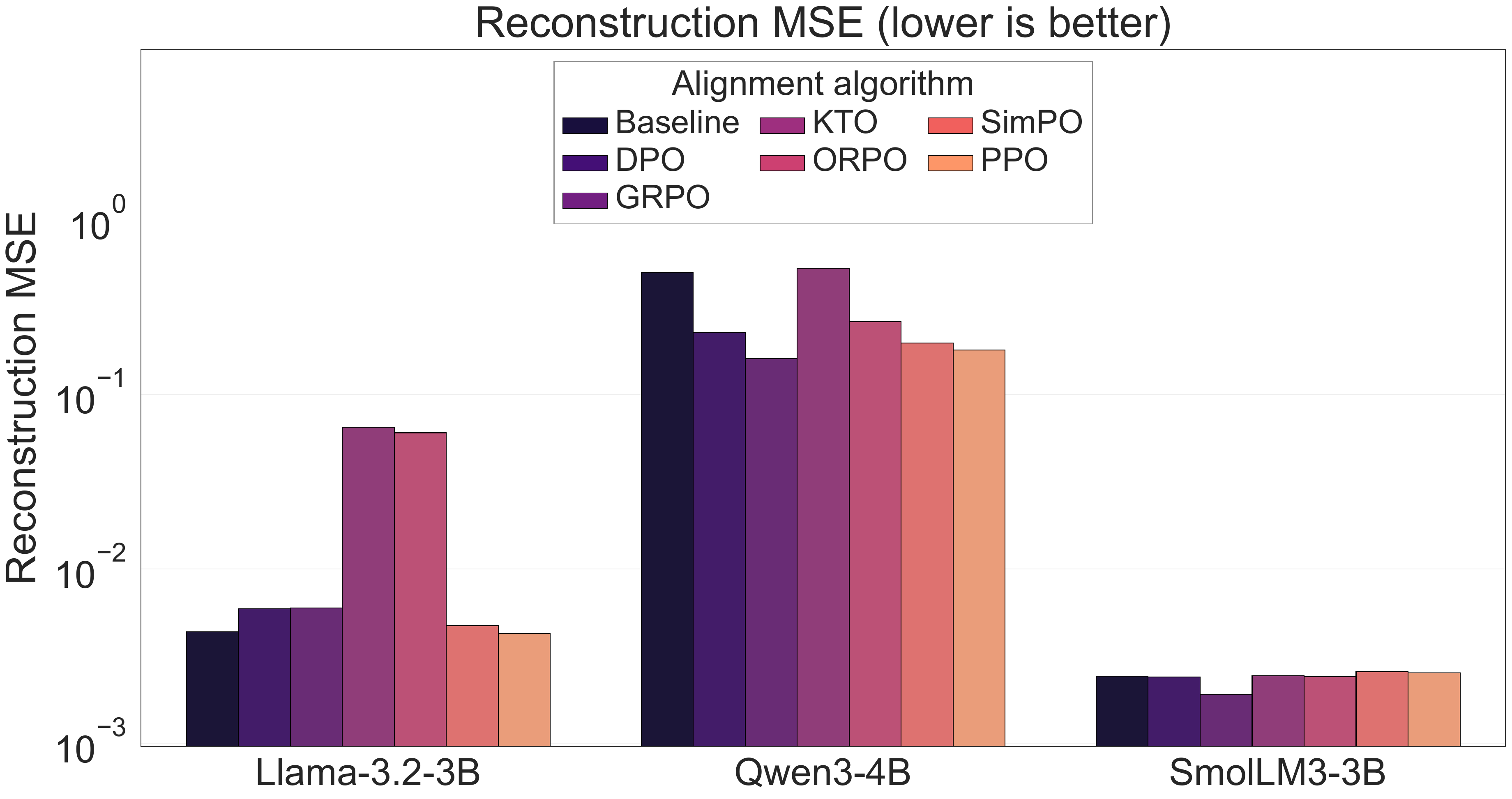}
        \caption{Reconstruction MSE (Lower is better)}
        \label{fig:sae_bar_mse}
    \end{subfigure}
    
    \vspace{1em}
    \begin{subfigure}[b]{0.48\linewidth}
        \includegraphics[width=\linewidth]{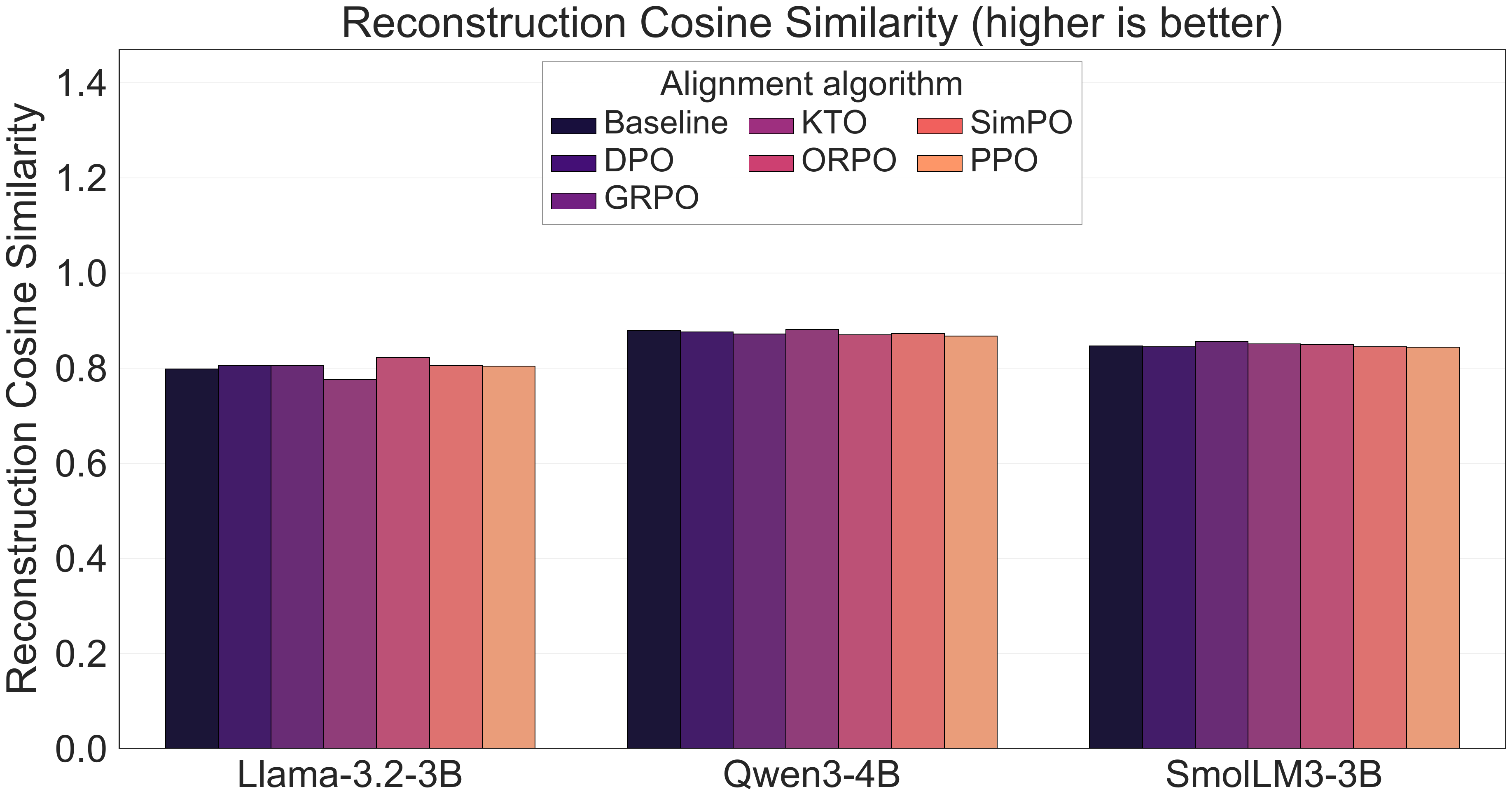}
        \caption{Cosine Similarity (Higher is better)}
        \label{fig:sae_bar_cossim}
    \end{subfigure}
    \hfill
    \begin{subfigure}[b]{0.48\linewidth}
        \includegraphics[width=\linewidth]{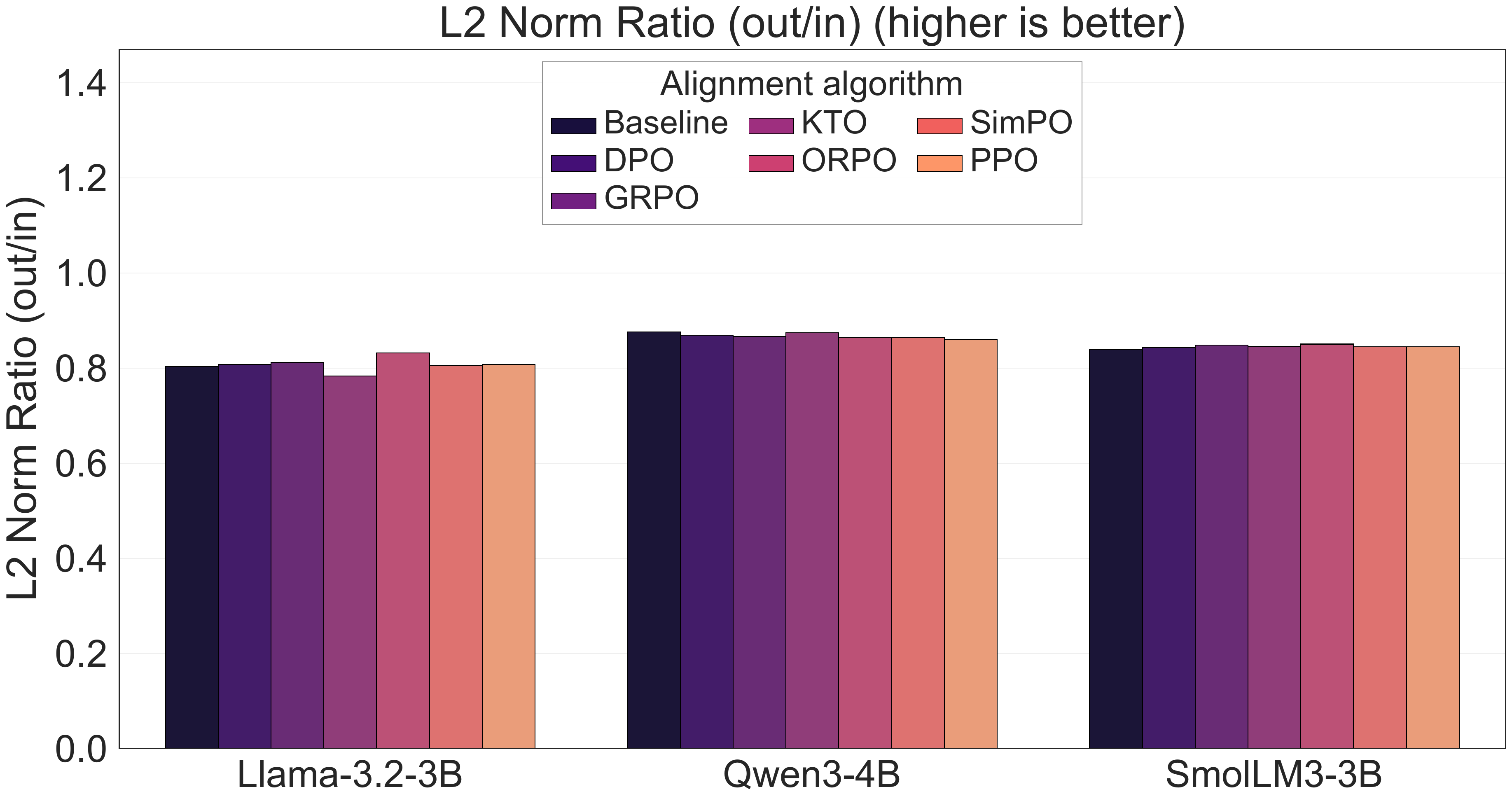}
        \caption{$L_2$ Norm Ratio (Higher is better)}
        \label{fig:sae_bar_l2}
    \end{subfigure}
    
    \vspace{1em}
    \begin{subfigure}[b]{0.48\linewidth}
        \includegraphics[width=\linewidth]{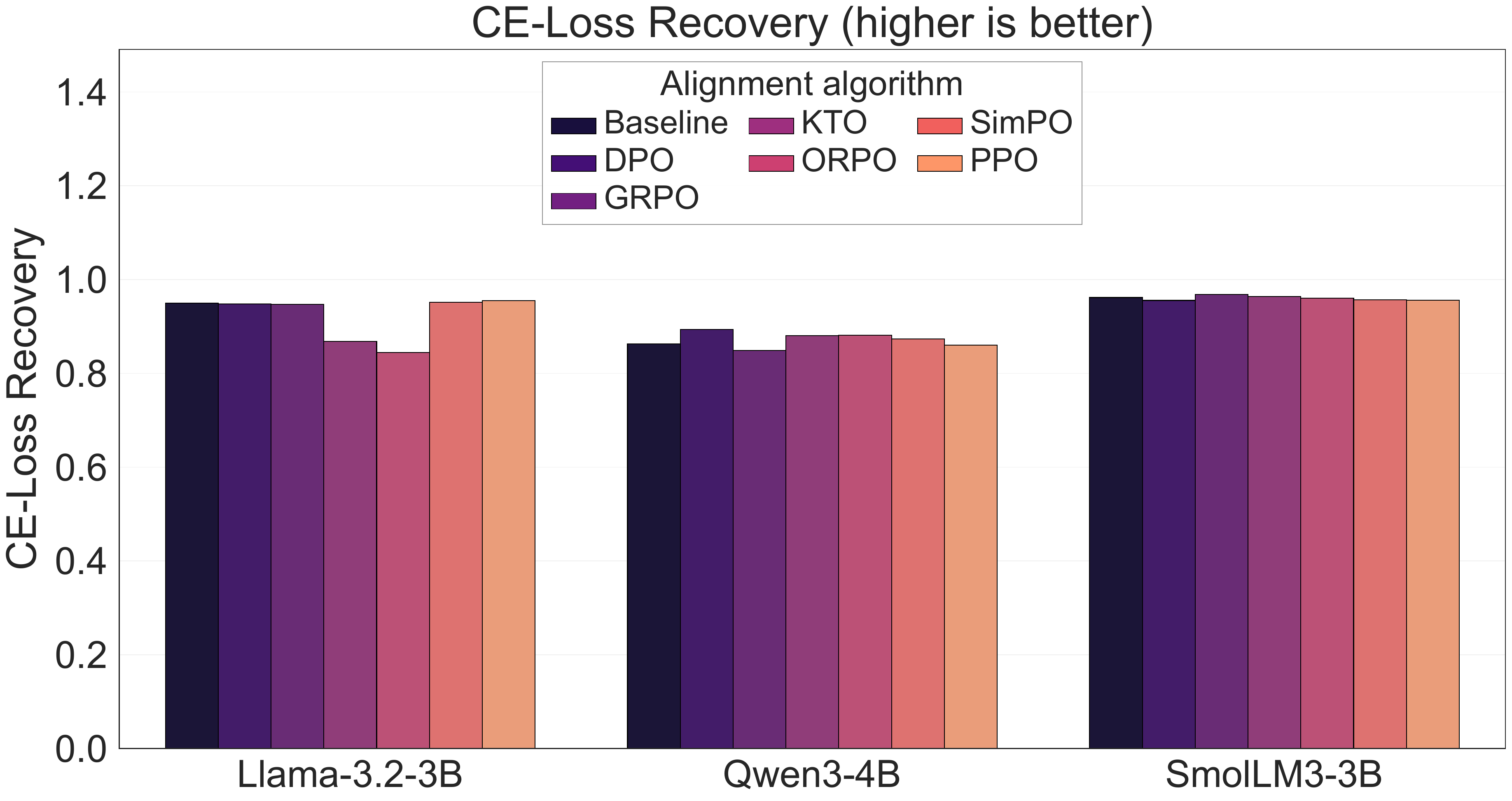}
        \caption{CE-Loss Recovery (Higher is better)}
        \label{fig:sae_bar_ce_loss}
    \end{subfigure}
    \hfill
    \begin{subfigure}[b]{0.48\linewidth}
        \includegraphics[width=\linewidth]{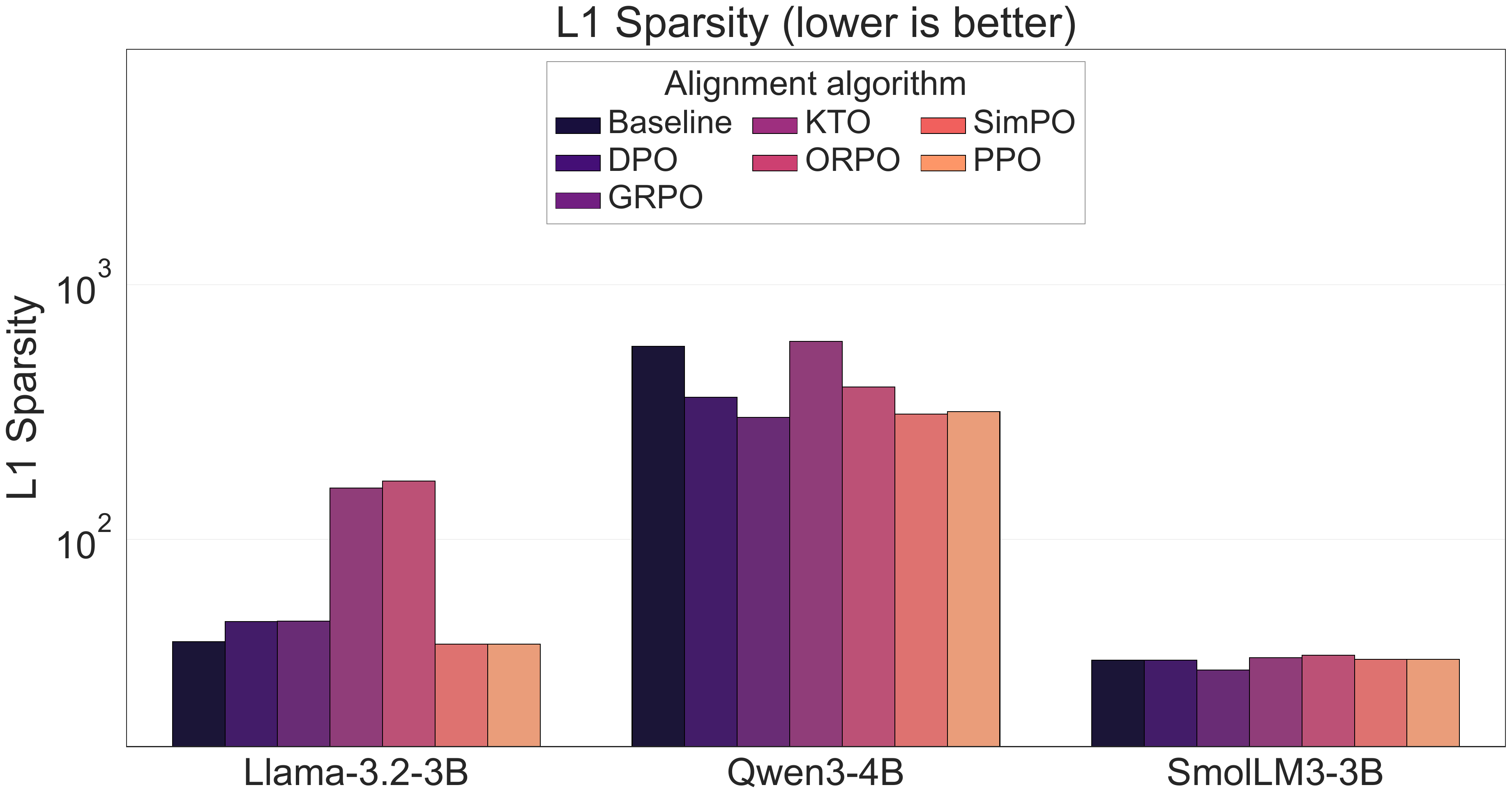}
        \caption{$L_1$ Sparsity (Lower is better)}
        \label{fig:sae_bar_l1}
    \end{subfigure}
    
    \caption{Comprehensive SAE reconstruction and preservation metrics. Across almost all metrics, concentrated-modification algorithms like KTO and ORPO (especially on Llama-3.2) show the largest deviations from baseline performance, indicating intense representational shifts that complicate SAE fidelity. $L_1$ sparsity penalties (log scale) further show that these methods induce higher-magnitude activations despite fixed $K=64$.}
    \label{fig:sae_bar_metrics}
\end{figure}

\paragraph{Algorithm choice drives sparsity--fidelity trade-offs.}
While all SAEs are constrained to $K=64$ active features, the $L_1$ penalty on encoder weights a proxy for feature activation magnitude varies dramatically by algorithm. \Cref{fig:sae_bar_l1} reveals a sharp dichotomy: on Llama-3.2-3B and Qwen3-4B, KTO and ORPO induce $L_1$ penalties $\sim$4$\times$ higher than DPO, GRPO, PPO, or SimPO (log-scale difference of $\approx 0.6$). This indicates that KTO/ORPO concentrate preference signals into fewer, higher-magnitude feature directions, whereas other methods distribute adjustments more diffusely. SmolLM3-3B exhibits attenuated algorithmic variance, suggesting smaller models may have less capacity to support highly specialized representational updates.

\paragraph{Interpretability implications.}
These findings expose a fundamental tension: alignment improves behavioral metrics but complicates mechanistic interpretability. Higher MSE implies that post-alignment activations are less efficiently compressible by a fixed-capacity SAE, while elevated $L_1$ norms for KTO/ORPO suggest that their preference representations are encoded in sparse, high-salience features that may be harder to localize and ablate. Critically, the algorithm-specific patterns in \cref{fig:sae_mse_loss,fig:sae_bar_l1} motivate our crosscoder analysis (\S\ref{sec:crosscoder}): if alignment methods reshape feature dictionaries in divergent ways, direct feature-matching between base and aligned models becomes essential for tracking representational drift.

This section provides a comprehensive breakdown of SAE performance across reconstruction and downstream behavioral preservation metrics. Figure~\ref{fig:sae_bar_metrics} maps out how different alignment objectives degrade or alter the underlying feature landscape across five key metrics, while Figure~\ref{fig:sae_mse_loss} illustrates the training dynamics and instability induced by aggressive behavioral fine-tuning. Finally, Table~\ref{tab:app_sae_eval} provides the precise numerical grid.

\begin{table}[h]
\centering
\caption{Complete SAE evaluation at the probe-best layer of each model. CE-loss score captures downstream language-modeling preservation; FVE represents the fraction of variance explained to measure the reconstruction quality. }
\label{tab:app_sae_eval}
\small
\begin{tabular}{llrrrrrr}
\toprule
Base & Algorithm & Layer & CE-score & FVE & MSE & L2 ratio & L1 \\
\midrule
SmolLM3-3B & Baseline & 19 & 0.962 & 0.944 & 0.0024 & 0.839 & 33.5 \\
SmolLM3-3B & DPO & 18 & 0.956 & 0.945 & 0.0024 & 0.843 & 33.6 \\
SmolLM3-3B & GRPO & 17 & 0.968 & 0.955 & 0.0019 & 0.848 & 30.7 \\
SmolLM3-3B & KTO & 19 & 0.964 & 0.945 & 0.0024 & 0.846 & 34.4 \\
SmolLM3-3B & ORPO & 18 & 0.961 & 0.933 & 0.0024 & 0.850 & 35.1 \\
SmolLM3-3B & SimPO & 18 & 0.957 & 0.941 & 0.0026 & 0.845 & 33.8 \\
\midrule
Llama-3.2-3B & Baseline & 11 & 0.950 & 0.933 & 0.0044 & 0.803 & 39.7 \\
Llama-3.2-3B & DPO & 13 & 0.948 & 0.917 & 0.0059 & 0.808 & 47.5 \\
Llama-3.2-3B & GRPO & 13 & 0.947 & 0.916 & 0.0060 & 0.812 & 47.7 \\
Llama-3.2-3B & KTO & 24 & 0.868 & 0.705 & 0.0647 & 0.783 & 158.9 \\
Llama-3.2-3B & ORPO & 25 & 0.844 & 0.798 & 0.0603 & 0.832 & 169.7 \\
Llama-3.2-3B & SimPO & 11 & 0.952 & 0.927 & 0.0047 & 0.805 & 38.8 \\
\midrule
Qwen3-4B & Baseline & 24 & 0.863 & 0.941 & 0.5019 & 0.876 & 572.5 \\
Qwen3-4B & DPO & 22 & 0.894 & 0.969 & 0.2265 & 0.869 & 362.0 \\
Qwen3-4B & GRPO & 20 & 0.849 & 0.977 & 0.1601 & 0.866 & 301.2 \\
Qwen3-4B & KTO & 24 & 0.881 & 0.938 & 0.5289 & 0.875 & 599.7 \\
Qwen3-4B & ORPO & 22 & 0.881 & 0.982 & 0.2611 & 0.865 & 396.8 \\
Qwen3-4B & SimPO & 21 & 0.873 & 0.972 & 0.1968 & 0.864 & 310.8 \\
\bottomrule
\end{tabular}
\end{table}

\begin{table*}[t]
\centering
\caption{Across models and alignment methods, mean-activated features consistently match the base anchor (rank 1). Max-activation shows variability but does not affect average identity.}
\label{tab:anchor_stability}
\small
\setlength{\tabcolsep}{5pt}
\renewcommand{\arraystretch}{1.15}

\begin{tabular}{l l c c c c c c}
\toprule
Family & Algo & Readout & Layer & Base Anchor & Mean Feature & Rank & Max Feature \\
\midrule

\multirow{6}{*}{Llama}
& DPO   & same/probe & 11/13 & 15132 & 15132 & 1 & 5272 \\
& PPO   & same/probe & 11/11 & 15132 & 15132 & 1 & 5272 \\
& KTO   & same/probe & 11/24 & 15132 & 15132 & 1 & 5272 \\
& GRPO  & same/probe & 11/13 & 15132 & 15132 & 1 & 5272 \\
& ORPO  & same/probe & 11/25 & 15132 & 15132 & 1 & 15132/15879 \\
& SimPO & same/probe & 11/11 & 15132 & 15132 & 1 & 5272 \\
\midrule

\multirow{6}{*}{Qwen}
& DPO   & same/probe & 24/22 & 6910 & 6910 & 1 & 3455 \\
& PPO   & same/probe & 24/21 & 6910 & 6910 & 1 & 3455 \\
& KTO   & same/probe & 24/24 & 6910 & 6910 & 1 & 3455 \\
& GRPO  & same/probe & 24/20 & 6910 & 6910 & 1 & 3455 \\
& ORPO  & same/probe & 24/22 & 6910 & 6910 & 1 & 3455 \\
& SimPO & same/probe & 24/21 & 6910 & 6910 & 1 & 3455 \\
\midrule

\multirow{6}{*}{SmolLM}
& DPO   & same/probe & 19/18 & 10154 & 10154 & 1 & 1428/10154 \\
& PPO   & same/probe & 19/18 & 10154 & 10154 & 1 & 1428/10154 \\
& KTO   & same/probe & 19/19 & 10154 & 10154 & 1 & 1428 \\
& GRPO  & same/probe & 19/17 & 10154 & 10154 & 1 & 1428/10154 \\
& ORPO  & same/probe & 19/18 & 10154 & 10154 & 1 & 1428/10154 \\
& SimPO & same/probe & 19/18 & 10154 & 10154 & 1 & 1428/10154 \\
\bottomrule
\end{tabular}
\end{table*}
\clearpage
\section{Crosscoder Details}
\label{app:crosscoder-details}

\subsection{Training and Hyperparameters}
\label{app:crosscoder-training-hyperparam}
The shared Sparse Autoencoder takes an expansion factor $\alpha=8$ (yielding $M = 8d$-dimensional latent space, given input dimensionality, $d$), Top-$K = 400$. From \cite{mishrasharma-2025}'s work, we use a $6\%$ forced shared fraction (of decoder columns), a shared-subspace multiplier $\lambda_{\text{shared}} = 0.05$, cross-reconstruction weight $\lambda_{\text{cross}} = 0.4$, $L_1$ sparsity weight $\lambda_{\text{sparse}} = 10^{-3}$. For training, we initialize the decoder norm to $0.1$ and train with a weight decay $10^{-5}$ and gradient clipping at norm $1.0$. The Adam optimizer has $\beta_1 = 0.9$, $\beta_2 = 0.999$, learning rate $3 \times 10^{-4}$ with $5\%$ warmup. Lastly, a batch size of $32$ is used for training over $4$ epochs.

\begin{table}[t!]
\centering
\caption{%
    Crosscoder training statistics for all (base model, alignment) pairs.
    FVE denotes the fractional variance explained for the base
    and aligned streams measured on the validation set for final checkpoint, $L_0$ denotes the mean number of active latents per sample,
    and Dead Frac denotes the fraction of dead neurons at the end of training.%
}
\label{tab:app_crosscoder_training_stats}
\small
\begingroup
\setlength{\tabcolsep}{4pt}
\begin{tabular}{llrrrrrr}
\toprule
Base & Algorithm & Layers & FVE Base & FVE Aligned & $L_0$ Base & $L_0$ Aligned & Dead Frac \\
\midrule
Llama-3.2-3B & DPO   & L12--14 & 0.7760 & 0.7697 & 213.3 & 212.1 & 0.9537 \\
Llama-3.2-3B & GRPO  & L12--14 & 0.7756 & 0.7701 & 214.0 & 212.4 & 0.9547 \\
Llama-3.2-3B & KTO   & L23--25 & 0.7094 & 0.7683 & 221.6 & 221.4 & 0.9672 \\
Llama-3.2-3B & ORPO  & L24--26 & 0.6937 & 0.8142 & 220.5 & 215.8 & 0.9580 \\
Llama-3.2-3B & PPO   & L10--12 & 0.7550 & 0.7482 & 200.0 & 198.7 & 0.9518 \\
Llama-3.2-3B & SimPO & L10--12 & 0.7551 & 0.7487 & 200.2 & 199.0 & 0.9518 \\
\midrule
Qwen3-4B     & DPO   & L21--23 & 0.8958 & 0.8953 & 211.6 & 211.5 & 0.9626 \\
Qwen3-4B     & GRPO  & L19--21 & 0.8942 & 0.8935 & 212.8 & 212.4 & 0.9636 \\
Qwen3-4B     & KTO   & L23--25 & 0.8979 & 0.9068 & 214.4 & 213.8 & 0.9562 \\
Qwen3-4B     & ORPO  & L21--23 & 0.8912 & 0.9013 & 223.4 & 224.7 & 0.9543 \\
Qwen3-4B     & PPO   & L20--22 & 0.8947 & 0.8936 & 212.3 & 212.1 & 0.9634 \\
Qwen3-4B     & SimPO & L20--22 & 0.8950 & 0.8944 & 211.9 & 211.7 & 0.9636 \\
\midrule
SmolLM3-3B   & DPO   & L17--19 & 0.8131 & 0.8086 & 214.4 & 213.2 & 0.9439 \\
SmolLM3-3B   & GRPO  & L16--18 & 0.8102 & 0.8038 & 210.1 & 207.2 & 0.9434 \\
SmolLM3-3B   & KTO   & L18--20 & 0.7872 & 0.7896 & 214.7 & 215.2 & 0.9548 \\
SmolLM3-3B   & ORPO  & L17--19 & 0.7959 & 0.7902 & 214.0 & 224.1 & 0.9500 \\
SmolLM3-3B   & PPO   & L17--19 & 0.8139 & 0.8107 & 215.1 & 214.0 & 0.9443 \\
SmolLM3-3B   & SimPO & L17--19 & 0.8138 & 0.8096 & 214.7 & 213.6 & 0.9440 \\
\bottomrule
\end{tabular}
\endgroup
\end{table}

\begin{table}[t!]
\centering
\caption{%
    Run-specific GMM-adaptive $\rho$ thresholds used for Crosscoder feature
    classification. Features below $\rho_{\text{base}}$ are assigned to the
    base-only class, features above $\rho_{\text{aligned}}$ are assigned to the
    aligned-only class, and intermediate values are partitioned using the shared
    thresholds.%
}
\label{tab:app_crosscoder_rho_thresholds}
\small
\begin{tabular}{llrrrrr}
\toprule
Base & Algorithm & Layers & $\rho_{\text{base}}$ & $\rho_{\text{sh-low}}$ & $\rho_{\text{sh-high}}$ & $\rho_{\text{aligned}}$ \\
\midrule
Llama-3.2-3B & DPO   & L12--14 & 0.4000 & 0.4542 & 0.4000 & 0.5000 \\
Llama-3.2-3B & GRPO  & L12--14 & 0.4000 & 0.4532 & 0.4000 & 0.5000 \\
Llama-3.2-3B & KTO   & L23--25 & 0.4000 & 0.4187 & 0.4000 & 0.5000 \\
Llama-3.2-3B & ORPO  & L24--26 & 0.4000 & 0.4582 & 0.7819 & 0.7819 \\
Llama-3.2-3B & PPO   & L10--12 & 0.4000 & 0.4492 & 0.4000 & 0.5000 \\
Llama-3.2-3B & SimPO & L10--12 & 0.4000 & 0.4497 & 0.4000 & 0.5000 \\
\midrule
Qwen3-4B     & DPO   & L21--23 & 0.4000 & 0.4612 & 0.4067 & 0.5000 \\
Qwen3-4B     & GRPO  & L19--21 & 0.4000 & 0.4442 & 0.4242 & 0.5000 \\
Qwen3-4B     & KTO   & L23--25 & 0.4000 & 0.4527 & 0.4437 & 0.5000 \\
Qwen3-4B     & ORPO  & L21--23 & 0.4000 & 0.4657 & 0.4367 & 0.5000 \\
Qwen3-4B     & PPO   & L20--22 & 0.4000 & 0.4607 & 0.4167 & 0.5000 \\
Qwen3-4B     & SimPO & L20--22 & 0.4000 & 0.4622 & 0.4187 & 0.5000 \\
\midrule
SmolLM3-3B   & DPO   & L17--19 & 0.4000 & 0.4297 & 0.4000 & 0.5000 \\
SmolLM3-3B   & GRPO  & L16--18 & 0.4000 & 0.4257 & 0.4000 & 0.5000 \\
SmolLM3-3B   & KTO   & L18--20 & 0.4000 & 0.4482 & 0.4000 & 0.5000 \\
SmolLM3-3B   & ORPO  & L17--19 & 0.4000 & 0.4352 & 0.4000 & 0.5000 \\
SmolLM3-3B   & PPO   & L17--19 & 0.4000 & 0.4312 & 0.4000 & 0.5000 \\
SmolLM3-3B   & SimPO & L17--19 & 0.4000 & 0.4312 & 0.4000 & 0.5000 \\
\bottomrule
\end{tabular}
\end{table}

\subsection{Crosscoder Metrics and Decision Thresholds}
\label{app:crosscoder_metrics}
After training, the final crosscoder statistics on the 10\% validation set are reported in Table~\ref{tab:app_crosscoder_training_stats}. These are used to choose the best SAE architecture. We deem a particular architecture ($\alpha, K$) valid if the fraction of variance explained (FVE) is above 0.75 and $L_0 \leq 250$ for both the base and aligned versions. For identifying the feature geometry on the 20\% held-out set, the classification thresholds are reported in Table~\ref{tab:app_crosscoder_rho_thresholds}.

\clearpage
\section{LLM Usage}
\label{app:llm_usage}

Large language models were used as writing assistants for editing, grammar correction, and improving clarity and flow. They were not used to generate experimental results, run analyses, or make scientific conclusions. All technical content, claims, experiments, and final wording were reviewed and approved by the authors, who take full responsibility for the paper.



\end{document}